%% file: main.tex
\documentclass[a4paper, twocolumn, 10pt]{article}
\usepackage{times}
\usepackage{flushend}
\usepackage{natbib}
\usepackage{authblk}

\title{Policy Search with High-Dimensional Context Variables}
\author[1]{Voot Tangkaratt} 
\author[2]{Herke van Hoof}
\author[2]{Simone Parisi}
\author[2]{\\Gerhard Neumann}
\author[2]{Jan Peters}
\author[3,4,1]{Masashi Sugiyama}

\affil[1]{Department of Computer Science, The University of Tokyo}
\affil[2]{Department of Computer Science, Technical University of Darmstadt}
\affil[3]{Center for Advanced Integrated Intelligence Research, RIKEN}
\affil[4]{Department of Complexity Science and Engineering, The University of Tokyo}

\date{}

\usepackage{amsmath}
\usepackage{amsfonts}
\usepackage{amssymb}
\usepackage{graphicx}
\usepackage[ruled, vlined, linesnumbered]{algorithm2e}
\usepackage[subrefformat=parens]{subcaption} 

\newcommand{\argmin}{\mathop{\mathrm{argmin}}}

\newcommand{\bb}{\boldsymbol{b}}
\newcommand{\bc}{\boldsymbol{c}}

\newcommand{\bff}{\boldsymbol{f}}

\newcommand{\bm}{\boldsymbol{m}}
\newcommand{\br}{\boldsymbol{r}}

\newcommand{\bx}{\boldsymbol{x}}
\newcommand{\by}{\boldsymbol{y}}

\newcommand{\btheta}{\boldsymbol{\theta}}

\newcommand{\bvphi}{\boldsymbol{\varphi}}

\newcommand{\bA}{\boldsymbol{A}}
\newcommand{\bB}{\boldsymbol{B}}

\newcommand{\bD}{\boldsymbol{D}}
\newcommand{\bF}{\boldsymbol{F}}

\newcommand{\bH}{\boldsymbol{H}}
\newcommand{\bI}{\boldsymbol{I}}
\newcommand{\bK}{\boldsymbol{K}}
\newcommand{\bL}{\boldsymbol{L}}
\newcommand{\bM}{\boldsymbol{M}}
\newcommand{\bP}{\boldsymbol{P}}
\newcommand{\bQ}{\boldsymbol{Q}}

\newcommand{\bU}{\boldsymbol{U}}
\newcommand{\bT}{\boldsymbol{T}}
\newcommand{\bV}{\boldsymbol{V}}
\newcommand{\bW}{\boldsymbol{W}}
\newcommand{\bX}{\boldsymbol{X}}
\newcommand{\bY}{\boldsymbol{Y}}
\newcommand{\bZ}{\boldsymbol{Z}}
\newcommand{\bSigma}{\boldsymbol{\Sigma}}

\newcommand{\dtheta}{\mathrm{d}\boldsymbol{\theta}}

\newcommand{\dc}{\mathrm{d}\boldsymbol{c}}

\newcommand{\cd}{{d}_{\boldsymbol{\mathrm{c}}}}
\newcommand{\trued}{{d}_{\boldsymbol{\mathrm{\tilde{c}}}}}
\newcommand{\thetad}{{d}_{\boldsymbol{\mathrm{\theta}}}}

\newcommand{\zd}{{d}_{\boldsymbol{\mathrm{z}}}}

\usepackage{wrapfig}
\usepackage{pgfplots, tikz}
\usetikzlibrary{plotmarks,spy,shadows}
\tikzset{spy using overlaysshadow/.style = 
  {spy scope = {#1,
                every spy on node/.style = {circle,
                                            fill,
                                            fill opacity = 0.2,
                                            text opacity = 1},
                every spy in node/.style = {circle,
                                            circular drop shadow,
                                            fill = white,
                                            draw,
                                            ultra thick,
                                            cap = round}
                }
  }
}

\usepackage[colorinlistoftodos, textwidth=15mm, textsize=footnotesize, shadow]{todonotes}

\begin{document}
\maketitle

\begin{abstract}
{
Direct contextual policy search methods learn to improve policy parameters and simultaneously generalize these parameters to different context or task variables. 
However, learning from high-dimensional context variables, such as camera images, is still a prominent problem in many real-world tasks.
A naive application of unsupervised dimensionality reduction methods to the context variables, such as principal component analysis, is insufficient as task-relevant input may be ignored.
In this paper, we propose a contextual policy search method in the model-based relative entropy stochastic search framework with integrated dimensionality reduction.
We learn a model of the reward that is locally quadratic in both the policy parameters and the context variables. 
Furthermore, we perform supervised linear dimensionality reduction on the context variables by nuclear norm regularization.
The experimental results show that the proposed method outperforms naive dimensionality reduction via principal component analysis and a state-of-the-art contextual policy search method.
}

\end{abstract}

\section{Introduction}

An autonomous agent often requires different policies for solving tasks with different contexts.
For instance, in a ball hitting task the robot has to adapt his controller according to the ball position, i.e., the context.
Direct policy search approaches \citep{Baxter00reinforcementlearning,DBLP:conf/ijcai/RosensteinB01,DBLP:journals/ftrob/DeisenrothNP13} allow the agent to learn a separate policy for each context through trial and error.
However, learning optimal policies for many large contexts, such as in the presence of continuous context variables, is impracticable. On the other hand, direct \emph{contextual} policy search approaches  \citep{DBLP:conf/ijcai/KoberOP11,DBLP:conf/icml/Neumann11,DBLP:conf/icml/SilvaKB12} represent the contexts by real-valued vectors and are able to learn a context-dependent distribution over the policy parameters. Such a distribution can generalize across context values and therefore the agent is able to adapt to unseen contexts.

Yet, direct policy search methods (both contextual and plain) usually require a lot of evaluations of the objective and may converge prematurely. 
To alleviate these issues, \citeauthor{DBLP:conf/nips/AbdolmalekiLPLR15} (\citeyear{DBLP:conf/nips/AbdolmalekiLPLR15}) recently proposed a stochastic search framework called \emph{model-based relative entropy stochastic search (MORE)}.
In this framework, the new search distribution can be computed efficiently in a closed form using a learned model of the objective function. 
MORE outperformed state-of-the-art methods in stochastic optimization problems and single-context policy search problems, but its application to contextual policy search has not been explored yet.
One of the contributions in this paper is a novel contextual policy search method in the MORE framework.



However, a naive extension of the original MORE would still suffer from high-dimensional contexts.
Learning from high-dimensional variables, in fact, is still an important problem in statistics and machine learning~\citep{Bishop:2006:PRM:1162264}.
Nowadays, high-dimensional data (e.g., camera images) can often be obtained quite easily, but obtaining informative low-dimensional variables (e.g., exact ball positions) is non-trivial and requires prior knowledge and/or human guidance.


In this paper, we propose to handle high-dimensional context variables by learning a low-rank representation of the objective function.
We show that learning a low-rank representation corresponds to performing linear dimensionality reduction on the context variables.
Since optimization with a rank constraint is generally NP-hard,
we minimize the \textit{nuclear norm} (also called {trace norm}), which is a \emph{convex} surrogate of the rank function \citep{DBLP:journals/siamrev/RechtFP10}.
This minimization allows us to learn a low-rank representation in a fully supervised manner by just solving a convex optimization problem.
We evaluate the proposed method on a synthetic task with known ground truth and on robotic ball hitting tasks based on camera images. The evaluation shows that the proposed method with nuclear norm minimization outperforms the methods that naively perform principal component analysis to reduce the dimensionality of context variables.

\section{Contextual Policy Search}
In this section, we formulate the direct contextual policy search problem
and briefly discuss existing methods.

\subsection{Problem Formulation}

The direct contextual policy search is formulated as follows.
An agent observes the context variable $\bc \in \mathbb{R}^{\cd}$
and  draws a parameter $\btheta \in \mathbb{R}^{\thetad}$ from a search distribution $\pi(\btheta|\bc)$.
Subsequently, the agent executes a policy with the parameter $\btheta$ and observes a scalar reward computed by a reward function $R(\btheta,\bc)$.
The goal is to find a search distribution $\pi(\btheta|\bc)$ maximizing the expected reward
\begin{align}
\iint \mu(\bc) \pi(\btheta|\bc) R(\btheta, \bc) \dtheta \dc, \label{eq:cps_obj}
\end{align}
where $\mu(\bc)$ denotes the context distribution.
We assume that the reward function $R(\btheta, \bc)$ itself is unknown, but the agent can always access the reward value.

\subsection{Related Work}

In the basic direct contextual policy search framework, the agent iteratively collects samples $\{( \btheta_n, \bc_n, R(\btheta_n, \bc_n) )\}_{n=1}^N$ using a sampling distribution $q(\btheta|\bc)$. Subsequently, it computes a new search distribution $\pi(\btheta|\bc)$ such that the expected reward increases or is maximized.
In literature, different approaches have been used to compute the new search distribution, e.g., 
evolutionary strategies~\citep{DBLP:journals/ec/HansenMK03},
expectation-maximization algorithms~\citep{DBLP:conf/ijcai/KoberOP11}, 
or information theoretic approaches~\citep{DBLP:journals/ftrob/DeisenrothNP13}. 


Most of the existing direct contextual policy search methods focus on tasks with low-dimensional context variables.
To learn from high-dimensional context variables, usually the problem of learning a low-dimensional context representation is separated from the direct policy search by preprocessing the context space.
However, unsupervised linear dimensionality reduction techniques are insufficient in problems where the latent representation contains distractor dimensions that do not influence the reward. A prominent example is principal component analysis (PCA)~\citep{PCA:Jolliffe:1986}, 
that does not take the supervisory signal into account and therefore cannot discriminate between relevant and irrelevant latent dimensions.
On the other hand, supervised linear dimensionality reduction techniques require a suitable response variable. 
However, defining such a variable can be subjective.
Moreover, they often involve non-convex optimization and suffer from local optima~\citep{fukumizu2009,Neco:SuzukiT:2013}. 

In the last years, non-linear dimensionality reduction techniques based on deep learning have gained popularity~\citep{DBLP:journals/ftml/Bengio09}.
For instance, \citeauthor{DBLP:conf/nips/WatterSBR15}
(\citeyear{DBLP:conf/nips/WatterSBR15})
proposed a generative deep network to learn low-dimensional representations of images in order to capture information about the system transition dynamics and allow optimal control problems to be solved in low-dimensional spaces.
More recently, \citeauthor{Nature44806}~(\citeyear{Nature44806}) successfully trained a machine to play a high-level game of \emph{go} using a deep convolutional network.
Although their work does not directly focus on dimensionality reduction, 
the deep convolutional network is known to be able to extract meaningful representation of data.
Thus, the effect of dimensionality reduction is achieved.

However, deep learning approaches generally require large datasets that are difficult to obtain in real-world scenarios (e.g., robotics). Furthermore, they involve solving non-convex optimization, which can suffer from local optima.

In this paper, we tackle the issues raised above. First, the proposed approach integrates supervised linear dimensionality reduction on the context variables by learning a \emph{low-rank representation} for the reward model. Second, the problem is formalized as a \emph{convex} optimization problem and is therefore guaranteed to converge to a global optimum.

\section{Contextual MORE}

The original MORE~\citep{DBLP:conf/nips/AbdolmalekiLPLR15} finds a search distribution (without context) that maximizes the expected reward while upper-bounding the Kullback-Leibler (KL) divergence~\citep{Kullback51klDivergence} and lower-bounding the entropy.
The KL and the entropy are bounded to control the exploration-exploitation trade-off.
The key insight of MORE is to learn a reward model to efficiently compute a new search distribution in closed form. 
Below, we propose our method called \textit{contextual model-based relative entropy stochastic search} (C-MORE), which is a direct contextual policy search method in the MORE framework.

\subsection{Learning the Search Distribution}

The goal of C-MORE is to find a search distribution $\pi(\btheta|\bc)$ that maximizes the expected reward while upper-bounding the expected KL divergence between $\pi(\btheta|\bc)$ and $q(\btheta|\bc)$,
and lower-bounding the expected entropy of $\pi(\btheta|\bc)$. Formally,
\begin{align*}
\max_{\pi} &\iint \mu(\bc) \pi(\btheta|\bc) R(\btheta,\bc) \dtheta \dc, \\
\mathrm{s.t.}
	&\iint \mu(\bc) \pi(\btheta|\bc) \log \frac{\pi(\btheta|\bc)}{q(\btheta|\bc)} \dtheta \dc \leq \epsilon, \\
	-&\iint \mu(\bc) \pi(\btheta|\bc) \log \pi(\btheta|\bc) \dtheta\dc \geq \beta, \\
	&\iint \mu(\bc) \pi(\btheta|\bc) \dtheta \dc = 1,
\end{align*}
where the KL upper-bound $\epsilon$ and the entropy lower-bound $\beta$ are parameters specified by the user.
The former is fixed for the whole learning process. The latter is adaptively changed according to the percentage of the relative difference between the sampling policy's expected entropy and the minimal entropy, as described by~\citeauthor{DBLP:conf/nips/AbdolmalekiLPLR15}~(\citeyear{DBLP:conf/nips/AbdolmalekiLPLR15}), i.e., 
\begin{align*}
\beta = \gamma( \mathbb{E}[H(q)] - H_0 ) + H_0,
\end{align*}
where $\mathbb{E}[H(q)] = -\iint \mu(\bc) q(\btheta|\bc) \log q(\btheta|\bc) \dtheta\dc$ is the sampling policy's expected entropy and $H_0$ is the minimal entropy. 
In the experiments, we set $\gamma = 0.99$ and $H_0 = -150$. 
The above optimization problem can be solved by the method of Lagrange multipliers\footnote{All derivations are given in the supplementary material.
}.
The solution is given by
\begin{align*}
\pi(\btheta|\bc) &= q(\btheta|\bc)^{\frac{\eta}{\eta+\omega}}
	 \exp\left( \frac{R(\btheta, \bc)}{\eta+\omega} \right) 
	 \exp	\left( -\frac{\eta+\omega-\gamma}{\eta+\omega} \right), 
\end{align*}
where $\eta > 0$ and $\omega > 0$ are the Lagrange multipliers obtained by minimizing the dual function
\begin{align}
g(\eta, \omega) &= 
\eta\epsilon - \omega\beta+ (\eta+\omega ) \times \int \mu(\bc) \notag \\
&\phantom{=} \log \left( \int q(\btheta|\bc)^{\frac{\eta}{\eta+\omega}}
	\exp\left( \frac{R(\btheta, \bc)}{\eta+\omega} \right) \dtheta \right) \dc.
	\label{dual_int}
\end{align}
Evaluating the above integral is not trivial due to the integration over $q(\btheta|\bc)^{\frac{\eta}{\eta+\omega}}$, that cannot be approximated straightforwardly by sample averages.
Below, we describe how to solve this issue and evaluate the dual function from data.

\subsection{Dual Function Evaluation via the Quadratic Model}

\begin{algorithm}[t]
\caption{C-MORE}
\label{algo_CMORE}
\DontPrintSemicolon
\KwIn{Parameters $\epsilon$ and $\beta$, initial distribution $\pi(\btheta|\bc)$}

\For{$k=1,\dots,K$}
	{
	\For{$n=1,\dots,N$}
		{
			Observe context $\bc_n \sim \mu(\bc)$ \;
			Draw parameter $\btheta_n \sim \pi(\btheta|\bc_n)$ \;
			Execute task with $\btheta_n$ and receive $R(\btheta_n, \bc_n)$ \;
		}
		
	Learn the quadratic model $\widehat{R}(\btheta, \bc)$ \;
	
	Solve $\argmin_{\eta>0, \omega>0} g(\eta, \omega)$ using Eq.~\eqref{dual} \;

	Set new search distribution $\pi(\btheta|\bc)$ using Eq.~\eqref{policy_update}	
	}
	
\end{algorithm}

We assume that the reward function 
$R(\btheta, \bc)$ can be approximated by a quadratic model
\begin{align}
\widehat{R}(\btheta,\bc) &= \btheta^\top \bA \btheta 
	+ \bc^\top \bB \bc 
	+ 2 \btheta^\top \bD \bc \notag \\
&\phantom{=}	+ \btheta^\top \br_1 + \bc^\top \br_2 + r_0,
	\label{quadratic_model}
\end{align}
where $\bA \in \mathbb{R}^{\thetad  \times \thetad}, 
\bB \in \mathbb{R}^{\cd \times \cd}, 
\bD \in \mathbb{R}^{\thetad \times \cd}, 
\br_1 \in \mathbb{R}^{\thetad}, 
\br_2 \in \mathbb{R}^{\cd}$, 
and $r_0 \in \mathbb{R}$ are the model parameters. Matrices $\bA$ and $\bB$ are symmetric.
We also assume the sampling distribution $q(\btheta|\bc)$ to be Gaussian of the form
\begin{align}
q(\btheta|\bc) &= \mathcal{N}(\btheta|\bb + \bK\bc, \bQ). \label{gaussianpolicy}
\end{align}
Under these assumptions, the dual function in Eq.~\eqref{dual_int} can be expressed by
\begin{align}
g(\eta, \omega) 
&= \eta\epsilon - \omega\beta 
   + \frac{1}{2}\bigg(
   \bff^\top\bF^{-1}\bff -\eta \bb^\top\bQ^{-1}\bb \notag \\
&\phantom{=}   +  (\eta+\omega) \log | 2\pi \bF^{-1}(\eta+\omega) |
   - \eta \log | 2\pi\bQ |  \bigg) \notag \\
&\phantom{=} 
	+ \int \mu(\bc) \left( \bc^\top \bm + \frac{1}{2}\bc^\top \bM \bc \right) \dc,
	\label{dual}
\end{align}
where
\begin{align*}
\bff &= \eta \bQ^{-1}\bb + \br_1, \\
\bF &= \eta \bQ^{-1} - 2 \bA, \\
\bm &= \bL^\top \bF^{-1} \bff - \eta \bK^\top \bQ^{-1} \bb, \\
\bM &= \bL^\top \bF^{-1} \bL - \eta \bK^\top \bQ^{-1}\bK, \\
\bL &= \eta \bQ^{-1}\bK + 2\bD.
\end{align*}
Since the context distribution $\mu(\bc)$ is unknown, we approximate the expectation in Eq.~\eqref{dual} by sample averages.
The dual function can be minimized by standard non-linear optimization routines such as IPOPT~\citep{Waechter2006}.
Finally, using Eq.~\eqref{quadratic_model} and Eq.~\eqref{gaussianpolicy} the new search distribution $\pi(\btheta|\bc)$ is computed in closed form as
\begin{align}
\pi(\btheta|\bc) &=
\mathcal{N}\Big(\btheta | \bF^{-1}\bff + \bF^{-1}\bL\bc, \bF^{-1}(\eta+\omega)\Big).
\label{policy_update}
\end{align} 
To ensure that the covariance $\bF^{-1}(\eta + \omega)$ is positive definite,
the model parameter $\bA$ is constrained to be negative definite.
C-MORE is summarized in Algorithm~\ref{algo_CMORE}.

\section{Learning the Quadratic Model}
\label{section:model_learning}
The performance of C-MORE depends on the accuracy of the quadratic model.
For many problems, the reward function $R(\btheta,\bc)$ is not quadratic 
and the quadratic model is not suitable to approximate the entire reward function.
However, the reward function is often smooth and it can be \emph{locally} approximated by a quadratic model.
Therefore, we locally approximate the reward function by learning a new quadratic model for each policy update.
The quadratic model can be learned by regression methods such as ridge regression\footnote{After learning the parameters, $\bA$ is enforced to be negative definite by truncating its positive eigenvalues. Subsequently, we re-learn the remainder parameters.
An alternative approach is projected gradient descend, but it is more computationally demanding and requires step size tuning.}%
~\citep{Bishop:2006:PRM:1162264}.
However, ridge regression is prone to error when the context is high-dimensional.
Below, we address this issue by firstly showing that performing linear dimensionality reduction on the context variables yields a low-rank matrix of parameters. Secondly, we propose a nuclear norm minimization approach to learn a low-rank matrix without explicitly performing dimensionality reduction.

\subsection{Dimensionality Reduction and Low-Rank Representation}
Linear dimensionality reduction learns a low-rank matrix $\bW$ and projects the data onto a lower dimensional subspace.
Performing linear dimensionality reduction on the context variables yields the following quadratic model
\begin{align}
\widehat{R}(\btheta, \bc) 
&= \btheta^\top \bA \btheta 
	+ \bc^\top \bW^\top \widetilde{\bB} \bW \bc 
	+ 2 \btheta^\top \widetilde{\bD} \bW \bc \notag \\
&\phantom{=}  
	+ \btheta^\top \br_{1} + \bc^\top \bW^\top \widetilde{\br}_{2} + r_0, \label{eq:model_DR}
\end{align}
where $\bW \in \mathbb{R}^{\zd \times \cd}$ denotes a rank-$\zd$ matrix with $\zd < \cd$.
The model parameters $\bA, \widetilde{\bB}, \widetilde{\bD}, \br_1, \widetilde{\br}_2$ and $r_0$ can be learned by ridge regression.
However, the matrix $\bB = \bW^\top \widetilde{\bB} \bW$ is low-rank, i.e., $\mathrm{rank}(\bB) = \zd < \cd$.
Thus, performing linear dimensionality reduction on the contexts 
makes $\bB$ low-rank.
Note that the rank of $\bD = \widetilde{\bD}\bW$ depends on $\btheta$ and is problem dependent.
Hence, we do not consider the rank of $\bD$ for dimensionality reduction.

There are several linear dimensionality reduction methods that can be applied to learn $\bW$.
{Principal component analysis (PCA)}~\citep{PCA:Jolliffe:1986} is a common method used in statistics and machine learning.
However, being unsupervised, it does not take the regression targets into account, i.e., the reward.
Alternative supervised techniques, such as KDR \citep{fukumizu2009} and LSDR \citep{Neco:SuzukiT:2013}, do not take the regression model, i.e., the quadratic model, into account.
%
On the contrary, in {projection regression}~\citep{Friedman1981,DBLP:conf/icml/VijayakumarS00} the model parameters and the projection matrix are learned simultaneously.
However, applying this approach to the model in Eq.~\eqref{eq:model_DR} requires alternately optimizing for the model parameters and the projection matrix and is computationally expensive.

In the original MORE, Bayesian dimensionality reduction~\citep{DBLP:journals/tcyb/Gonen13} is applied to perform linear supervised dimensionality reduction on $\btheta$, i.e., the algorithm considers a projection $\bW\btheta$.
The matrix $\bW$ is sampled from a prior distribution and the algorithm learns the model parameters using weighted average over the sampled $\bW$.
However, for high-dimensional $\bW$, this approach requires an impractically large amount of samples $\bW$ to obtain an accurate model, leading to computationally expensive updates.

\subsection{Learning a Low-Rank Matrix with Nuclear Norm Regularization}

The quadratic model in Eq.~\eqref{quadratic_model} 
can be re-written as 
\begin{align*}
\widehat{R}(\bx) = \bx^\top \bH \bx,
\end{align*}
where the input vector $\bx$ and the parameter matrix $\bH$ are defined as
\begin{align*}
\bx &= 
\left[
\begin{array}{c}
\btheta\\ 
\bc \\ 
1 
\end{array}
\right], \quad
\bH &= 
\begin{bmatrix}
\bA 	 & \bD 	& 0.5\br_1 \\
\bD^\top 	 & \bB 	& 0.5\br_2 \\
0.5\br_1^\top	 & 0.5\br_2^\top & r_0
\end{bmatrix}.
\end{align*}
Note that $\bH$ is symmetric since both $\bA$ and $\bB$ are symmetric.
As discussed in the previous section, we desire $\bB$ to be low-rank. 
Unlike Eq.~\eqref{eq:model_DR}, we do not consider dimensionality reduction for the linear terms in $\bc$, i.e., $2\btheta^\top \bD \bc$ and $\bc^\top \br_2$.
Instead, we learn $\bH$ by solving the following convex optimization problem
\begin{align}
\min_{\bH} &\left[ \mathcal{J}(\bH) + \lambda_{*}\|\bB\|_{*} \right],
\notag \\
\text{s.t. } &\bA \text{ is negative definite}, \label{eq:low_rank_opt}
\end{align}
where $\mathcal{J}(\bH)$ denotes the differentiable part
\begin{align*}
\mathcal{J}(\bH) = \frac{1}{2N}\sum_{n=1}^N \left( \bx_n^\top \bH \bx_n - R(\btheta_n,\bc_n) \right)^2 + \frac{\lambda}{2} \|\bH\|_{\mathrm{F}}^2,
\end{align*}
where $\lambda>0$ and $\lambda_* >0$ are regularization parameters.
The Frobenius norm $\|\cdot\|_{\mathrm{F}}$ is defined as 
$\| \bH \|_{\mathrm{F}} = \sqrt{\mathrm{tr}(\bH\bH^\top)}$.
The nuclear norm of a matrix $\|\cdot\|_{*}$ is defined as the $\ell_1$-norm of its singular values
.
This optimization problem can be explained as follows.
The term $\mathcal{J}(\bH)$ consists of the mean squared error and the $\ell_2$-regularization term.
Thus, minimizing $\mathcal{J}(\bH)$ corresponds to ridge regression.
Minimizing the nuclear norm $\|\bB\|_{*}$ shrinks the singular values of $\bB$. Thus, the solution tends to have sparse singular values and to be low-rank.
The negative definite constraint further ensures that the covariance matrix in Eq.~\eqref{policy_update} is positive definite.

The {convexity} of this optimization problem can be verified by checking the following conditions. 
First, the convexity of the mean squared error can be proven following
\citeauthor{Covex:Boyd:2004}, \citeyear[page 74]{Covex:Boyd:2004}.
Let $g(t) = \widehat{\mathcal{J}}(\bZ + t\bV)$ be the mean squared error and 
$\bZ$ and $\bV$ are symmetric matrices.
Then we have that $\nabla^2 g(t) = \frac{1}{N}\sum(\bx_n^\top \bV \bx_n)^2 \geq 0$.
Thus, the mean squared error is convex.
Since the Frobenius norm is convex, $\mathcal{J}(\bH)$ is convex as well.
Second, a set of negative definite matrices is convex since $ \by^\top ( a\bX + (1-a)\bY ) \by < 0 $ for any negative definite matrices $\bX$ and $\bY$, $0 \leq a \leq 1$, and any vector $\by$ \citep{Covex:Boyd:2004}. Third, the nuclear norm is a convex function \citep{DBLP:journals/siamrev/RechtFP10}.
Note that, since the gradient $\nabla \mathcal{J}(\bH)$ is symmetric, $\bH$ is guaranteed to be symmetric as well given that the initial solution is also symmetric.

It is also possible to enforce the matrix $\bH$ (rather than $\bB$) to be low-rank, implying that both $\btheta$ and $\bc$ can be projected onto a common low-dimensional subspace.
However, this is often not the case, and regularizing by the nuclear norm of $\bH$ did not perform well in our experiments.
We may also directly constrain $\mathrm{rank}(\bB) = \zd$ in Eq.~\eqref{eq:low_rank_opt} instead of performing nuclear norm regularization.
However, minimization problems with rank constraints are NP-hard.
On the contrary, the nuclear norm is the convex envelop of the rank function
and can be optimized more efficiently~\citep{DBLP:journals/siamrev/RechtFP10}.
For this reason, the nuclear norm has been a popular surrogate to a low-rank constraint in many applications, such as matrix completion \citep{DBLP:journals/tit/CandesT10} and multi-task learning \citep{DBLP:journals/siamjo/PongTJY10}.

Since the optimization problem in Eq.~\eqref{eq:low_rank_opt} is convex, any convex optimization method can be used \citep{Covex:Boyd:2004}.
For our experiments, we use the \emph{accelerated proximal gradient descend (APG)}~\citep{Toh09anaccelerated}.
The pseudocode of our implementation of APG for solving Eq.~\eqref{eq:low_rank_opt} is given in the supplementary material.
Note that APG requires computing the SVD of the matrix $\bB$. Since computing the exact SVD of a high-dimensional matrix can be computationally expensive, we approximate it by randomized SVD \citep{DBLP:journals/siamrev/HalkoMT11}.


\begin{figure*}[t]
\centering
\begin{minipage}[t]{.47\textwidth}
	\centering
	\input{fig_quadexp2}
	\caption{\label{fig:quad_reward} Average reward for the quadratic cost function problem. Shaded area denotes standard deviation (results are averaged over ten trials). Only C-MORE Nuc. Norm converges within 100 iterations to an almost optimal policy.}
\end{minipage}\hfill
\begin{minipage}[t]{.49\textwidth}
    \centering
	\input{fig_2dof_reward}
	\caption{\label{fig:2dof_reward}Averaged reward for the 2-DoF hitting task. C-REPS outperforms C-MORE early on. However, it prematurely converges to suboptimal solutions, while C-MORE continues to improve and soon outperforms C-REPS.}
\end{minipage}
\end{figure*}


\section{Experiments}
We evaluate the proposed method on three problems. We start by studying C-MORE behavior in a scenario where we know the true reward model and the true low-dimensional context. 
Subsequently, we focus our attention on two simulated robotic ball hitting tasks.
In the first task, a toy 2-DoF planar robot arm has to hit a ball placed on a plane. 
In the second task, a simulated 6-DoF robot arm has to hit a ball placed in a three-dimensional space. 
In both cases, the robots accomplish their task by using raw camera images as context variables.
However, in the latter case we have limited data and therefore sample efficiency is of primary importance.

The evaluation is performed on three different versions of C-MORE, according to the model learning approach: using only ridge regression (\emph{C-MORE Ridge}), aided by a low-dimensional context variables learned by PCA (\emph{C-MORE Ridge+PCA}) and aided by nuclear norm regularization (\emph{C-MORE Nuc. Norm}). We also use \emph{C-REPS} \citep{DBLP:journals/ftrob/DeisenrothNP13} with PCA as baseline.
For the ball hitting tasks, we also tried to preprocess the context space with an autoencoder. However, the learned representation performed poorly, possibly due to the limited amount of data at our disposal, and therefore this method is not reported.

For each case study, first, the experiments are presented and then the results are reported and discussed. For additional details, we refer to the supplementary material.

\subsection{Quadratic Cost Function Optimization}
In the first experiment, we want to study the performance of the algorithms in a setup where we are able to analytically compute both the reward and the true low-dimensional context. To this aim, we define the following problem
\begin{gather*}
R(\btheta, \bc) = -(||\btheta - \bT_1\tilde{\bc}||_2)^2,
\quad
\tilde{\bc} = \widetilde{\bI}  \bT_2^{-1}  \bc,
\\
\bT_1 \in \mathbb{R}^{\thetad\times\trued}, 
\enspace \bT_2 \in \mathbb{R}^{\cd\times\cd}, 
\enspace \widetilde{\bI} \in \mathbb{R}^{\trued\times\cd}, 
\enspace \trued < \cd,
\end{gather*}
where $\bI$ is the identity matrix, $\widetilde{\bI}$ is a rectangular matrix with ones in its main diagonal and zeros otherwise, $\tilde{\bc}$ is the true low-dimensional context, and $\bT_1$ is to match the dimension of the true context and the parameter $\btheta $ in order to compute the reward.
This setup is particularly interesting because only a subset of the observed context influences the reward. First, the observed context $\bc$ is linearly transformed by $\bT_2^{-1}$. Subsequently, thanks to the matrix $\widetilde{\bI}$, only the first $\trued$ elements are kept to compose the true context, while the remainder is treated as noise. Finally, the reward is computed by linearly transforming the true context by $\bT_1$.

\paragraph{Setup.}
We set $\trued = 3, \thetad = 10, \cd = 25$, while the elements of $\bT_1, \bT_2$ are chosen uniformly randomly in $[0,1]$.
The sampling Gaussian distribution is initialized with random mean and covariance $\bQ = 10,000 \bI$. 
For learning, we collect 35 new samples and keeps track of the samples collected during the last 20 iterations to stabilize the policy update. 
The evaluation is performed at each iteration over 1,000 contexts. Each context element is drawn from a uniform random distribution in $[-10,10]$.
Since we can generate a large amount of data in this setting, 
we pre-train PCA using 10,000 random context samples and fixed the dimensionality to $\zd = 20$ (chosen by cross-validation).
The learning is performed for a maximum of 100 iterations. If the KL divergence is lower than 0.1, then the learning is considered to be converged and the policy is not updated anymore.

\paragraph{Results.}
As shown in Figure~\ref{fig:quad_reward} 
, C-MORE Nuc. Norm 
clearly outperforms all the competitors, learning an almost optimal policy and being the only one to converge within the maximum number of iterations. It is also the only algorithm correctly learning the true context dimensionality, as nuclear norm successfully regularizes $\bB$ to have rank three.
On the contrary, PCA does not help C-MORE much and yields only slightly better results than plain ridge regression. PCA cannot in fact determine task-relevant dimensions as non-relevant dimensions have equally-high variance.

\begin{figure*}[t]
\centering
\begin{minipage}[t]{.53\textwidth}
	\centering
	\captionsetup[subfigure]{aboveskip=-10pt}
	\begin{subfigure}[t]{.49\linewidth}
        \centering
		\includegraphics[trim={0cm -3.1cm 0cm 0cm},clip,width=.92\linewidth]{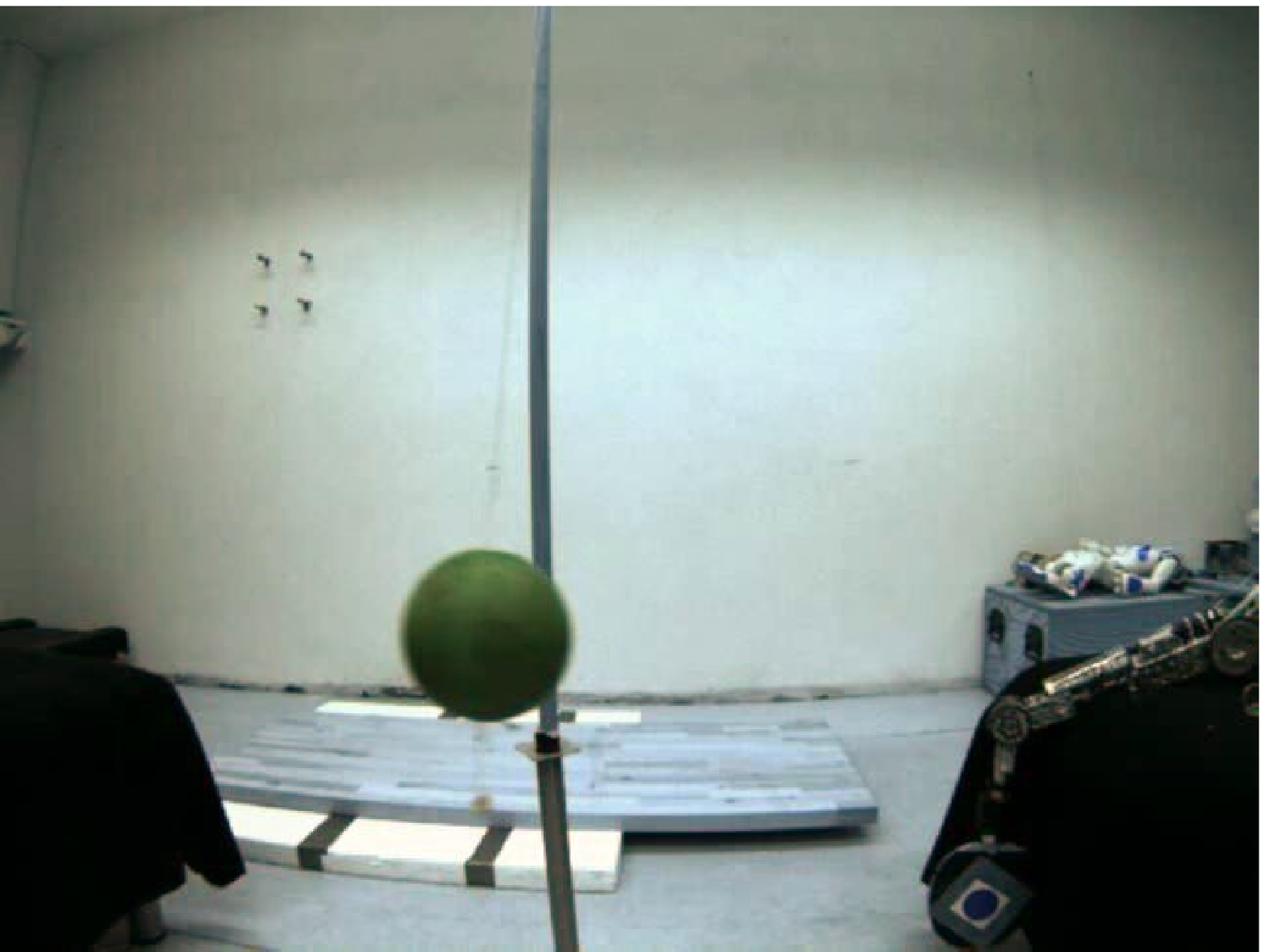}
		\caption{\label{fig:6dof_1}}
	\end{subfigure}
	\hfill
	\begin{subfigure}[t]{.49\linewidth}
		\centering
		\includegraphics[trim={0cm -2.3cm 0cm 0cm},clip,width=\linewidth]{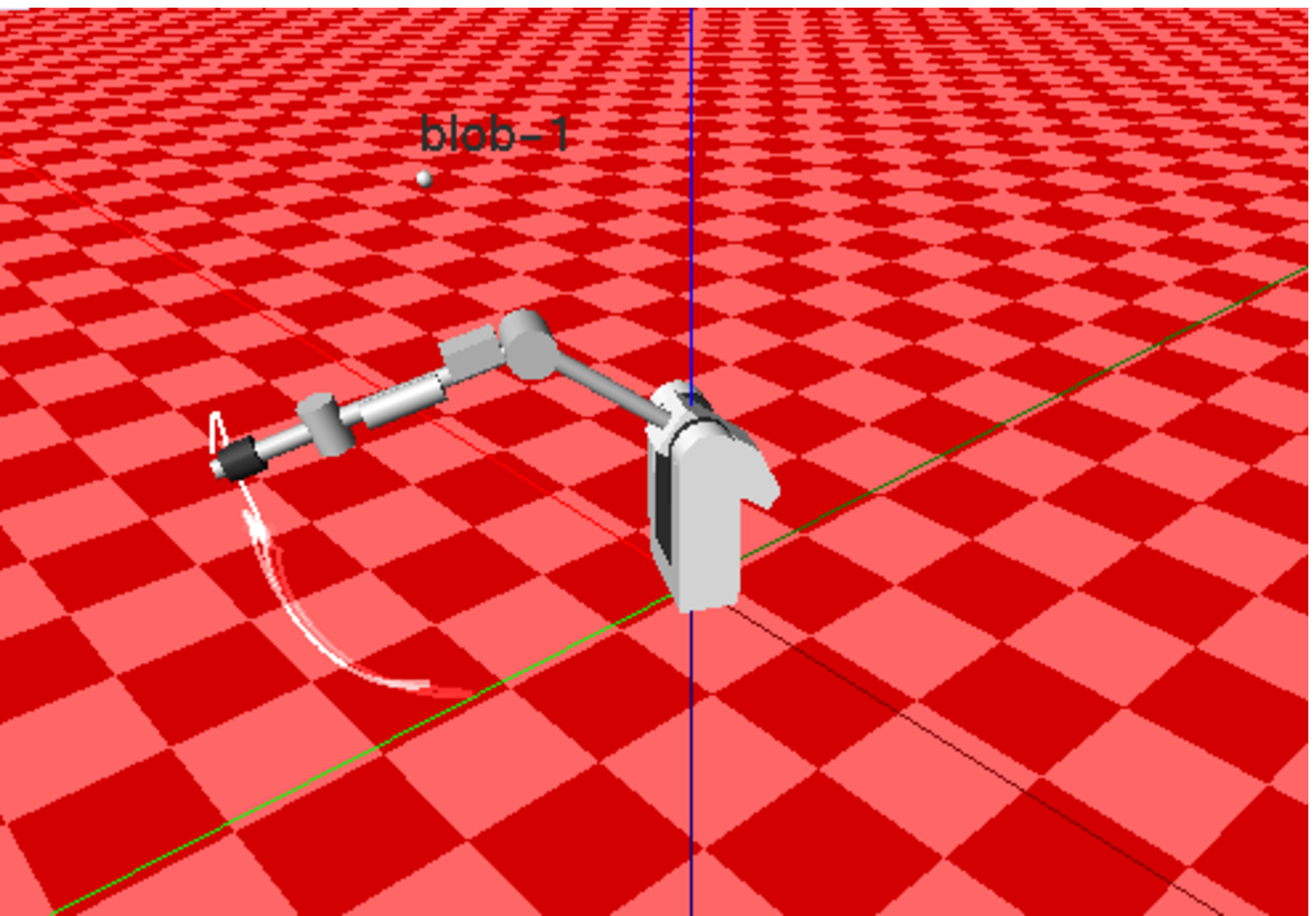}
		\caption{\label{fig:6dof_2}}
	\end{subfigure}
	\caption{\label{fig:6dof_robot} The 6-DoF robot as seen from the camera (Figure~\ref{fig:6dof_1}, bottom right) and in simulation (Figure~\ref{fig:6dof_2}). The goal is to control the robot to hit the real green ball according to camera images, resized to $32 \times 24$.}
\end{minipage}\hfill
\begin{minipage}[t]{.44\textwidth}
    \centering
	\input{fig_6dof_reward}
	\caption{\label{fig:6dof_reward}6-DoF hitting task results (averaged over three trials). 
	Nuclear norm regularization outperforms PCA, both in terms of reward and accuracy.}
\end{minipage}
\vspace{-0.1cm}
\end{figure*}

{
\subsection{Ball Hitting with a 2-DoF Robot Arm}
\setlength{\columnsep}{12pt}%
\begin{wrapfigure}[14]{l}{3.7cm}
\vspace{-.4cm}
\centering
\includegraphics[width=\linewidth]{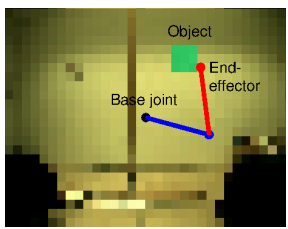}
\caption{\label{fig:2dof_robot} 2-DoF hitting task. The context observed by the robot (blue and red lines) consists of a virtual green ball and the background image.}
\end{wrapfigure}
In this task, a simulated planar robot arm (shown aside) has to hit a green virtual ball placed on RGB camera images of size $32 \times 24$. The context is defined by the observed pixels, for a total of 2304 context variables.
The ball is randomly and uniformly placed in the robot workspace.
Noise drawn from a uniform random distribution in $[-30,30]$ is also added to the context, to simulate different light conditions.
The robot controls the joint accelerations at each time step by a linear-in-parameter controller with Gaussian basis functions, for a total of 32 parameters $\btheta$ to be learned.
The reward ${R}(\btheta, \bc)$ is the negative cumulative joint accelerations plus the negative distance between the end-effector and the ball at the final time step. 

\paragraph{Setup.}
For learning, the agent collects $50$ samples at each iteration and keeps samples from the last four previous iterations.
The evaluation is performed at each iteration over 500 contexts.
Pixel values are normalized in $\left[ -1, 1 \right]$.
The sampling Gaussian distribution is initialized with random mean and identity covariance.
For both C-MORE Nuc. Norm and C-MORE PCA, we perform 5-fold cross-validation every $100$ policy updates to choose the values of $\lambda_{*}$ and $\zd$, respectively, based on the mean squared error between the collected returns and the model-predicted ones.
For C-REPS PCA, we tried different values of $\zd \in \{10, 20, 30, 40 \}$ and selected $\zd = 10$ which gave the best result.

\paragraph{Results.}
Figure~\ref{fig:2dof_reward} shows the averaged reward against the number of iterations. 
Once again, C-MORE aided by nuclear norm regularization performs the best, achieving the highest average reward. 
At the $1000$th iteration, 
the learned controller hits the ball with $76\%$ accuracy.
The rank of its learned matrix $\bB$ is approximately $31$, which shows that the algorithm successfully learns a low-rank model representation.
On the contrary, preprocessing the context space through PCA still helps C-MORE (the rank of its learned $\bB$ is approximately 25), but yields poor results for C-REPS, which suffers of premature convergence. 
}

\subsection{Ball Hitting with a 6-DoF Robot Arm}
Similarly to the previous task, here a 6-DoF robotic arm has to hit a ball placed on a three-dimensional space, as shown in Figure~\ref{fig:6dof_robot}.
The context is once again defined by the vectorized pixels of RGB images of size $32 \times 24$, for a total of $2304$ context variables. 
Note that Figure~\ref{fig:6dof_1} shows an image before we rescale it to size $32 \times 24$.
However, unlike the 2-DoF task, the ball is directly recorded by a real camera placed near the physical robot, and it is not virtually generated on the images.
Furthermore, the robot is controlled by dynamic motor primitives~\citep{ijspeert2002learning} (DMPs), which are non-linear dynamical systems. We use one DMP per joint, with five basis functions per DMP. We also learn the goal attractor of the DMPs, for a total of 36 parameters $\btheta$ to be learned.
The reward ${R}(\btheta, \bc)$ is computed as the negative cumulative joint accelerations and minimum distance between the end-effector and the ball as well.
\paragraph{Setup.}
The image dataset is collected by taking pictures with the ball placed at $50$ different positions. 
To increase the number of data, we add a uniform random noise in $[-30,30]$ to the context to simulate different light conditions.
Therefore, although some samples determine the same ball position, they are considered different due to the added noise.
The search distribution is initialized by imitation learning using $50$ demonstration samples.
For learning, the agent collects $50$ samples at each iteration and always keeps samples from the last four previous iterations.
\paragraph{Results.}
We only evaluate C-MORE with {nuclear norm} and {PCA} since they performed well in the previous evaluation.
Figure~\ref{fig:6dof_reward} shows that nuclear norm again outperforms PCA.
At the $500$th iteration, 
the learned controller hits the ball with $80\%$ accuracy. 
Considering that the robot is not able to hit the ball in some contexts and can achieve a maximum accuracy of $90\%$, this accuracy is impressive for the task.
The averaged rank of matrix $\bB$ learned by the nuclear norm approach is approximately $25$, which shows that minimizing the nuclear norm successfully learns a low-rank matrix. For PCA, the averaged rank of $\bB$ is approximately 30.

\section{Conclusion}

Learning with high-dimensional context variables is a challenging and prominent problem in machine learning.
In this paper, we proposed C-MORE, a novel contextual policy search method with integrated dimensionality reduction.
C-MORE learns a reward model that is locally quadratic in the policy parameters and the context variables.
By enforcing the model representation to be low-rank, we perform supervised linear dimensionality reduction.
Unlike existing techniques relying on non-convex formulations, the nuclear norm allows us to learn the low-rank representation by solving a \emph{convex} optimization problem, thus guaranteeing convergence to a global optimum.
The main disadvantage of the proposed method is that it demands more computation time due to the nuclear norm regularization.
Although we did not encounter severe problems in our experiments, for very large dimensional tasks this issue can be mitigated by using more efficient techniques, such as active subspace selection~\citep{DBLP:conf/icml/HsiehO14}.

In this paper, we only focused on linear dimensionality reduction techniques.
Recently, non-linear techniques based on deep network has been showing impressive performance~\citep{DBLP:journals/ftml/Bengio09,DBLP:conf/nips/WatterSBR15}.
In future work, we will incorporate deep network into C-MORE, e.g., by using a deep convolutional network to represent the reward model.





\bibliography{references}

\onecolumn

\Huge
\centerline{
---Supplementary Material---
}

\large


\section{Derivations of C-MORE}
In this section, we derive C-MORE in details.
C-MORE solves 
\begin{align*}
\max_{\pi}&~ \iint \mu(\bc) \pi(\btheta|\bc) R(\btheta,\bc) \dtheta \dc, \\
\mathrm{s.t.}
	&~ \iint \mu(\bc) \pi(\btheta|\bc) \log \frac{\pi(\btheta|\bc)}{q(\btheta|\bc)} \dtheta \dc \leq \epsilon, \\
	&~ -\iint \mu(\bc) \pi(\btheta|\bc) \log \pi(\btheta|\bc) \dtheta\dc \geq \beta, \\
	&~ \iint \mu(\bc) \pi(\btheta|\bc) \dtheta \dc = 1,
\end{align*}
by the method of Lagrange multipliers.
Firstly, we write the Lagrangian $\mathcal{L}$ with the Lagrange multipliers $\eta > 0, \omega > 0$, and $\gamma$, which correspond to the first, second, and third constraints, respectively
\begin{align*}
\mathcal{L}(\pi, \eta, \omega, \gamma) 
&= \iint \mu(\bc)\pi(\btheta|\bc) R(\btheta,\bc) \dtheta \dc
 + \eta \left( \epsilon - \iint \mu(\bc) \pi(\btheta|\bc)\log \frac{\pi(\btheta|\bc)}{q(\btheta|\bc)} \dtheta \dc \right) \\
&\phantom{=} + \omega \left( -\iint \mu(\bc) \pi(\btheta|\bc) \log \pi(\btheta|\bc) \dtheta\dc - \beta \right) 
 + \gamma \left( \iint \mu(\bc) \pi(\btheta|\bc) \dtheta \dc - 1 \right).
\end{align*}
Then, we maximize the Lagrangian $\mathcal{L}(\pi, \eta, \omega, \gamma)$ w.r.t. the primal variable $\pi$.
The derivative of the Lagrangian w.r.t. $\pi$ is
\begin{align*}
\partial_{\pi}{\mathcal{L}(\pi, \eta, \omega, \gamma)} 
&= \iint \mu(\bc) \Big( R(\btheta, \bc) 
	-(\eta+\omega) \log \pi(\btheta|\bc) + \eta \log q(\btheta|\bc) \Big) \dtheta \dc - (\eta + \omega - \gamma). 
\end{align*}
By setting this derivative to zero, we have
\begin{align*}
0 &= \iint \mu(\bc) \Big( R(\btheta, \bc) 
	-(\eta+\omega) \log \pi(\btheta|\bc) + \eta \log q(\btheta|\bc) \Big) \dtheta \dc - (\eta + \omega -\gamma) \\
  &= R(\btheta, \bc) 
	-(\eta+\omega) \log \pi(\btheta|\bc) + \eta \log q(\btheta|\bc) - (\eta + \omega - \gamma).
\end{align*}
This gives us
\begin{align}
\log \pi(\btheta|\bc) &= \frac{R(\btheta, \bc)}{\eta+\omega} + \frac{\eta}{\eta+\omega} \log q(\btheta|\bc) - \frac{\eta+\omega-\gamma}{\eta+\omega}, \notag \\
\pi(\btheta|\bc) &= q(\btheta|\bc)^{\frac{\eta}{\eta+\omega}}
	 \exp\left( \frac{R(\btheta, \bc)}{\eta+\omega} \right) 
	 \exp	\left( -\frac{\eta+\omega-\gamma}{\eta+\omega} \right). \label{eq:weight_policy_normalize}
\end{align}
The last exponential term in Eq.~\eqref{eq:weight_policy_normalize} is the normalization constant for the search distribution $\pi(\btheta|\bc)$ since it does not depend on $\btheta$ or $\bc$.
Thus, we have 
\begin{align*}
\exp \left( \frac{\eta+\omega-\gamma}{\eta+\omega} \right) 
&= \int q(\btheta|\bc)^{\frac{\eta}{\eta+\omega}}
	 \exp\left( \frac{R(\btheta, \bc)}{\eta+\omega} \right) \dtheta, \\
\eta+\omega-\gamma 
&= (\eta+\omega) \log \left( \int q(\btheta|\bc)^{\frac{\eta}{\eta+\omega}}
	 \exp\left( \frac{R(\btheta, \bc)}{\eta+\omega} \right) \dtheta \right).
\end{align*}
(The minus sign in the exponent in Eq.~\eqref{eq:weight_policy_normalize} becomes the inverse operator and cancels out).
This normalization term will be used to derive the dual function.
Next, we substitute the term $\log \pi(\btheta|\bc)$ back to the Lagrangian
\begin{align*}
\mathcal{L}(\pi^*, \eta, \omega, \gamma) 
&= \iint \mu(\bc)\pi(\btheta|\bc) R(\btheta,\bc) \dtheta \dc \\
&\phantom{=} - \eta \left(\iint \mu(\bc) \pi(\btheta|\bc) 
	\left[ \frac{R(\btheta, \bc)}{\eta+\omega} + \frac{\eta}{\eta+\omega} \log q(\btheta|\bc) - \frac{\eta+\omega-\gamma}{\eta+\omega} \right]
 \dtheta \dc \right) \\
&\phantom{=} -\omega\left(\iint \mu(\bc)\pi(\btheta|\bc) 
	\left[ \frac{R(\btheta, \bc)}{\eta+\omega} + \frac{\eta}{\eta+\omega} \log q(\btheta|\bc) - \frac{\eta+\omega-\gamma}{\eta+\omega} \right]
 \dtheta \dc \right) \\
&\phantom{=} + \eta \iint \mu(\bc) \pi(\btheta|\bc) \log q(\btheta|\bc) \dtheta \dc
+ \gamma\left( \iint \mu(\bc) \pi(\btheta|\bc) \dtheta \dc - 1 \right) + \eta\epsilon -\omega\beta.
\end{align*}
Most terms cancel out and we only have
\begin{align*}
\mathcal{L}(\pi^*, \eta, \omega, \gamma) 
	&= \eta\epsilon - \omega\beta -\gamma + \int \mu(\bc) \left( \eta + \omega \right) \dc \\
	&= \eta\epsilon - \omega\beta + \int \mu(\bc) \left( \eta + \omega -\gamma \right) \dc \\
	&= \eta\epsilon - \omega\beta + 
	\left(\eta+\omega\right) \int \mu(\bc) \log \left( \int q(\btheta|\bc)^{\frac{\eta}{\eta+\omega}}
	 \exp\left( \frac{R(\btheta, \bc)}{\eta+\omega} \right) \dtheta \right) \dc \\
	&= g(\eta, \omega).
\end{align*}
The Lagrange multipliers $\eta > 0$ and $\omega > 0$ are obtained by minimizing the dual function $g(\eta, \omega)$.
Then, the search distribution in Eq.~\eqref{eq:weight_policy_normalize} can be computed using these Lagrange multipliers.


\section{Evaluating the Dual Function}
Here, we show how to compute the new search distribution in closed form.
Recall that our quadratic model is
\begin{align*}
\widehat{R}(\btheta,\bc) = \btheta^\top \bA \btheta 
	+ \bc^\top \bB \bc 
	+ 2 \btheta^\top \bD \bc
	+ \btheta^\top \br_1 + \bc^\top \br_2 + r_0,
\end{align*}
with symmetric $\bA$ and $\bB$.
Also recall that the sampling distribution $q(\btheta|\bc)$ is Gaussian
\begin{align*}
q(\btheta|\bc) 
	&= \mathcal{N}(\btheta|\bb + \bK\bc, \bQ) \\
	&= \frac{1}{{| 2\pi\bQ |}^{\frac{1}{2}}} 
		\exp\left( -\frac{1}{2}[\btheta - (\bb+\bK\bc)]^\top \bQ^{-1} [\btheta - (\bb+\bK\bc)] \right).
\end{align*}
The quadratic model and the Gaussian distribution allow us to compute the dual function from data as follows.
Firstly, we consider the term 
\begin{align*}
q(\btheta|\bc)^{\frac{\eta}{\eta+\omega}} \exp\left( \frac{{R}(\btheta, \bc)}{\eta+\omega} \right).
\end{align*}
Using the Gaussian distribution $q(\btheta|\bc)$
and replacing $R(\btheta, \bc)$ with $\widehat{R}(\btheta, \bc)$ yield
\begin{align*}
& q(\btheta|\bc)^{\frac{\eta}{\eta+\omega}}
	 \exp\left( \frac{\widehat{R}(\btheta, \bc)}{\eta+\omega} \right) \\
&=
\frac{1}{| 2\pi\bQ |^{\frac{\eta}{2(\eta+\omega)}}}
		\exp\left( - \frac{\eta}{2(\eta+\omega)} [\btheta - (\bb+\bK\bc)]^\top \bQ^{-1} [\btheta - (\bb+\bK\bc)] \right) \\
&\phantom{=} \times \exp\left( \frac{
	\btheta^\top \bA \btheta 
	+ \bc^\top \bB \bc 
	+ 2 \btheta^\top \bD \bc
	+ \btheta^\top \br_{1} + \bc^\top \br_{2} + r_0 }
	{ \eta+\omega }
	\right) \\
&= 	\frac{1}{| 2\pi\bQ |^{\frac{\eta}{2(\eta+\omega)}}} 
	\exp \bigg( - \frac{1}{\eta+\omega} 
	\Big(
	- \frac{\eta}{2} [\btheta - (\bb+\bK\bc)]^\top \bQ^{-1} [\btheta - (\bb+\bK\bc)] \\
&\phantom{=} 
	+ \btheta^\top \bA \btheta 
	+ \bc^\top \bB \bc 
	+ 2 \btheta^\top \bD \bc
	+ \btheta^\top \br_{1} + \bc^\top \br_{2} + r_0
	\Big) 		
	\bigg) \\
&= \frac{1}{| 2\pi\bQ |^{\frac{\eta}{2(\eta+\omega)}}} 
	\exp \bigg( \frac{1}{2(\eta+\omega)} \Big( 
	-\eta \btheta^\top \bQ^{-1} \btheta 
	+ 2 \eta \btheta^\top \bQ^{-1} \bb
	+ 2 \eta \btheta^\top \bQ^{-1} \bK \bc 
	- \eta \bb^\top \bQ^{-1} \bb 
\\ &\phantom{=} 
- 2 \eta \bb^\top \bQ^{-1} \bK \bc 
	- \eta \bc^\top \bK^\top \bQ^{-1} \bK \bc
	+ 2 \btheta^\top \bA \btheta 
	+ 2 \bc^\top \bB \bc 
	+ 4 \btheta^\top \bD \bc
	+ 2 \btheta^\top \br_1
	+ 2 \bc^\top \br_2 
	+ 2 r_0
	\Big) \bigg) \\
&= 	\frac{1}{| 2\pi\bQ |^{\frac{\eta}{2(\eta+\omega)}}} 
	\exp \bigg( \frac{1}{2(\eta+\omega)} \Big( 
	\btheta^\top \left(-\eta\bQ^{-1} + 2\bA \right) \btheta
	+ \btheta^\top \left( 2\eta\bQ^{-1}\bb + 2\br_1 \right)
\\ &\phantom{=}
	+ \btheta^\top \left( 2\eta\bQ^{-1}\bK + 4\bD \right) \bc
	   + G \Big) \bigg) \\
&= \frac{1}{| 2\pi\bQ |^{\frac{\eta}{2(\eta+\omega)}}} 
	\exp \bigg( \frac{1}{2(\eta+\omega)} 
	\Big(
	- ( \btheta^\top \bF \btheta - 2\btheta^\top\bff - 2\btheta^\top\bL\bc)
	\Big)	\bigg)
	 \exp\bigg( \frac{G}{2(\eta+\omega)}\bigg),
\end{align*}
where 
\begin{align*}
\bF &= \eta \bQ^{-1} - 2 \bA, \\
\bff &= \eta \bQ^{-1}\bb + \br_1, \\
\bL &= \eta \bQ^{-1}\bK + 2\bD, \\
G &=  -\eta\bb^\top\bQ^{-1} \bb 
	   - 2\eta\bb^\top\bQ^{-1}\bK\bc
	   - \eta \bc^\top \bK^\top \bQ^{-1} \bK \bc 
	   + 2\bc^\top \bB \bc
	   + 2\bc^\top \br_2 
	   + 2 r_0.
\end{align*}
Next, we ``complete the square'' by considering the following quadratic term
\begin{align*}
&[\btheta - (\bF^{-1}\bff+\bF^{-1}\bL\bc)]^\top \bF [\btheta - (\bF^{-1}\bff+\bF^{-1}\bL\bc)]
\\
&= \btheta^\top\bF\btheta - 2\btheta^\top\bff - 2\btheta^\top\bL\bc 
	+ (\bF^{-1}\bff+\bF^{-1}\bL\bc )^\top\bF(\bF^{-1}\bff+\bF^{-1}\bL\bc) \\
&= \left( \btheta^\top\bF\btheta - 2\btheta^\top\bff - 2\btheta^\top\bL\bc \right)
	+ \bff^\top \bF^{-1} \bff + 2 \bff^\top \bF^{-1}\bL\bc + \bc^\top \bL^\top \bF^{-1}\bL\bc.
\end{align*}
Therefore, we have
\begin{align}
&q(\btheta|\bc)^{\frac{\eta}{\eta+\omega}}
	 \exp\left( \frac{R(\btheta, \bc)}{\eta+\omega} \right) \notag \\
&= \frac{1}{| 2\pi\bQ |^{\frac{\eta}{2(\eta+\omega)}}} 
	\exp \bigg( \frac{1}{2(\eta+\omega)} 
	\Big(
	- ( [\btheta - (\bF^{-1}\bff+\bF^{-1}\bL\bc)]^\top \bF [\btheta - (\bF^{-1}\bff+\bF^{-1}\bL\bc)]
 \notag \\ &\phantom{=}
+ \bff^\top \bF^{-1} \bff + 2 \bff^\top \bF^{-1}\bL\bc + \bc^\top \bL^\top \bF^{-1}\bL\bc
	) \Big)	\bigg)  \exp\bigg( \frac{G}{2(\eta+\omega)}\bigg). \label{eq:weight_policy}
\end{align}
Using the above result, the inner integral term in the dual function is
\begin{align*}
&\int q(\btheta|\bc)^{\frac{\eta}{\eta+\omega}}
	 \exp\left( \frac{\widehat{R}(\btheta, \bc)}{\eta+\omega} \right) \dtheta \\
&= \frac{|2\pi\bF^{-1}(\eta+\omega)|^{\frac{1}{2}} }{ |2\pi\bQ|^{\frac{\eta}{2(\eta+\omega)}} }
\exp\bigg( \frac{1}{2(\eta+\omega)} \Big(
\bff^\top \bF^{-1} \bff + 2 \bff^\top \bF^{-1}\bL\bc + \bc^\top \bL^\top \bF^{-1}\bL\bc
\Big) \bigg) 
\exp\bigg( \frac{G}{2(\eta+\omega)} \bigg),
\end{align*}
where the squared exponential term in Eq.~\eqref{eq:weight_policy} depending on $\btheta$ is ``integrated out'' and becomes the inverted normalization term $|2\pi\bF^{-1}(\eta+\omega)|^{\frac{1}{2}} $.
Plugging this term back to the dual function yields
\begin{align*}
g(\eta, \omega) 
&= \eta\epsilon - \omega\beta 
   + \frac{1}{2}\bigg(
   \bff^\top\bF^{-1}\bff -\eta \bb^\top\bQ^{-1}\bb
   + (\eta+\omega) \log | 2\pi \bF^{-1}(\eta+\omega) |
   - \eta \log | 2\pi\bQ |  \bigg) 
\\ &\phantom{=} 
 + \int \mu(\bc) \bigg( \bff^\top \bF^{-1}\bL -\eta \bb^\top \bQ^{-1}\bK \bigg) \bc \dc
+ \frac{1}{2} \int \mu(\bc) \bc^\top \bigg( \bL^\top \bF^{-1} \bL - \eta \bK^\top \bQ^{-1}\bK \bigg) \bc \dc \\
&= \eta\epsilon - \omega\beta 
   + \frac{1}{2}\bigg(
   \bff^\top\bF^{-1}\bff -\eta \bb^\top\bQ^{-1}\bb
   + (\eta+\omega) \log | 2\pi \bF^{-1}(\eta+\omega) |
   - \eta \log | 2\pi\bQ |  \bigg) 
\\ &\phantom{=} 
	+ \int \mu(\bc) \left( \bc^\top \bm + \frac{1}{2}\bc^\top \bM \bc \right) \dc,
\end{align*}
where
\begin{align*}
\bm &= \bL^\top \bF^{-1} \bff - \eta \bK^\top \bQ^{-1} \bb, \\
\bM &= \bL^\top \bF^{-1} \bL - \eta \bK^\top \bQ^{-1}\bK.
\end{align*}
The expectation over $\mu(\bc)$ can be approximated by the context samples.
The term $2\bc^\top \bB \bc + 2\bc^\top\br_2 + 2r_0$ in $G$ does not appear in the dual function since this is constant w.r.t. $\eta$ and $\omega$.
Similarly to the dual function, by using Eq.~\eqref{eq:weight_policy_normalize} and Eq.~\eqref{eq:weight_policy} we compute the new search distribution in closed form as
\begin{align*}
\pi(\btheta|\bc) 
&\propto q(\btheta|\bc)^{\frac{\eta}{\eta+\omega}}
	 \exp\left( \frac{\widehat{R}(\btheta, \bc)}{\eta+\omega} \right) \\
&\propto 
\exp \bigg( \frac{1}{2(\eta+\omega)} 	\Big(
	- ( [\btheta - (\bF^{-1}\bff+\bF^{-1}\bL\bc)]^\top \bF [\btheta 
	- 	(\bF^{-1}\bff+\bF^{-1}\bL\bc)] \Big) \bigg) \\
&= \mathcal{N}\Big(\btheta | \bF^{-1}\bff + \bF^{-1}\bL\bc, \bF^{-1}(\eta+\omega)\Big).
\end{align*}

\section{Proof of Convexity of $\widehat{\mathcal{J}}(\bH)$}

To show that 
$
\widehat{\mathcal{J}}(\bH) = \frac{1}{2N}\sum_{n=1}^N \left( \bx_n^\top \bH \bx_n - R(\btheta_n, \bc_n) \right)^2
$ 
is a convex function, we follow the proof in  Boyd and Vandenberghe 2004, page 74.
Let $g(t) = \widehat{\mathcal{J}}(\bZ + t\bV)$ with symmetric matrices $\bZ$ and $\bV$ and scalar $t$.
Then, we can verify the convexity of $\widehat{\mathcal{J}}$ through $g$.
Through simple calculation, we have that
\begin{align*}
g(t) &= \frac{1}{2N} \sum_{n=1}^N \left( \bx_n^\top (\bZ+t\bV) \bx_n - R(\btheta_n,\bc_n) \right)^2 \\
&= \frac{1}{2N} \sum_{n=1}^N \bigg( (\bx_n^\top \bZ \bx_n)^2 + 2 t(\bx_n^\top \bZ \bx_n) (\bx_n^\top \bV \bx_n) - 2(\bx_n^\top \bZ \bx_n) R(\bx_n) \\
&\phantom{=} \quad\quad\quad\quad + t^2 (\bx_n^\top\bV\bx_n)^2 - 2t (\bx_n^\top\bV\bx_n) R(\bx_n) + R(\bx_n)^2
\bigg).
\end{align*}
The first and second derivatives of $g(t)$ are
\begin{align*}
\nabla g(t) &= \frac{1}{N} \sum_{n=1}^N \bigg(
(\bx^\top \bZ \bx)(\bx^\top\bV\bx) + t(\bx_n^\top \bV \bx_n)^2 
- (\bx_n^\top \bV \bx_n)R(\bx_n)
\bigg), \\
\nabla^2 g(t)&= \frac{1}{N} \sum_{n=1}^N (\bx_n^\top \bV \bx_n)^2  \geq 0.
\end{align*}
Since the second derivative is non-negative, the function $\widehat{\mathcal{J}}(\bH)$ is convex.

\section{Experiments Details}
Here, we provide all additional details regarding the experiments that are not mentioned in the paper.

\subsection{Quadratic Cost Function Optimization}

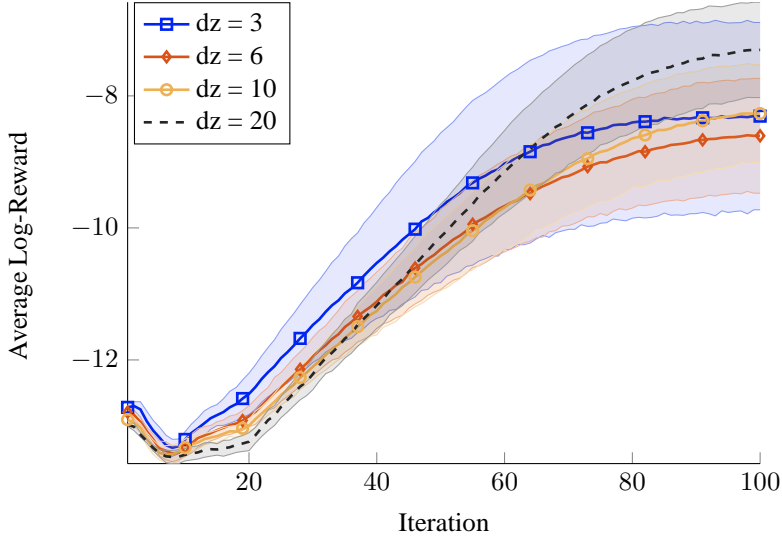
\begin{figure*}[h]
	\centering
	\input{fig_quadexp_pca}
	\caption{\label{fig:fig_quadexp_pca} Average reward for the quadratic cost function problem with C-MORE PCA using different dimensionality $\zd$. Shaded area denotes standard deviation (results are averaged over ten trials). Results do not differ much, since PCA always fails in reducing the dimensionality of the context variables. Clearly, the more principal components we keep, the better results we obtain.}
\end{figure*}

The sampling Gaussian distribution mean is initialized uniformly randomly in $[0,5]$, while the initial covariance is $\bQ = 10,000 \bI$.
For nuclear norm, the regularization parameters are set to $\lambda_{*} = 0.00002$ and $\lambda = 0.00001$. 
For PCA, the dimensionality candidates are
$\zd \in \{3, 6, 10, 20\}$. Figure~\ref{fig:fig_quadexp_pca} shows the performance with different dimensionality.
The regularization parameter for the $\ell_2$-regularization is set at $\lambda = 0.00001$.
The step size for accelerated proximal gradient (APG) is fixed at $0.001$. 
We also normalize the gradient such that the Frobenius norm is $1$.
The maximum iteration of APG is set at $K = 300$.
C-REPS and C-MORE KL divergence is set to $0.9$.

\subsection{Ball Hitting with a 2-DoF Robot Arm}
We use a linear-in-parameter controller with Gaussian basis functions to control the joint accelerations $\ddot{\boldsymbol{q}}$ of the robot
\begin{align*}
\ddot{\boldsymbol{q}} = \left[ \btheta_1^\top \bvphi(\boldsymbol{q}), 
\quad \btheta_2^\top \bvphi(\boldsymbol{q}) \right]^\top,
\end{align*}
where $\btheta = \left[ \btheta_1^\top, \btheta_2^\top\right]^\top$ is the policy parameters vector, $\boldsymbol{q} \in \mathbb{R}^2$ is the joint angles vector, and $\bvphi$ is the Gaussian basis functions vector with $16$ Gaussian centers placed at $\{ 0, \frac{\pi}{2}, \pi, \frac{3\pi}{2}\} \times \{ 0, \frac{\pi}{2}, \pi, \frac{3\pi}{2}\}$ for both joints.
The number of total parameter $\btheta$ is $32$.
The reward ${R}(\btheta, \bc)$ is computed as the negative cumulative joint accelerations plus the negative distance between the end-effector and the ball at the final time step
\begin{align*}
R(\btheta, \bc) = -0.05\sum_{t=1}^{40} | \ddot{\boldsymbol{q}}_t | 
+ 10\exp\left( -\frac{ (\mathrm{ball}_\mathrm{x} - \mathrm{eff}_\mathrm{x})^2 
- (\mathrm{ball}_\mathrm{y} - \mathrm{eff}_\mathrm{y})^2}{50}\right),
\end{align*}
where $\ddot{\boldsymbol{q}}_t$ denotes the joint accelerations at time step $t$, 
$\mathrm{ball}_\mathrm{x}$ and $\mathrm{eff}_\mathrm{x}$ denotes the $\mathrm{x}$ coordinate of the ball and end-effector at the final time step, respectively. Variables $\mathrm{ball}_\mathrm{y}$ and $\mathrm{eff}_\mathrm{y}$ are also also defined similarly.
We use the following transition dynamics to govern the joint angles
\begin{gather*}
\dot{\boldsymbol{q}} \leftarrow \dot{\boldsymbol{q}} + \Delta \ddot{\boldsymbol{q}},
\\
{\boldsymbol{q}} \leftarrow {\boldsymbol{q}} + \Delta \dot{\boldsymbol{q}},
\end{gather*}
where $\Delta = 0.1$.
Each arm has length $7.5$. 
The robot is initialized such that the end-effector is at the bottom position.

The sampling Gaussian distribution $\mathcal{N}(\btheta|\boldsymbol{b} + \bK\bc, \bQ)$ is initialized by ${b}_{ij} \sim \mathcal{N}({0}, {1})$, $K_{ij} \sim \mathcal{N}({0}, {0.01^2})$, for each entry $(i,j)$ and $\bQ = \boldsymbol{I}$.
For nuclear norm, the regularization parameter candidates are
$\lambda_{*} \in \{1 \times 10^{-7}, 5\times 10^{-7}, 1\times 10^{-6}, 2\times 10^{-6}\}$.
For PCA, the dimensionlaity candidates are
$\zd \in \{10, 20, 30, 40\}$.
The regularization parameter for the $\ell_2$-regularization is set at $\lambda = 0.0001$.
The step size for accelerated proximal gradient (APG) is fixed at $0.001$. 
We also normalize the gradient such that the Frobenius norm is $1$.
The maximum iteration of APG is initialize at $K = 500$ and it is reduced by $2$ after each update to reduce the computation time until a minimum of $K=300$.
$K$ is reset to 500 when cross-validation is performed.

\begin{figure}[h]
\centering
	\begin{subfigure}[b]{0.49\linewidth}
	\centering
	\includegraphics[width=1\linewidth]{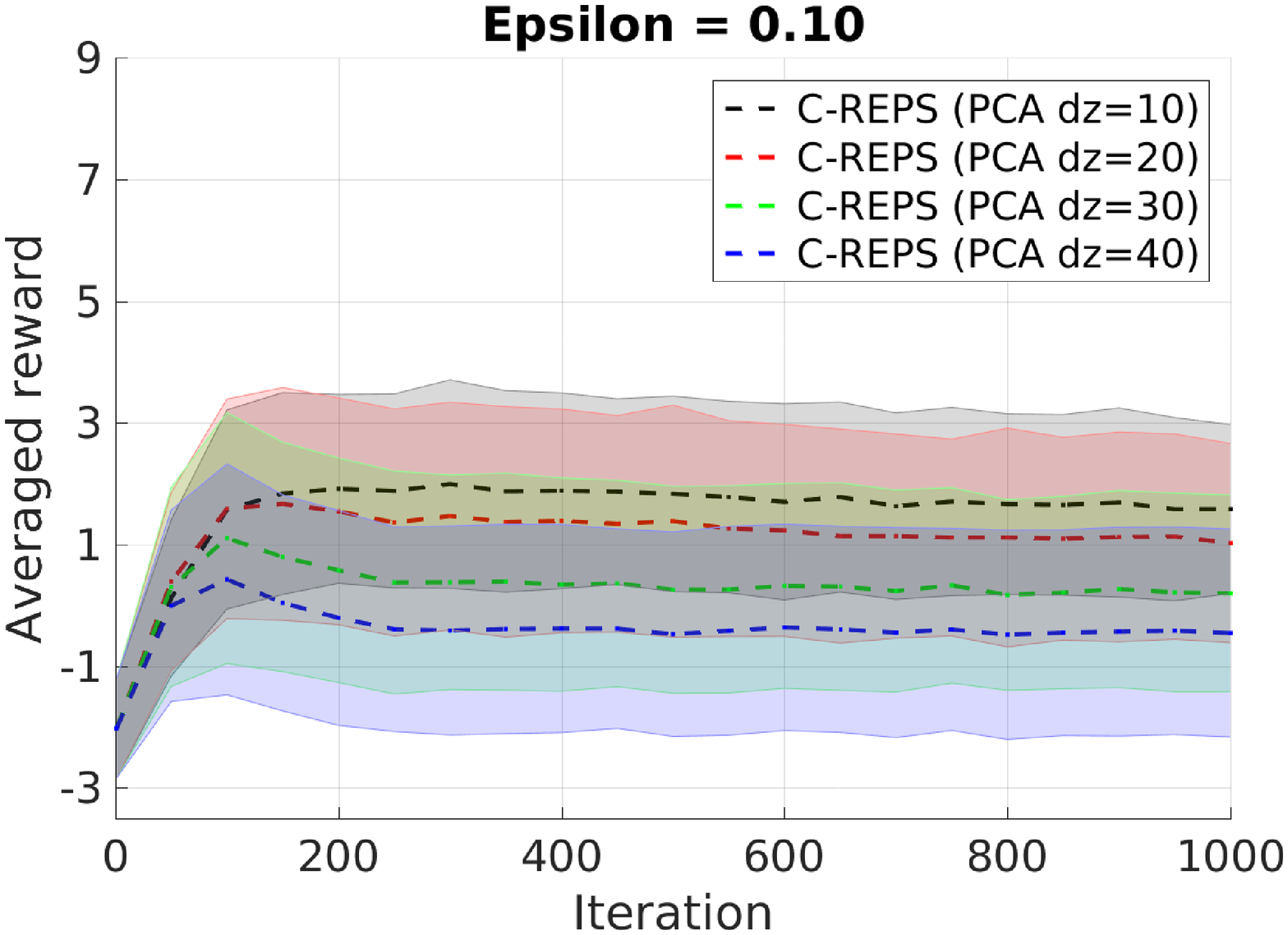}
	\end{subfigure}
	\begin{subfigure}[b]{0.49\linewidth}
	\centering
	\includegraphics[width=1\linewidth]{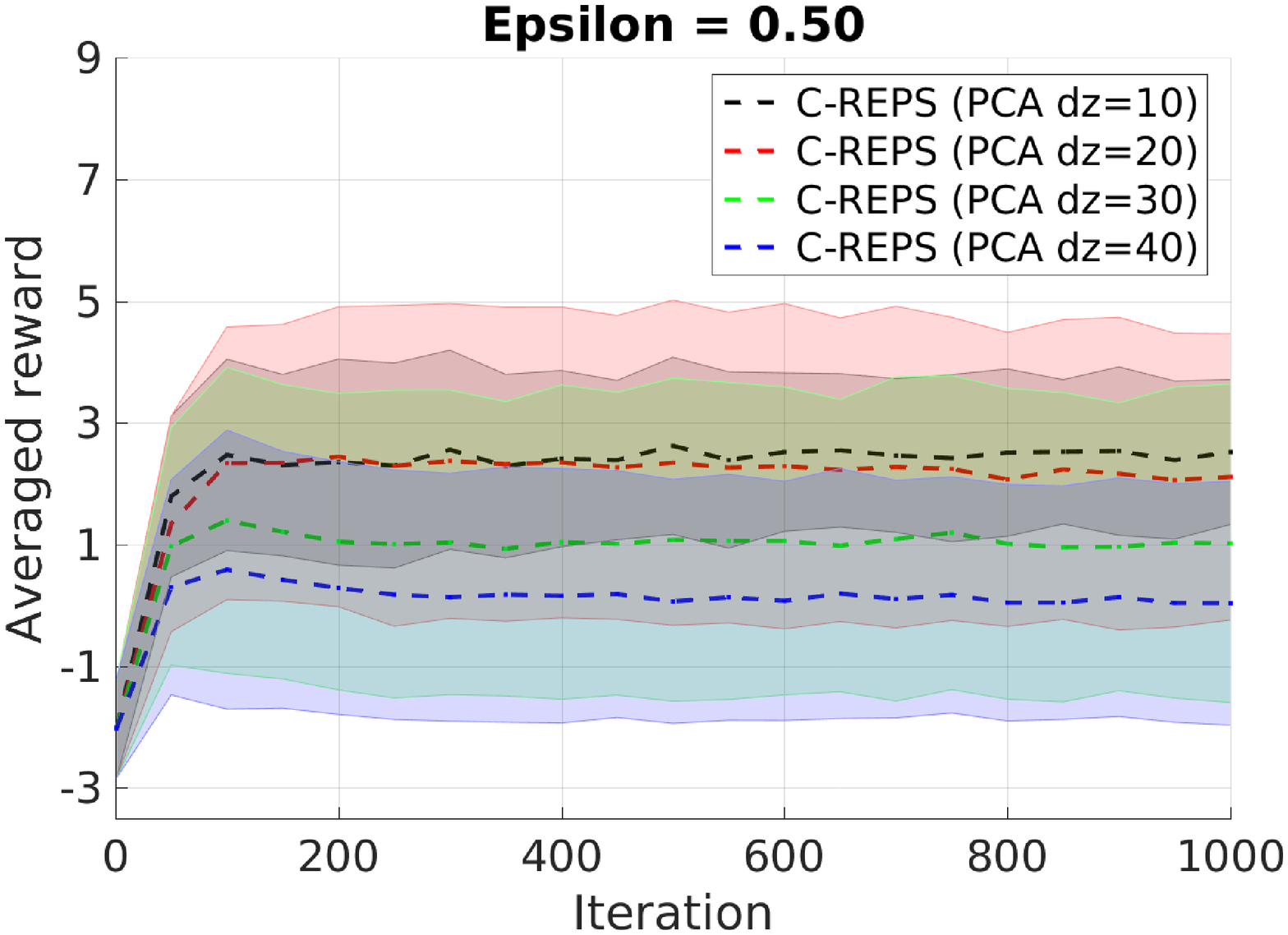}
	\end{subfigure}
	\begin{subfigure}[b]{0.49\linewidth}
	\centering
	\includegraphics[width=1\linewidth]{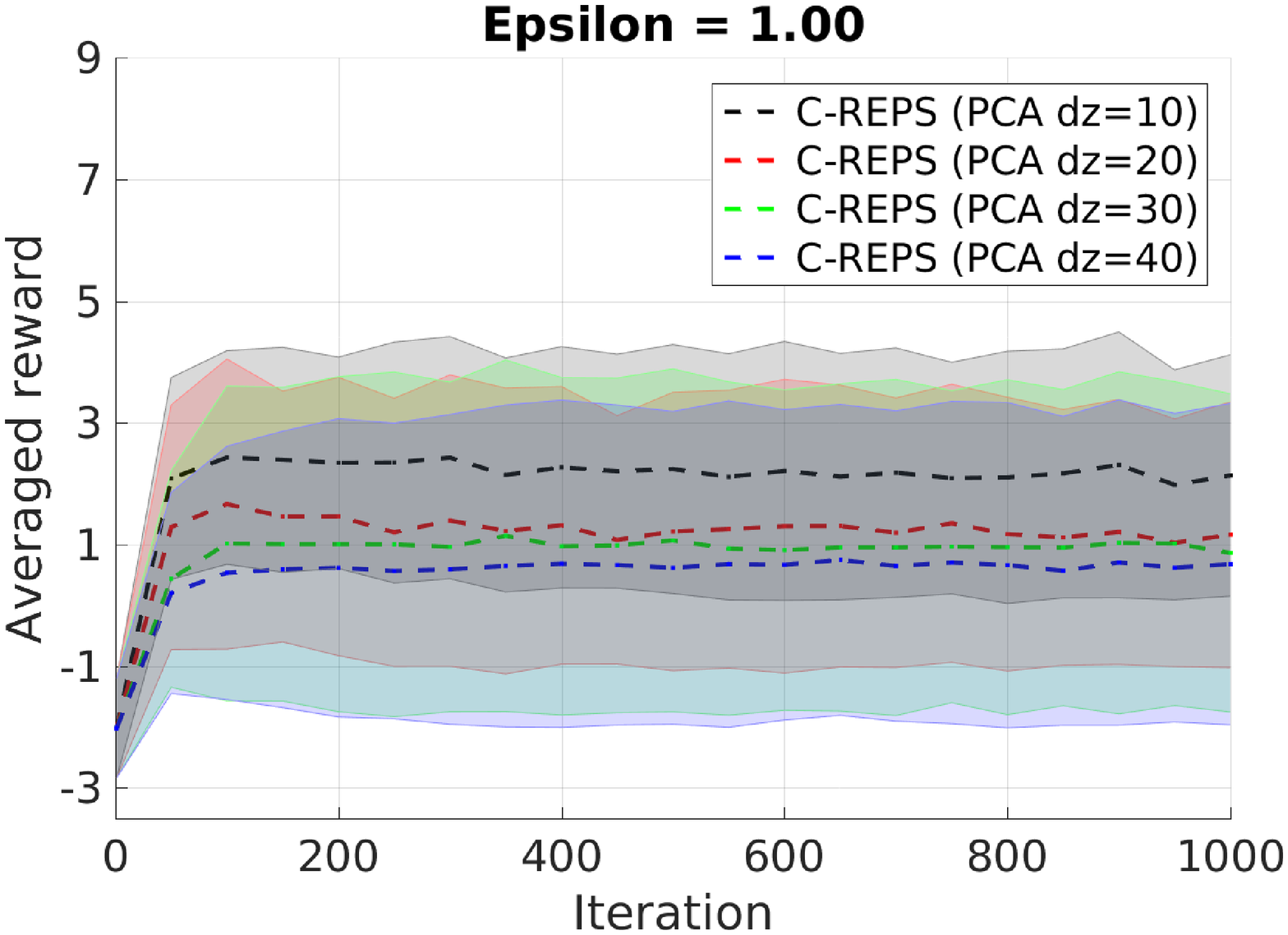}
	\end{subfigure}
	\caption{Results of C-REPS with different $\epsilon$ and $\zd$ for the 2-DoF hitting task.}
	\label{fig:creps_result}
\end{figure}

\begin{figure}[h]
\centering
	\includegraphics[width=0.50\linewidth]{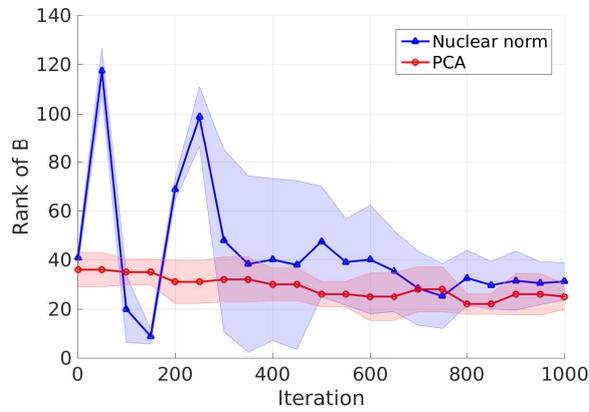}
	\caption{Rank of $\bB$ for the 2-DoF hitting task.
	At the $1000$th iteration the rank of $\bB$ learned by nuclear norm is approximately 31, while PCA is approximately 25.}
	\label{fig:2dof_rank}
\end{figure}

Figure \ref{fig:creps_result} shows the performance of C-REPS with PCA on different dimensionality $\zd$ and different KL upper-bound $\epsilon$.
The best result of C-REPS is the one reported in the main paper for the comparison with C-MORE.
Figure \ref{fig:2dof_rank} shows the rank of $\bB$ against update iteration averaged over ten trials.
C-MORE with nuclear norm is quite unstable early on. 
However, the rank stabilizes after some iterations and the rank converges to $31$.

\subsection{Ball Hitting with a 6-DoF Robot Arm}
\begin{figure}[h]
\centering
	\begin{subfigure}[b]{0.32\linewidth}
	\centering
	\includegraphics[width=1\linewidth]{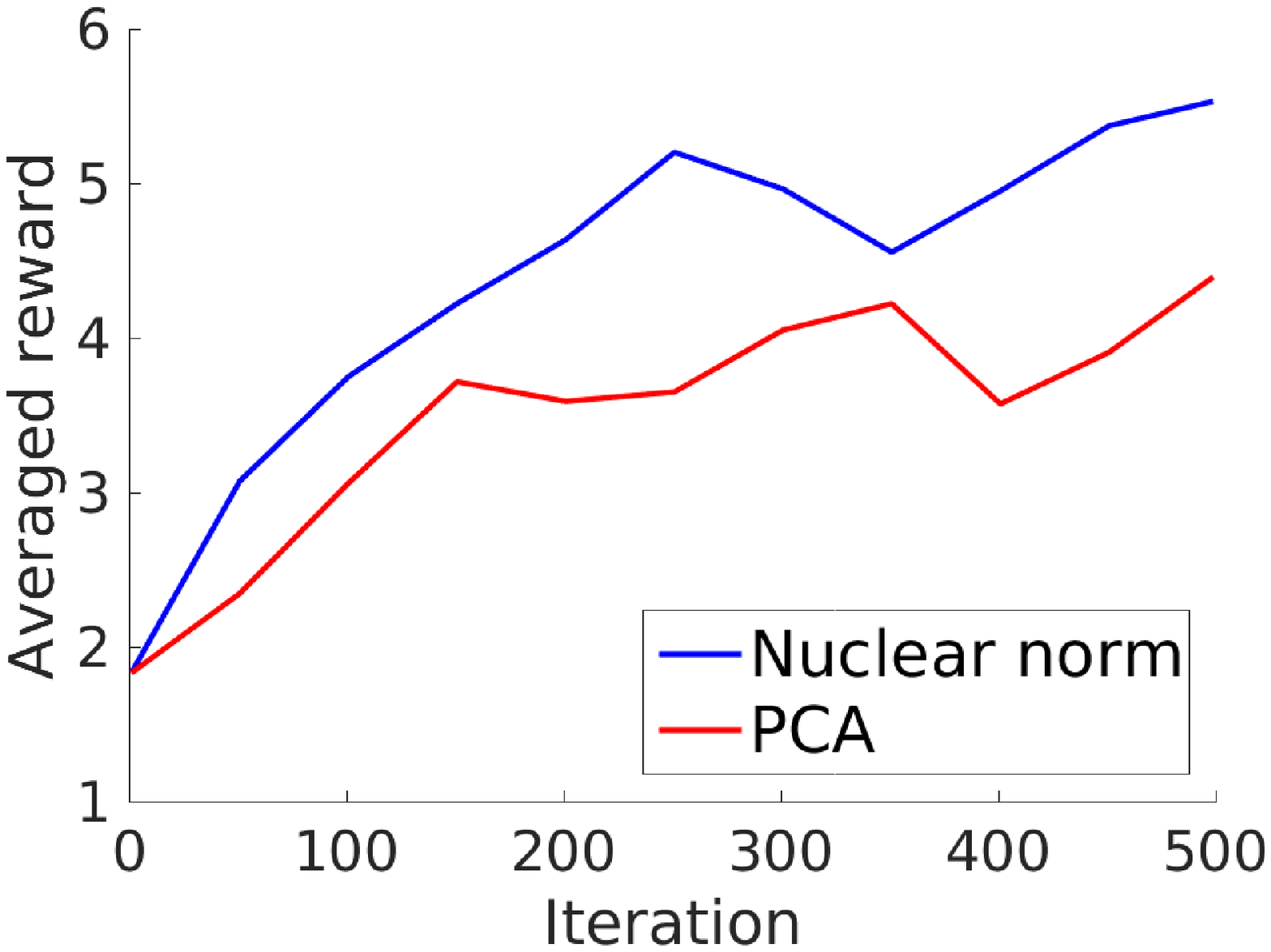}
	\end{subfigure}
	\begin{subfigure}[b]{0.32\linewidth}
	\centering
	\includegraphics[width=1\linewidth]{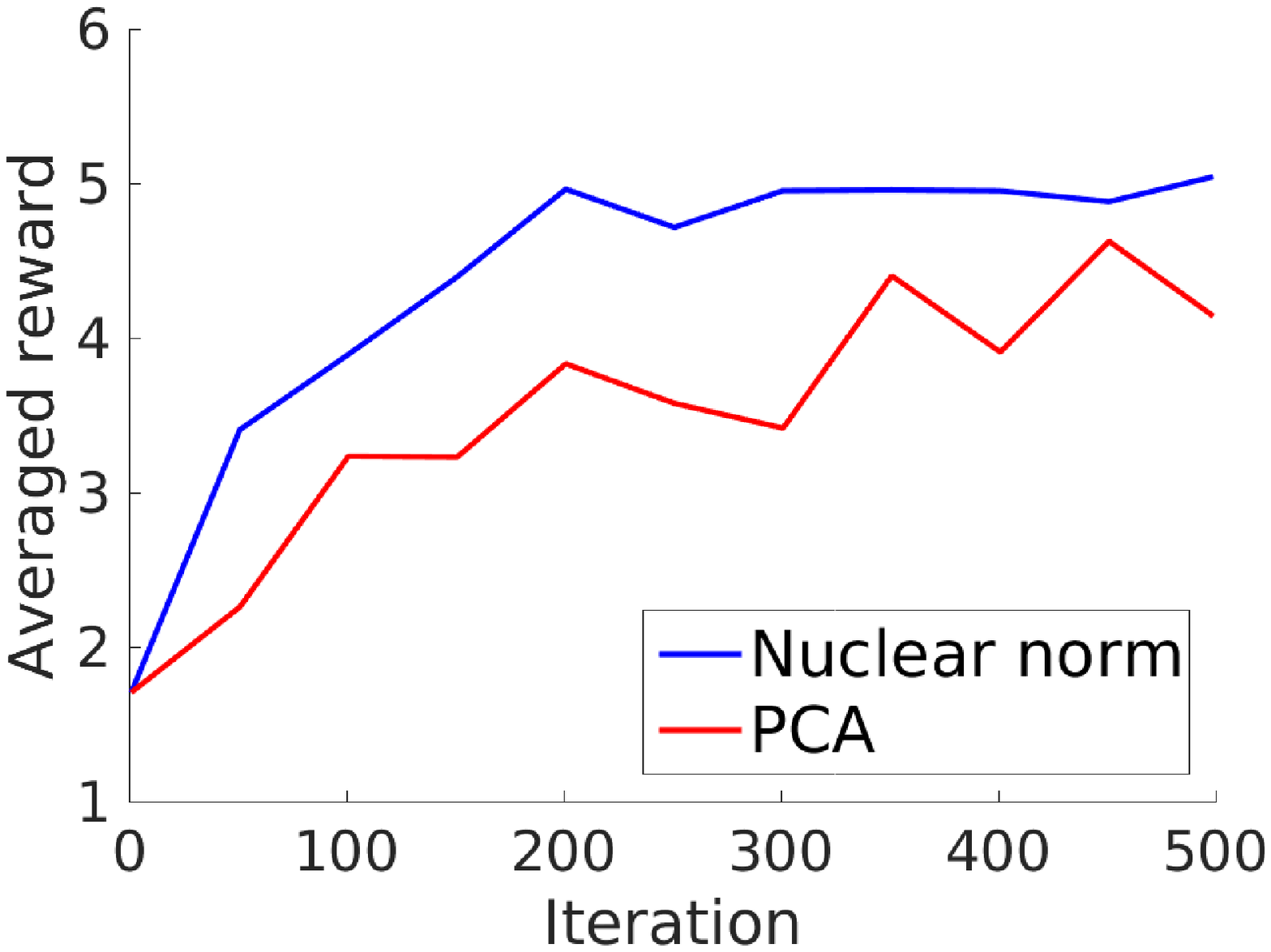}
	\end{subfigure}
	\begin{subfigure}[b]{0.32\linewidth}
	\centering
	\includegraphics[width=1\linewidth]{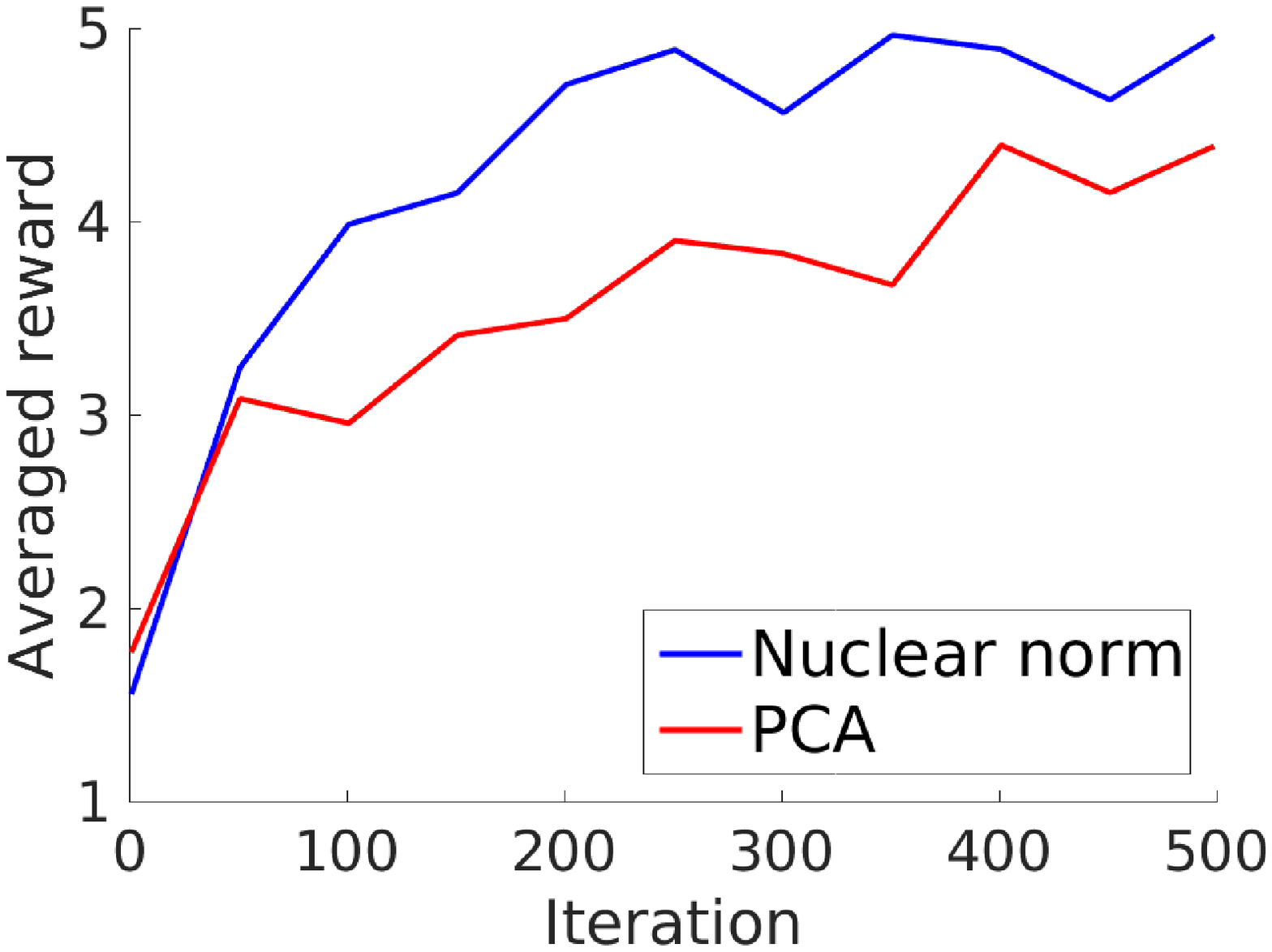}
	\end{subfigure}
	\caption{Average reward of three individual trials on the 6-DoF ball hitting task.}
	\label{fig:creps_indiv_reward}
\centering
	\begin{subfigure}[b]{0.32\linewidth}
	\centering
	\includegraphics[width=1\linewidth]{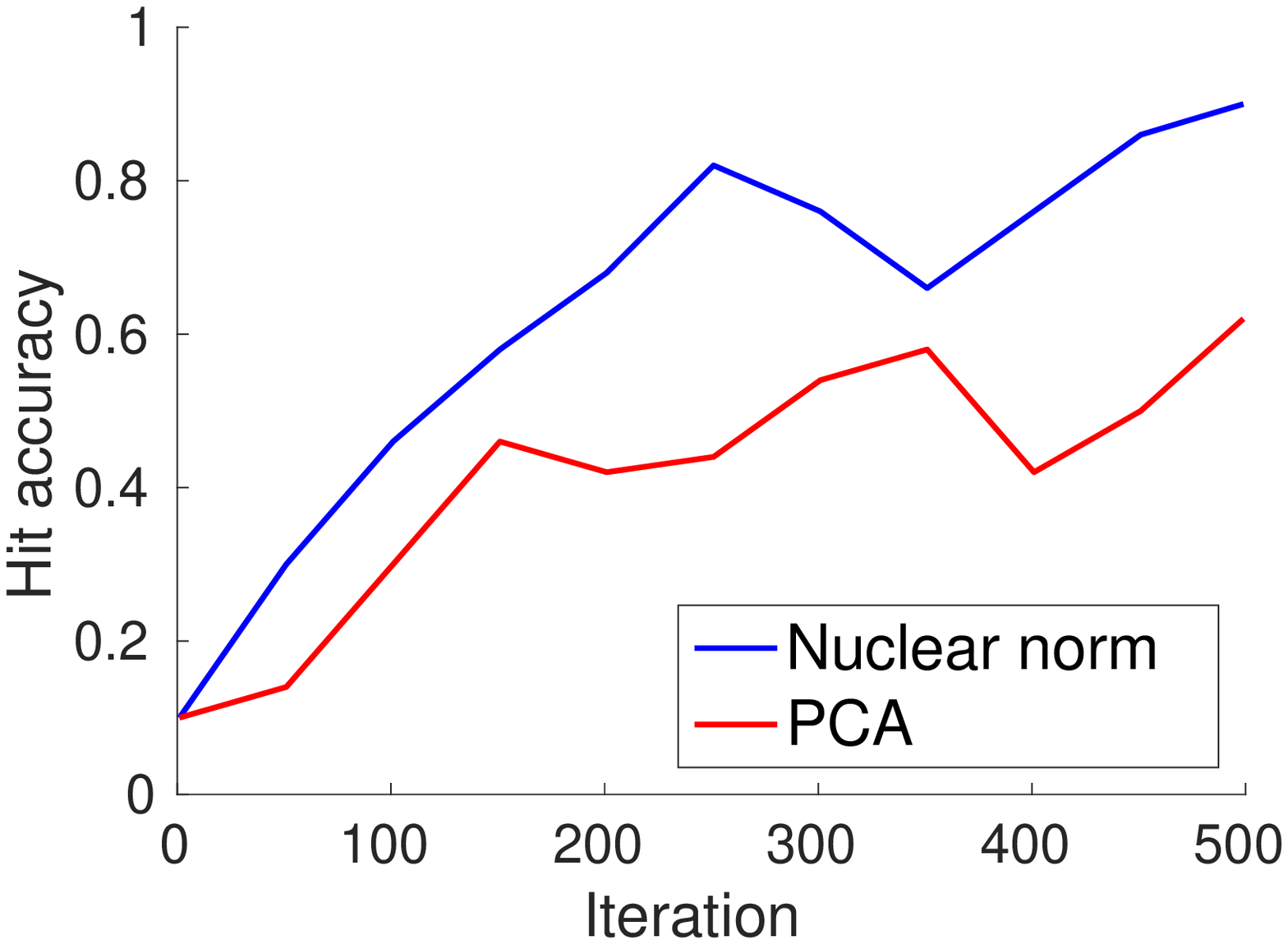}
	\end{subfigure}
	\begin{subfigure}[b]{0.32\linewidth}
	\centering
	\includegraphics[width=1\linewidth]{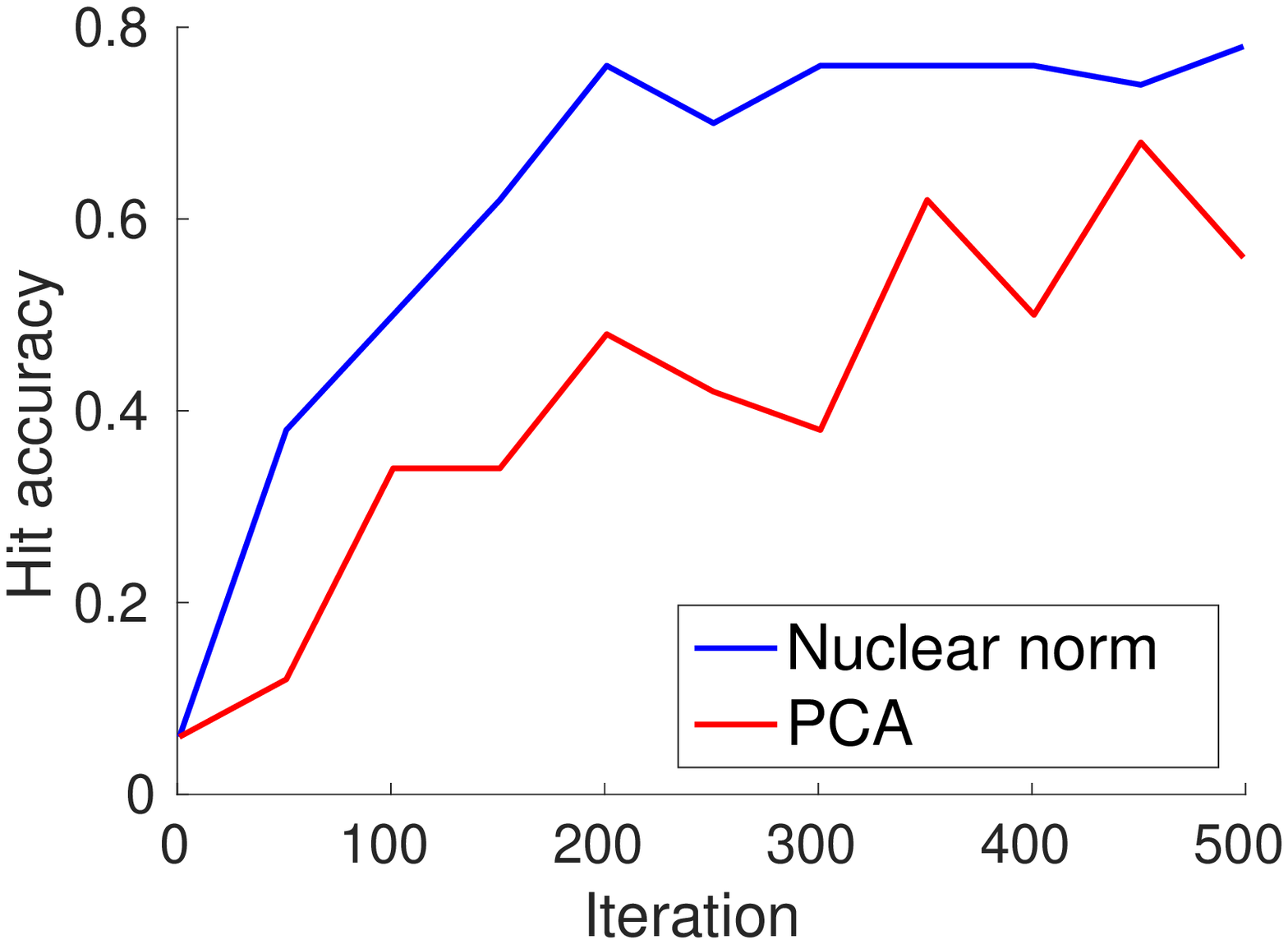}
	\end{subfigure}
	\begin{subfigure}[b]{0.32\linewidth}
	\centering
	\includegraphics[width=1\linewidth]{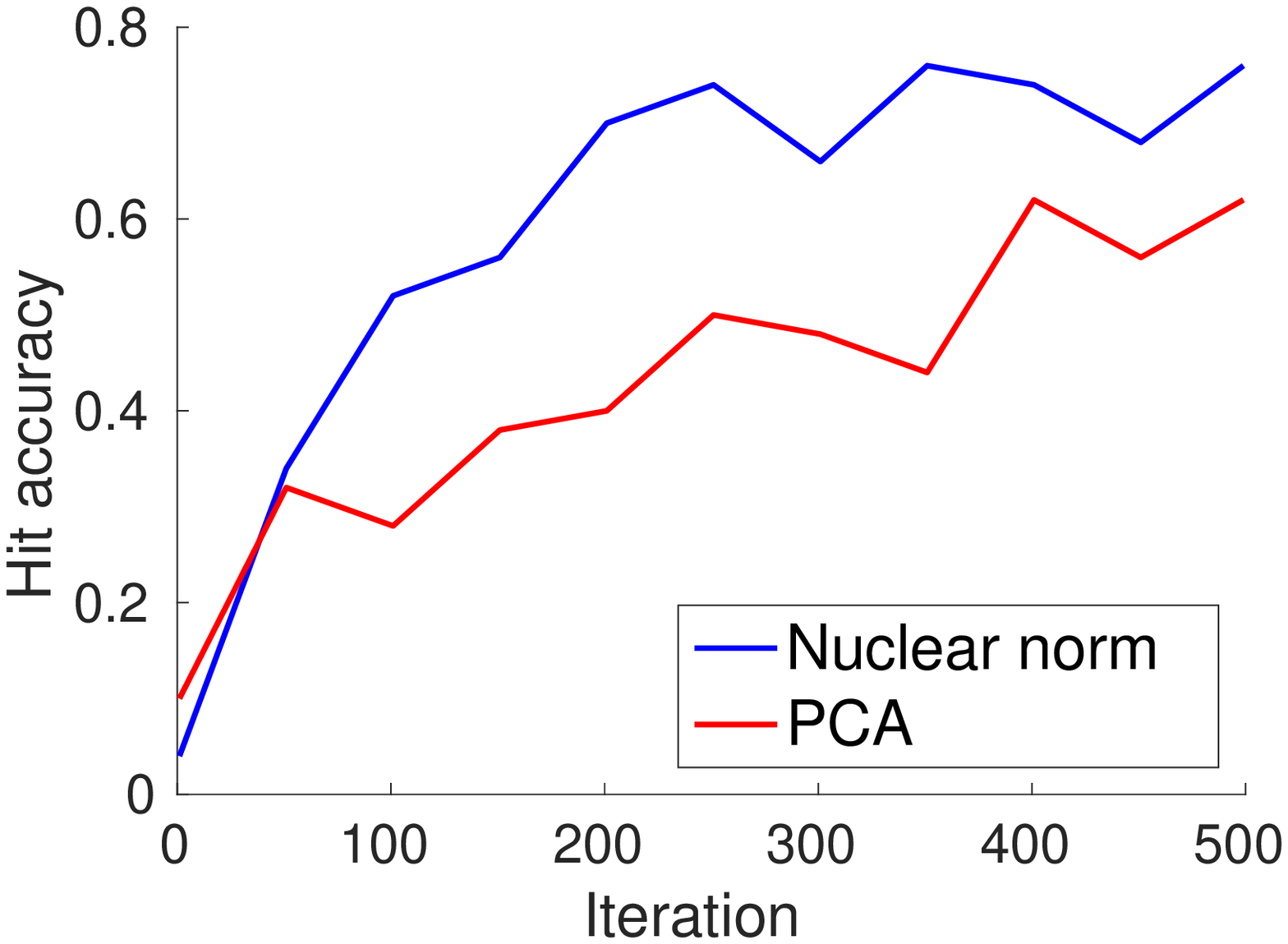}
	\end{subfigure}
	\caption{Hit accuracy of three individual trials on the 6-DoF ball hitting task.}
	\label{fig:creps_indiv_hit}
\end{figure}

In this task, the reward function is defined as
\begin{align*}
R(\btheta, \bc) =
\begin{cases}
-0.05\sum|\boldsymbol{q}_t| + 10 * \exp( - \mathrm{distance}^2), &\text{if } \mathrm{distance} > 0.1 \text{ cm.} \\
-0.05\sum\boldsymbol{q}_t + 10 * \exp( - \mathrm{distance}^2) + 20, &\text{otherwise},
\end{cases}
\end{align*}
where $\mathrm{distance}$ denotes the minimum distance between the ball and the end-effector along the trajectory.
APG and PCA setup is the same as in the 2-DoF robot experiment.
Figure~\ref{fig:creps_indiv_reward} and Figure~\ref{fig:creps_indiv_hit} show the average reward and hit accuracy averaged over 50 contexts on three individual trial, respectively.
On all trials, the nuclear norm performs better than PCA and consistently achieves $80\%$ hit accuracy.

\begin{algorithm}[b]
\caption{APG for solving the nuclear norm minimization problem}
\label{algo_APG}
\DontPrintSemicolon
\KwIn{Parameters $\lambda$ and $\lambda_{*}$, gradient step size $\tau$, 
maximum number of iteration $K$, initial solution $\bH_0$}
Initialize $\bH_{-1} = \bH_0$ and $t_{-1} = t_0 = 1$ \;
\For{$k=1,\dots,K$}{

	Set intermediate point
	\begin{align*}
	\bY_k = \bH_k + \frac{t_{k_{-1}} - 1}{t_k} \left( \bH_k - \bH_{k-1} \right)
	\end{align*} 

	\bigskip

	Do gradient descent using the differentiable term
	\begin{align*}
	\bY_{+} = \bY_k + \tau \nabla \mathcal{J}(\bY_k),
	\end{align*} 
	where
	\begin{align*}
	\bY_{+} &= 
	\begin{bmatrix}
	\bA_+ 	 & \bD_+ 	& 0.5\br_{1+} \\
	\bD^\top_+ 	 & \bB_+ 	& 0.5\br_{2+} \\
	0.5\br_{1+}^\top 	 & 0.5\br_{2+}^\top & r_{0+}
	\end{bmatrix}
	\end{align*}
	
	\bigskip

	Shrink singular values of $\bB_{+}$
	\begin{align*}
		\bB_{*} = \bU \max(\bSigma-\lambda_{*}\bI, \boldsymbol{0}) \bV^\top,
	\end{align*}
	where $\bB_{+} = \bU \bSigma \bV^\top$ is the SVD of $\bB_{+}$ \label{algo:APG:line_SVT}
	
	\bigskip

	Truncate positive eigenvalues of $\bA_{+}$
	\begin{align*}
		\bA_{*} = \bP \min(\boldsymbol{\Lambda}, \boldsymbol{0}) \bP^\top,
	\end{align*}
	where $\bA_{+} = \bP \boldsymbol{\Lambda} \bP^\top$ is the eigendecomposition of $\bA_{+}$\; \label{algo:APG:line_NSD}
	
	\bigskip
	
	Update solution
	\begin{align*}
	\bH_{k+1} &= 
	\begin{bmatrix}
	\bA_* 	 & \bD_+ 	& 0.5\br_{1+} \\
	\bD^\top_+ 	 & \bB_* 	& 0.5\br_{2+} \\
	0.5\br_{1+}^\top 	 & 0.5\br_{2+}^\top & r_{0+}
	\end{bmatrix}
	\end{align*}
	
	\bigskip

	Update parameter
	$
	t_{k+1} = \frac{1 + \sqrt{1+4(t_k)^2}}{2}
	$	\;
	
	\bigskip

	\If{stopping criterion is met}	{
	\Return	}	
	}
\end{algorithm}

\end{document}

%% file: fig_quadexp2.tex
%
%
\definecolor{mycolor1}{rgb}{0.00000,0.14700,0.94100}%
\definecolor{mycolor2}{rgb}{0.85000,0.32500,0.09800}%
\definecolor{mycolor3}{rgb}{0.92900,0.69400,0.32500}%
\definecolor{mycolor4}{rgb}{0.15000,0.15000,0.15000}%
\definecolor{mycolor5}{rgb}{0.13500,0.67800,0.18400}%
\begin{tikzpicture}
\begin{axis}[%
height = 5.7cm,
width = \columnwidth,
ylabel style = {yshift = -.9em,
                xshift = -0.5em},
xlabel style = {yshift =  0.5em},
xlabel       = {Iteration},
ylabel       = {Average Log-Reward},
axis lines*  = left,
enlargelimits = false,
mark repeat = {9},
legend style = {at = {(0.01, 1)}, 
                anchor = north west,
                draw = black,
                fill = white,
                legend cell align = left},
]

\addplot[area legend,solid,fill=mycolor1,opacity=1.000000e-01,draw=none,forget plot]
table[row sep=crcr] {%
x	y\\
1	-15.9681846010789\\
2	-15.9223640611152\\
3	-15.8578471991556\\
4	-15.8128933421292\\
5	-15.7815167936868\\
6	-15.7622646328381\\
7	-15.731495143117\\
8	-15.6648347091568\\
9	-15.6134893662733\\
10	-15.6037529947446\\
11	-15.5719291979671\\
12	-15.5457656522361\\
13	-15.5493668605805\\
14	-15.5234912138058\\
15	-15.5018209936012\\
16	-15.4943621660078\\
17	-15.4720041556167\\
18	-15.4585722712799\\
19	-15.4421502830984\\
20	-15.3552643129143\\
21	-15.2640195458174\\
22	-15.2039076340396\\
23	-15.1313885302101\\
24	-15.0370110095517\\
25	-14.93722676607\\
26	-14.8554229231605\\
27	-14.7433829356225\\
28	-14.6581647678332\\
29	-14.590979759751\\
30	-14.4942070047855\\
31	-14.3970046457896\\
32	-14.3091246587774\\
33	-14.2238684977014\\
34	-14.1077569519555\\
35	-14.0019844940087\\
36	-13.9216995040953\\
37	-13.8155878170787\\
38	-13.6872739260006\\
39	-13.5750815192958\\
40	-13.4677825507212\\
41	-13.3414672465585\\
42	-13.2502647210338\\
43	-13.1217991921758\\
44	-12.9748655191938\\
45	-12.8401501904771\\
46	-12.6922258403686\\
47	-12.5550380823679\\
48	-12.4211956838528\\
49	-12.2688126005735\\
50	-12.1097310262218\\
51	-11.9470448870979\\
52	-11.7787453390275\\
53	-11.5960803427397\\
54	-11.4090622045821\\
55	-11.2336005615662\\
56	-11.0470354011729\\
57	-10.8604466145872\\
58	-10.6476877289129\\
59	-10.4371722210309\\
60	-10.3054362807062\\
61	-11.3328014769001\\
62	-12.1184973185311\\
63	-11.3022535985647\\
64	-10.7150722622284\\
65	-9.04939443870921\\
66	-8.30224054138511\\
67	-7.66655285597978\\
68	-6.1687616892046\\
69	-4.34190355772136\\
70	-3.42953668785212\\
71	-2.7601441398133\\
72	-1.08104131335988\\
73	1.19456433942269\\
74	2.00769231067421\\
75	2.11131017178945\\
76	2.36913500940107\\
77	2.61577429685034\\
78	2.79652242237423\\
79	3.291989244586\\
80	4.03014325731437\\
81	6.79063936803398\\
82	8.90478683458204\\
83	9.26475596812976\\
84	9.49957753439859\\
85	9.51467016210989\\
86	9.49916315878987\\
87	9.47271843271242\\
88	9.43949335787496\\
89	9.46688218903189\\
90	9.40574530680893\\
91	9.30295708470338\\
92	9.28684023179784\\
93	9.15634900418322\\
94	9.1227782590925\\
95	9.43198676088103\\
96	9.61358381352794\\
97	9.61358381352794\\
98	9.61358381352794\\
99	9.61358381352794\\
100	9.61358381352794\\
100	10.565167509212\\
99	10.565167509212\\
98	10.565167509212\\
97	10.565167509212\\
96	10.565167509212\\
95	10.6105968966435\\
94	10.7383370317039\\
93	10.7232825906638\\
92	10.6668998900982\\
91	10.6602371576912\\
90	10.6200356628829\\
89	10.5987598443063\\
88	10.6079765829195\\
87	10.5968724617245\\
86	10.5887103216956\\
85	10.5842606568668\\
84	10.5707754376625\\
83	10.5962923315921\\
82	10.5009067269846\\
81	10.9769911421028\\
80	12.3245379829578\\
79	12.3182975133657\\
78	12.3127699843567\\
77	12.635021891415\\
76	12.6207798933615\\
75	12.676022677183\\
74	12.8277699000917\\
73	12.451844123348\\
72	12.2111889621293\\
71	12.3051377756045\\
70	12.5250745647911\\
69	11.8227784793866\\
68	10.6246161966516\\
67	10.2350271850612\\
66	10.3061761062047\\
65	8.07179702841007\\
64	6.3700306084957\\
63	4.86499331616408\\
62	3.51319968873645\\
61	-1.16346315885549\\
60	-6.07608667553577\\
59	-7.12284659830572\\
58	-7.78019080647087\\
57	-8.31880329371318\\
56	-8.77968871266303\\
55	-9.14977591922991\\
54	-9.49285605982691\\
53	-9.78871090530967\\
52	-10.0812332928205\\
51	-10.3741450164754\\
50	-10.6248844554837\\
49	-10.8775057306434\\
48	-11.1365345344176\\
47	-11.3675285245095\\
46	-11.5689793881613\\
45	-11.7616304105832\\
44	-11.9579126927148\\
43	-12.1577755112148\\
42	-12.3487937637776\\
41	-12.5323558217449\\
40	-12.7000494507308\\
39	-12.8576266271621\\
38	-13.0108396861857\\
37	-13.148761339257\\
36	-13.3002281596692\\
35	-13.4663566336343\\
34	-13.5942436560897\\
33	-13.7151926833716\\
32	-13.8323361433791\\
31	-13.9464620574989\\
30	-14.075266705828\\
29	-14.2079006261223\\
28	-14.3388422363656\\
27	-14.4603277539029\\
26	-14.5499246928511\\
25	-14.6615530304562\\
24	-14.7877514273156\\
23	-14.8946870399048\\
22	-15.0009014780539\\
21	-15.1145192174994\\
20	-15.2019809950932\\
19	-15.2540537738628\\
18	-15.2863716591393\\
17	-15.3053112479041\\
16	-15.3237195601527\\
15	-15.3375720850073\\
14	-15.3427409912914\\
13	-15.3668450270654\\
12	-15.3759535859507\\
11	-15.3822254631149\\
10	-15.4098086814154\\
9	-15.4308794038657\\
8	-15.4388948954596\\
7	-15.4780545190018\\
6	-15.5261741163602\\
5	-15.5631192610321\\
4	-15.6008033091517\\
3	-15.6651604364093\\
2	-15.7831413019799\\
1	-15.8829506033992\\
}--cycle;

\addplot [color=white!55!mycolor1,solid,forget plot]
  table[row sep=crcr]{%
1	-15.9681846010789\\
2	-15.9223640611152\\
3	-15.8578471991556\\
4	-15.8128933421292\\
5	-15.7815167936868\\
6	-15.7622646328381\\
7	-15.731495143117\\
8	-15.6648347091568\\
9	-15.6134893662733\\
10	-15.6037529947446\\
11	-15.5719291979671\\
12	-15.5457656522361\\
13	-15.5493668605805\\
14	-15.5234912138058\\
15	-15.5018209936012\\
16	-15.4943621660078\\
17	-15.4720041556167\\
18	-15.4585722712799\\
19	-15.4421502830984\\
20	-15.3552643129143\\
21	-15.2640195458174\\
22	-15.2039076340396\\
23	-15.1313885302101\\
24	-15.0370110095517\\
25	-14.93722676607\\
26	-14.8554229231605\\
27	-14.7433829356225\\
28	-14.6581647678332\\
29	-14.590979759751\\
30	-14.4942070047855\\
31	-14.3970046457896\\
32	-14.3091246587774\\
33	-14.2238684977014\\
34	-14.1077569519555\\
35	-14.0019844940087\\
36	-13.9216995040953\\
37	-13.8155878170787\\
38	-13.6872739260006\\
39	-13.5750815192958\\
40	-13.4677825507212\\
41	-13.3414672465585\\
42	-13.2502647210338\\
43	-13.1217991921758\\
44	-12.9748655191938\\
45	-12.8401501904771\\
46	-12.6922258403686\\
47	-12.5550380823679\\
48	-12.4211956838528\\
49	-12.2688126005735\\
50	-12.1097310262218\\
51	-11.9470448870979\\
52	-11.7787453390275\\
53	-11.5960803427397\\
54	-11.4090622045821\\
55	-11.2336005615662\\
56	-11.0470354011729\\
57	-10.8604466145872\\
58	-10.6476877289129\\
59	-10.4371722210309\\
60	-10.3054362807062\\
61	-11.3328014769001\\
62	-12.1184973185311\\
63	-11.3022535985647\\
64	-10.7150722622284\\
65	-9.04939443870921\\
66	-8.30224054138511\\
67	-7.66655285597978\\
68	-6.1687616892046\\
69	-4.34190355772136\\
70	-3.42953668785212\\
71	-2.7601441398133\\
72	-1.08104131335988\\
73	1.19456433942269\\
74	2.00769231067421\\
75	2.11131017178945\\
76	2.36913500940107\\
77	2.61577429685034\\
78	2.79652242237423\\
79	3.291989244586\\
80	4.03014325731437\\
81	6.79063936803398\\
82	8.90478683458204\\
83	9.26475596812976\\
84	9.49957753439859\\
85	9.51467016210989\\
86	9.49916315878987\\
87	9.47271843271242\\
88	9.43949335787496\\
89	9.46688218903189\\
90	9.40574530680893\\
91	9.30295708470338\\
92	9.28684023179784\\
93	9.15634900418322\\
94	9.1227782590925\\
95	9.43198676088103\\
96	9.61358381352794\\
97	9.61358381352794\\
98	9.61358381352794\\
99	9.61358381352794\\
100	9.61358381352794\\
};
\addplot [color=white!55!mycolor1,solid,forget plot]
  table[row sep=crcr]{%
1	-15.8829506033992\\
2	-15.7831413019799\\
3	-15.6651604364093\\
4	-15.6008033091517\\
5	-15.5631192610321\\
6	-15.5261741163602\\
7	-15.4780545190018\\
8	-15.4388948954596\\
9	-15.4308794038657\\
10	-15.4098086814154\\
11	-15.3822254631149\\
12	-15.3759535859507\\
13	-15.3668450270654\\
14	-15.3427409912914\\
15	-15.3375720850073\\
16	-15.3237195601527\\
17	-15.3053112479041\\
18	-15.2863716591393\\
19	-15.2540537738628\\
20	-15.2019809950932\\
21	-15.1145192174994\\
22	-15.0009014780539\\
23	-14.8946870399048\\
24	-14.7877514273156\\
25	-14.6615530304562\\
26	-14.5499246928511\\
27	-14.4603277539029\\
28	-14.3388422363656\\
29	-14.2079006261223\\
30	-14.075266705828\\
31	-13.9464620574989\\
32	-13.8323361433791\\
33	-13.7151926833716\\
34	-13.5942436560897\\
35	-13.4663566336343\\
36	-13.3002281596692\\
37	-13.148761339257\\
38	-13.0108396861857\\
39	-12.8576266271621\\
40	-12.7000494507308\\
41	-12.5323558217449\\
42	-12.3487937637776\\
43	-12.1577755112148\\
44	-11.9579126927148\\
45	-11.7616304105832\\
46	-11.5689793881613\\
47	-11.3675285245095\\
48	-11.1365345344176\\
49	-10.8775057306434\\
50	-10.6248844554837\\
51	-10.3741450164754\\
52	-10.0812332928205\\
53	-9.78871090530967\\
54	-9.49285605982691\\
55	-9.14977591922991\\
56	-8.77968871266303\\
57	-8.31880329371318\\
58	-7.78019080647087\\
59	-7.12284659830572\\
60	-6.07608667553577\\
61	-1.16346315885549\\
62	3.51319968873645\\
63	4.86499331616408\\
64	6.3700306084957\\
65	8.07179702841007\\
66	10.3061761062047\\
67	10.2350271850612\\
68	10.6246161966516\\
69	11.8227784793866\\
70	12.5250745647911\\
71	12.3051377756045\\
72	12.2111889621293\\
73	12.451844123348\\
74	12.8277699000917\\
75	12.676022677183\\
76	12.6207798933615\\
77	12.635021891415\\
78	12.3127699843567\\
79	12.3182975133657\\
80	12.3245379829578\\
81	10.9769911421028\\
82	10.5009067269846\\
83	10.5962923315921\\
84	10.5707754376625\\
85	10.5842606568668\\
86	10.5887103216956\\
87	10.5968724617245\\
88	10.6079765829195\\
89	10.5987598443063\\
90	10.6200356628829\\
91	10.6602371576912\\
92	10.6668998900982\\
93	10.7232825906638\\
94	10.7383370317039\\
95	10.6105968966435\\
96	10.565167509212\\
97	10.565167509212\\
98	10.565167509212\\
99	10.565167509212\\
100	10.565167509212\\
};
\addplot [color=mycolor1,solid,line width = 1.0pt,mark=square]
  table[row sep=crcr]{%
1	-15.9255676022391\\
2	-15.8527526815476\\
3	-15.7615038177824\\
4	-15.7068483256405\\
5	-15.6723180273594\\
6	-15.6442193745991\\
7	-15.6047748310594\\
8	-15.5518648023082\\
9	-15.5221843850695\\
10	-15.50678083808\\
11	-15.477077330541\\
12	-15.4608596190934\\
13	-15.4581059438229\\
14	-15.4331161025486\\
15	-15.4196965393042\\
16	-15.4090408630802\\
17	-15.3886577017604\\
18	-15.3724719652096\\
19	-15.3481020284806\\
20	-15.2786226540037\\
21	-15.1892693816584\\
22	-15.1024045560467\\
23	-15.0130377850575\\
24	-14.9123812184337\\
25	-14.7993898982631\\
26	-14.7026738080058\\
27	-14.6018553447627\\
28	-14.4985035020994\\
29	-14.3994401929366\\
30	-14.2847368553067\\
31	-14.1717333516443\\
32	-14.0707304010782\\
33	-13.9695305905365\\
34	-13.8510003040226\\
35	-13.7341705638215\\
36	-13.6109638318823\\
37	-13.4821745781679\\
38	-13.3490568060931\\
39	-13.216354073229\\
40	-13.083916000726\\
41	-12.9369115341517\\
42	-12.7995292424057\\
43	-12.6397873516953\\
44	-12.4663891059543\\
45	-12.3008903005302\\
46	-12.1306026142649\\
47	-11.9612833034387\\
48	-11.7788651091352\\
49	-11.5731591656085\\
50	-11.3673077408527\\
51	-11.1605949517867\\
52	-10.929989315924\\
53	-10.6923956240247\\
54	-10.4509591322045\\
55	-10.1916882403981\\
56	-9.91336205691797\\
57	-9.5896249541502\\
58	-9.2139392676919\\
59	-8.78000940966829\\
60	-8.19076147812099\\
61	-6.24813231787781\\
62	-4.30264881489732\\
63	-3.21863014120032\\
64	-2.17252082686637\\
65	-0.488798705149571\\
66	1.00196778240979\\
67	1.28423716454072\\
68	2.2279272537235\\
69	3.7404374608326\\
70	4.54776893846947\\
71	4.77249681789559\\
72	5.56507382438473\\
73	6.82320423138534\\
74	7.41773110538295\\
75	7.3936664244862\\
76	7.4949574513813\\
77	7.62539809413267\\
78	7.55464620336546\\
79	7.80514337897584\\
80	8.1773406201361\\
81	8.88381525506839\\
82	9.70284678078331\\
83	9.93052414986094\\
84	10.0351764860306\\
85	10.0494654094884\\
86	10.0439367402427\\
87	10.0347954472185\\
88	10.0237349703972\\
89	10.0328210166691\\
90	10.0128904848459\\
91	9.98159712119727\\
92	9.97687006094801\\
93	9.93981579742351\\
94	9.93055764539822\\
95	10.0212918287622\\
96	10.08937566137\\
97	10.08937566137\\
98	10.08937566137\\
99	10.08937566137\\
100	10.08937566137\\
};
\addlegendentry{C-MORE Nuc. Norm}

\addplot[area legend,solid,fill=mycolor2,opacity=1.000000e-01,draw=none,forget plot]
table[row sep=crcr] {%
x	y\\
1	-13.0093233198995\\
2	-13.0699283968992\\
3	-13.168102924309\\
4	-13.25501886154\\
5	-13.3770557105259\\
6	-13.4702366920475\\
7	-13.5544976807834\\
8	-13.5688292938794\\
9	-13.5704057064082\\
10	-13.5284574121554\\
11	-13.5150600798025\\
12	-13.5013373942813\\
13	-13.4747664673337\\
14	-13.4574505957264\\
15	-13.4427396708535\\
16	-13.4585058302509\\
17	-13.4308619050974\\
18	-13.4016561023494\\
19	-13.3861045535946\\
20	-13.374915824943\\
21	-13.2604097701112\\
22	-13.1570485105217\\
23	-13.0693208589553\\
24	-12.9917609101417\\
25	-12.8752699689822\\
26	-12.7903819656445\\
27	-12.6956981765132\\
28	-12.6044847833274\\
29	-12.5426162261595\\
30	-12.4500913889141\\
31	-12.3206290372721\\
32	-12.208733215278\\
33	-12.1667798668277\\
34	-12.0754718039909\\
35	-11.972960549013\\
36	-11.9000781644055\\
37	-11.7960662574513\\
38	-11.7073036910642\\
39	-11.6231034095974\\
40	-11.5229957141268\\
41	-11.4535862026712\\
42	-11.3203362792201\\
43	-11.2459047618642\\
44	-11.1768073987171\\
45	-11.0808524752649\\
46	-10.9882341317043\\
47	-10.8868744109134\\
48	-10.790742078185\\
49	-10.7077031098216\\
50	-10.6283556036508\\
51	-10.5402774050154\\
52	-10.439581551006\\
53	-10.3390744284692\\
54	-10.2819274653387\\
55	-10.2017256759679\\
56	-10.0786151651484\\
57	-10.0052587326235\\
58	-9.9398189046773\\
59	-9.84401844432278\\
60	-9.77689129583759\\
61	-9.70700046755686\\
62	-9.62900381634818\\
63	-9.55608644204551\\
64	-9.48404870340319\\
65	-9.39530967996337\\
66	-9.33255096676027\\
67	-9.25655134008261\\
68	-9.16846265128628\\
69	-9.14880932678653\\
70	-9.03692183289189\\
71	-8.99093432891906\\
72	-8.92408240436307\\
73	-8.87935313613355\\
74	-8.83093527085826\\
75	-8.74159392891783\\
76	-8.71993240684156\\
77	-8.67448682624136\\
78	-8.61923937001541\\
79	-8.60008336748339\\
80	-8.55632225971454\\
81	-8.50271452320554\\
82	-8.4678168525144\\
83	-8.432475809912\\
84	-8.42861624061416\\
85	-8.36248626498229\\
86	-8.34555396152373\\
87	-8.29341682899329\\
88	-8.2706518310779\\
89	-8.23146305260459\\
90	-8.19611015981461\\
91	-8.19129651737881\\
92	-8.12311003288404\\
93	-8.11940326478939\\
94	-8.12536160646154\\
95	-8.09226065163175\\
96	-8.07543901068241\\
97	-8.04884433450759\\
98	-8.02298708558375\\
99	-8.02535601592429\\
100	-8.02535601592429\\
100	-6.58046777795557\\
99	-6.58046777795557\\
98	-6.59488942862289\\
97	-6.59006766688653\\
96	-6.60765302831959\\
95	-6.62954573567894\\
94	-6.65006362001673\\
93	-6.64310898727998\\
92	-6.63827349103688\\
91	-6.67857138605831\\
90	-6.69277782629426\\
89	-6.69809806526489\\
88	-6.73980650730608\\
87	-6.7393328499844\\
86	-6.78010635787696\\
85	-6.7990286434263\\
84	-6.82100048797313\\
83	-6.8626507205978\\
82	-6.88938063499297\\
81	-6.95273764254415\\
80	-6.98553353363493\\
79	-7.02349351957975\\
78	-7.08932593239333\\
77	-7.13039300343793\\
76	-7.19483513694402\\
75	-7.24067105602368\\
74	-7.32344195625018\\
73	-7.37870747701521\\
72	-7.45841996801815\\
71	-7.52539280161199\\
70	-7.60869845455595\\
69	-7.69225101001094\\
68	-7.77285820966874\\
67	-7.86428013281205\\
66	-7.93677753903439\\
65	-8.03598731863427\\
64	-8.13200115332712\\
63	-8.20911547878492\\
62	-8.30948794508434\\
61	-8.40770390913123\\
60	-8.52469043034843\\
59	-8.63342336722362\\
58	-8.74768200115894\\
57	-8.84750837957862\\
56	-8.94825650808802\\
55	-9.08014976479898\\
54	-9.18835879218385\\
53	-9.31427419652767\\
52	-9.4236882877969\\
51	-9.54633188849379\\
50	-9.6424570852914\\
49	-9.75848934378318\\
48	-9.89931149753373\\
47	-10.009603649416\\
46	-10.1256402597366\\
45	-10.245362594074\\
44	-10.3683927425022\\
43	-10.4779788784247\\
42	-10.5933728951894\\
41	-10.701542687769\\
40	-10.8229056941158\\
39	-10.931035787037\\
38	-11.0389222357352\\
37	-11.138400700256\\
36	-11.3078464624635\\
35	-11.3828181186863\\
34	-11.5123888888422\\
33	-11.6133089293896\\
32	-11.7341641405029\\
31	-11.840862689805\\
30	-11.9762275246306\\
29	-12.1159620796656\\
28	-12.2033343422051\\
27	-12.3131923329921\\
26	-12.4248897174479\\
25	-12.5521837304323\\
24	-12.6903602494109\\
23	-12.7515285427362\\
22	-12.8924367668367\\
21	-13.0231739459018\\
20	-13.1027492732071\\
19	-13.145850081978\\
18	-13.1729273640114\\
17	-13.2146250132171\\
16	-13.2187795481802\\
15	-13.269747052581\\
14	-13.2479478601829\\
13	-13.2562636124247\\
12	-13.3001120226899\\
11	-13.3172091215739\\
10	-13.3439259587994\\
9	-13.3603458338341\\
8	-13.3709349146314\\
7	-13.3677403798137\\
6	-13.361180810945\\
5	-13.2880984895006\\
4	-13.2051813007078\\
3	-13.0956851331818\\
2	-12.9544200102255\\
1	-12.9609797795641\\
}--cycle;
\addplot [color=white!55!mycolor2,solid,forget plot]
  table[row sep=crcr]{%
1	-13.0093233198995\\
2	-13.0699283968992\\
3	-13.168102924309\\
4	-13.25501886154\\
5	-13.3770557105259\\
6	-13.4702366920475\\
7	-13.5544976807834\\
8	-13.5688292938794\\
9	-13.5704057064082\\
10	-13.5284574121554\\
11	-13.5150600798025\\
12	-13.5013373942813\\
13	-13.4747664673337\\
14	-13.4574505957264\\
15	-13.4427396708535\\
16	-13.4585058302509\\
17	-13.4308619050974\\
18	-13.4016561023494\\
19	-13.3861045535946\\
20	-13.374915824943\\
21	-13.2604097701112\\
22	-13.1570485105217\\
23	-13.0693208589553\\
24	-12.9917609101417\\
25	-12.8752699689822\\
26	-12.7903819656445\\
27	-12.6956981765132\\
28	-12.6044847833274\\
29	-12.5426162261595\\
30	-12.4500913889141\\
31	-12.3206290372721\\
32	-12.208733215278\\
33	-12.1667798668277\\
34	-12.0754718039909\\
35	-11.972960549013\\
36	-11.9000781644055\\
37	-11.7960662574513\\
38	-11.7073036910642\\
39	-11.6231034095974\\
40	-11.5229957141268\\
41	-11.4535862026712\\
42	-11.3203362792201\\
43	-11.2459047618642\\
44	-11.1768073987171\\
45	-11.0808524752649\\
46	-10.9882341317043\\
47	-10.8868744109134\\
48	-10.790742078185\\
49	-10.7077031098216\\
50	-10.6283556036508\\
51	-10.5402774050154\\
52	-10.439581551006\\
53	-10.3390744284692\\
54	-10.2819274653387\\
55	-10.2017256759679\\
56	-10.0786151651484\\
57	-10.0052587326235\\
58	-9.9398189046773\\
59	-9.84401844432278\\
60	-9.77689129583759\\
61	-9.70700046755686\\
62	-9.62900381634818\\
63	-9.55608644204551\\
64	-9.48404870340319\\
65	-9.39530967996337\\
66	-9.33255096676027\\
67	-9.25655134008261\\
68	-9.16846265128628\\
69	-9.14880932678653\\
70	-9.03692183289189\\
71	-8.99093432891906\\
72	-8.92408240436307\\
73	-8.87935313613355\\
74	-8.83093527085826\\
75	-8.74159392891783\\
76	-8.71993240684156\\
77	-8.67448682624136\\
78	-8.61923937001541\\
79	-8.60008336748339\\
80	-8.55632225971454\\
81	-8.50271452320554\\
82	-8.4678168525144\\
83	-8.432475809912\\
84	-8.42861624061416\\
85	-8.36248626498229\\
86	-8.34555396152373\\
87	-8.29341682899329\\
88	-8.2706518310779\\
89	-8.23146305260459\\
90	-8.19611015981461\\
91	-8.19129651737881\\
92	-8.12311003288404\\
93	-8.11940326478939\\
94	-8.12536160646154\\
95	-8.09226065163175\\
96	-8.07543901068241\\
97	-8.04884433450759\\
98	-8.02298708558375\\
99	-8.02535601592429\\
100	-8.02535601592429\\
};
\addplot [color=white!55!mycolor2,solid,forget plot]
  table[row sep=crcr]{%
1	-12.9609797795641\\
2	-12.9544200102255\\
3	-13.0956851331818\\
4	-13.2051813007078\\
5	-13.2880984895006\\
6	-13.361180810945\\
7	-13.3677403798137\\
8	-13.3709349146314\\
9	-13.3603458338341\\
10	-13.3439259587994\\
11	-13.3172091215739\\
12	-13.3001120226899\\
13	-13.2562636124247\\
14	-13.2479478601829\\
15	-13.269747052581\\
16	-13.2187795481802\\
17	-13.2146250132171\\
18	-13.1729273640114\\
19	-13.145850081978\\
20	-13.1027492732071\\
21	-13.0231739459018\\
22	-12.8924367668367\\
23	-12.7515285427362\\
24	-12.6903602494109\\
25	-12.5521837304323\\
26	-12.4248897174479\\
27	-12.3131923329921\\
28	-12.2033343422051\\
29	-12.1159620796656\\
30	-11.9762275246306\\
31	-11.840862689805\\
32	-11.7341641405029\\
33	-11.6133089293896\\
34	-11.5123888888422\\
35	-11.3828181186863\\
36	-11.3078464624635\\
37	-11.138400700256\\
38	-11.0389222357352\\
39	-10.931035787037\\
40	-10.8229056941158\\
41	-10.701542687769\\
42	-10.5933728951894\\
43	-10.4779788784247\\
44	-10.3683927425022\\
45	-10.245362594074\\
46	-10.1256402597366\\
47	-10.009603649416\\
48	-9.89931149753373\\
49	-9.75848934378318\\
50	-9.6424570852914\\
51	-9.54633188849379\\
52	-9.4236882877969\\
53	-9.31427419652767\\
54	-9.18835879218385\\
55	-9.08014976479898\\
56	-8.94825650808802\\
57	-8.84750837957862\\
58	-8.74768200115894\\
59	-8.63342336722362\\
60	-8.52469043034843\\
61	-8.40770390913123\\
62	-8.30948794508434\\
63	-8.20911547878492\\
64	-8.13200115332712\\
65	-8.03598731863427\\
66	-7.93677753903439\\
67	-7.86428013281205\\
68	-7.77285820966874\\
69	-7.69225101001094\\
70	-7.60869845455595\\
71	-7.52539280161199\\
72	-7.45841996801815\\
73	-7.37870747701521\\
74	-7.32344195625018\\
75	-7.24067105602368\\
76	-7.19483513694402\\
77	-7.13039300343793\\
78	-7.08932593239333\\
79	-7.02349351957975\\
80	-6.98553353363493\\
81	-6.95273764254415\\
82	-6.88938063499297\\
83	-6.8626507205978\\
84	-6.82100048797313\\
85	-6.7990286434263\\
86	-6.78010635787696\\
87	-6.7393328499844\\
88	-6.73980650730608\\
89	-6.69809806526489\\
90	-6.69277782629426\\
91	-6.67857138605831\\
92	-6.63827349103688\\
93	-6.64310898727998\\
94	-6.65006362001673\\
95	-6.62954573567894\\
96	-6.60765302831959\\
97	-6.59006766688653\\
98	-6.59488942862289\\
99	-6.58046777795557\\
100	-6.58046777795557\\
};
\addplot [color=mycolor2,solid,line width = 1.0pt,mark=diamond]
  table[row sep=crcr]{%
1	-12.9851515497318\\
2	-13.0121742035624\\
3	-13.1318940287454\\
4	-13.2301000811239\\
5	-13.3325771000132\\
6	-13.4157087514963\\
7	-13.4611190302985\\
8	-13.4698821042554\\
9	-13.4653757701211\\
10	-13.4361916854774\\
11	-13.4161346006882\\
12	-13.4007247084856\\
13	-13.3655150398792\\
14	-13.3526992279546\\
15	-13.3562433617173\\
16	-13.3386426892155\\
17	-13.3227434591572\\
18	-13.2872917331804\\
19	-13.2659773177863\\
20	-13.2388325490751\\
21	-13.1417918580065\\
22	-13.0247426386792\\
23	-12.9104247008458\\
24	-12.8410605797763\\
25	-12.7137268497072\\
26	-12.6076358415462\\
27	-12.5044452547526\\
28	-12.4039095627662\\
29	-12.3292891529125\\
30	-12.2131594567724\\
31	-12.0807458635385\\
32	-11.9714486778904\\
33	-11.8900443981086\\
34	-11.7939303464165\\
35	-11.6778893338496\\
36	-11.6039623134345\\
37	-11.4672334788536\\
38	-11.3731129633997\\
39	-11.2770695983172\\
40	-11.1729507041213\\
41	-11.0775644452201\\
42	-10.9568545872048\\
43	-10.8619418201444\\
44	-10.7726000706096\\
45	-10.6631075346694\\
46	-10.5569371957204\\
47	-10.4482390301647\\
48	-10.3450267878594\\
49	-10.2330962268024\\
50	-10.1354063444711\\
51	-10.0433046467546\\
52	-9.93163491940147\\
53	-9.82667431249845\\
54	-9.7351431287613\\
55	-9.64093772038343\\
56	-9.51343583661824\\
57	-9.42638355610104\\
58	-9.34375045291812\\
59	-9.2387209057732\\
60	-9.15079086309301\\
61	-9.05735218834404\\
62	-8.96924588071626\\
63	-8.88260096041521\\
64	-8.80802492836516\\
65	-8.71564849929882\\
66	-8.63466425289733\\
67	-8.56041573644733\\
68	-8.47066043047751\\
69	-8.42053016839873\\
70	-8.32281014372392\\
71	-8.25816356526552\\
72	-8.19125118619061\\
73	-8.12903030657438\\
74	-8.07718861355422\\
75	-7.99113249247076\\
76	-7.95738377189279\\
77	-7.90243991483965\\
78	-7.85428265120437\\
79	-7.81178844353157\\
80	-7.77092789667473\\
81	-7.72772608287484\\
82	-7.67859874375368\\
83	-7.6475632652549\\
84	-7.62480836429364\\
85	-7.5807574542043\\
86	-7.56283015970034\\
87	-7.51637483948884\\
88	-7.50522916919199\\
89	-7.46478055893474\\
90	-7.44444399305444\\
91	-7.43493395171856\\
92	-7.38069176196046\\
93	-7.38125612603469\\
94	-7.38771261323913\\
95	-7.36090319365534\\
96	-7.341546019501\\
97	-7.31945600069706\\
98	-7.30893825710332\\
99	-7.30291189693993\\
100	-7.30291189693993\\
};
\addlegendentry{C-MORE Ridge+PCA}

\addplot[area legend,solid,fill=mycolor3,opacity=1.000000e-01,draw=none,forget plot]
table[row sep=crcr] {%
x	y\\
1	-16.0184618623908\\
2	-15.9945318241882\\
3	-15.9938169087898\\
4	-16.0243655316737\\
5	-15.9989008377545\\
6	-16.0371764320209\\
7	-16.0270416018936\\
8	-16.0331496521698\\
9	-15.9771557617413\\
10	-16.0255621800201\\
11	-16.0239470611862\\
12	-15.9990232017427\\
13	-16.0177872761098\\
14	-16.0475229119188\\
15	-16.0306742568485\\
16	-16.0347446517718\\
17	-16.0246404090969\\
18	-16.0295621264705\\
19	-16.0067772075378\\
20	-16.0201891461582\\
21	-15.9793703867053\\
22	-15.9536088149754\\
23	-15.9119592035924\\
24	-15.8980100980858\\
25	-15.8371387324781\\
26	-15.7955835312459\\
27	-15.7813600425951\\
28	-15.7320111369447\\
29	-15.6901332654854\\
30	-15.6515355382475\\
31	-15.6266722438896\\
32	-15.5885175908668\\
33	-15.5314826584695\\
34	-15.5027470263149\\
35	-15.4312737253937\\
36	-15.4300557128552\\
37	-15.4039880600234\\
38	-15.3573813565423\\
39	-15.3208024262549\\
40	-15.3216782017977\\
41	-15.2329992522009\\
42	-15.2029371591391\\
43	-15.1617596296695\\
44	-15.1752874166643\\
45	-15.1114262071126\\
46	-15.0727394520051\\
47	-14.9886433249535\\
48	-14.9974342319007\\
49	-14.9451077383676\\
50	-14.9432603634318\\
51	-14.8741393252366\\
52	-14.8082846474907\\
53	-14.808690350805\\
54	-14.7697478448459\\
55	-14.7136088148511\\
56	-14.6761622317772\\
57	-14.6418981891094\\
58	-14.6069786049106\\
59	-14.5387323147523\\
60	-14.5319935424757\\
61	-14.4941780214831\\
62	-14.4134953069593\\
63	-14.4150639469363\\
64	-14.322966893672\\
65	-14.2841620419412\\
66	-14.2687185213752\\
67	-14.2220368069364\\
68	-14.1652452920552\\
69	-14.0864729253074\\
70	-14.0462029806996\\
71	-14.0437598836486\\
72	-13.9476060668782\\
73	-13.910423072961\\
74	-13.8519579914495\\
75	-13.8613592479022\\
76	-13.7434199157594\\
77	-13.7043204268747\\
78	-13.5703259571959\\
79	-13.6051187212182\\
80	-13.567977512883\\
81	-13.4904745379754\\
82	-13.4210577613649\\
83	-13.3757317415863\\
84	-13.3380126179876\\
85	-13.2533902394187\\
86	-13.2045988050021\\
87	-13.1655524845174\\
88	-13.0945153926952\\
89	-13.0510911892412\\
90	-12.9750639953836\\
91	-12.8812759266548\\
92	-12.863487887342\\
93	-12.7765565266703\\
94	-12.7031988948418\\
95	-12.6096702238807\\
96	-12.595631928039\\
97	-12.5043966046002\\
98	-12.4498768605842\\
99	-12.3735230080325\\
100	-12.3735230080325\\
100	-11.0821310939859\\
99	-11.0821310939859\\
98	-11.1731940238411\\
97	-11.3049621099047\\
96	-11.3794547952225\\
95	-11.5184086360456\\
94	-11.6317747359599\\
93	-11.7164252818235\\
92	-11.8220726629532\\
91	-11.9181012858725\\
90	-12.0120246212212\\
89	-12.0767991641684\\
88	-12.1476495970134\\
87	-12.2544605697836\\
86	-12.3142373763288\\
85	-12.4427636889029\\
84	-12.5005578364416\\
83	-12.6167062970115\\
82	-12.6135997468628\\
81	-12.7578026963016\\
80	-12.8100832270934\\
79	-12.8441919344345\\
78	-12.9414301012383\\
77	-13.0193699853913\\
76	-13.0836060041808\\
75	-13.1470134878631\\
74	-13.229912330532\\
73	-13.3062877862553\\
72	-13.348953347201\\
71	-13.432807534544\\
70	-13.4759130574596\\
69	-13.5717131042267\\
68	-13.6218988386581\\
67	-13.6681696381133\\
66	-13.7633427602914\\
65	-13.7568066790105\\
64	-13.8438144462215\\
63	-13.9124554213627\\
62	-13.9624550455674\\
61	-14.0463368225002\\
60	-14.1098234454961\\
59	-14.1120241507592\\
58	-14.17419616485\\
57	-14.2460311916392\\
56	-14.2955272900872\\
55	-14.3555636258494\\
54	-14.3821861628203\\
53	-14.4638861843168\\
52	-14.4966618243873\\
51	-14.5485787105068\\
50	-14.6343035634117\\
49	-14.6552192808454\\
48	-14.7163551468882\\
47	-14.7671117036185\\
46	-14.8026181170649\\
45	-14.8423150180051\\
44	-14.871102689872\\
43	-14.9253914229813\\
42	-14.9747345114508\\
41	-15.0313088525499\\
40	-15.0600084898386\\
39	-15.1308247384165\\
38	-15.189205582183\\
37	-15.2153810943333\\
36	-15.249467593862\\
35	-15.3048925270284\\
34	-15.3465874760297\\
33	-15.3892851665216\\
32	-15.4683918483885\\
31	-15.4809670750252\\
30	-15.5207192368968\\
29	-15.5799067722362\\
28	-15.5974386519158\\
27	-15.6366601343455\\
26	-15.6942519342178\\
25	-15.725469001086\\
24	-15.7486710122149\\
23	-15.7933938086135\\
22	-15.8301617355421\\
21	-15.8946631827286\\
20	-15.9157361690995\\
19	-15.9261680810672\\
18	-15.9482836704553\\
17	-15.9107722000449\\
16	-15.9463096514524\\
15	-15.9114870284967\\
14	-15.9446113004905\\
13	-15.9173915519054\\
12	-15.9260245200778\\
11	-15.8908831975905\\
10	-15.9303245475696\\
9	-15.8723868022902\\
8	-15.9201767492677\\
7	-15.9247569343653\\
6	-15.9275188151626\\
5	-15.9153633884466\\
4	-15.9516944749117\\
3	-15.9027894882714\\
2	-15.9001275941428\\
1	-15.9110391300612\\
}--cycle;

\addplot [color=white!55!mycolor3,solid,forget plot]
  table[row sep=crcr]{%
1	-16.0184618623908\\
2	-15.9945318241882\\
3	-15.9938169087898\\
4	-16.0243655316737\\
5	-15.9989008377545\\
6	-16.0371764320209\\
7	-16.0270416018936\\
8	-16.0331496521698\\
9	-15.9771557617413\\
10	-16.0255621800201\\
11	-16.0239470611862\\
12	-15.9990232017427\\
13	-16.0177872761098\\
14	-16.0475229119188\\
15	-16.0306742568485\\
16	-16.0347446517718\\
17	-16.0246404090969\\
18	-16.0295621264705\\
19	-16.0067772075378\\
20	-16.0201891461582\\
21	-15.9793703867053\\
22	-15.9536088149754\\
23	-15.9119592035924\\
24	-15.8980100980858\\
25	-15.8371387324781\\
26	-15.7955835312459\\
27	-15.7813600425951\\
28	-15.7320111369447\\
29	-15.6901332654854\\
30	-15.6515355382475\\
31	-15.6266722438896\\
32	-15.5885175908668\\
33	-15.5314826584695\\
34	-15.5027470263149\\
35	-15.4312737253937\\
36	-15.4300557128552\\
37	-15.4039880600234\\
38	-15.3573813565423\\
39	-15.3208024262549\\
40	-15.3216782017977\\
41	-15.2329992522009\\
42	-15.2029371591391\\
43	-15.1617596296695\\
44	-15.1752874166643\\
45	-15.1114262071126\\
46	-15.0727394520051\\
47	-14.9886433249535\\
48	-14.9974342319007\\
49	-14.9451077383676\\
50	-14.9432603634318\\
51	-14.8741393252366\\
52	-14.8082846474907\\
53	-14.808690350805\\
54	-14.7697478448459\\
55	-14.7136088148511\\
56	-14.6761622317772\\
57	-14.6418981891094\\
58	-14.6069786049106\\
59	-14.5387323147523\\
60	-14.5319935424757\\
61	-14.4941780214831\\
62	-14.4134953069593\\
63	-14.4150639469363\\
64	-14.322966893672\\
65	-14.2841620419412\\
66	-14.2687185213752\\
67	-14.2220368069364\\
68	-14.1652452920552\\
69	-14.0864729253074\\
70	-14.0462029806996\\
71	-14.0437598836486\\
72	-13.9476060668782\\
73	-13.910423072961\\
74	-13.8519579914495\\
75	-13.8613592479022\\
76	-13.7434199157594\\
77	-13.7043204268747\\
78	-13.5703259571959\\
79	-13.6051187212182\\
80	-13.567977512883\\
81	-13.4904745379754\\
82	-13.4210577613649\\
83	-13.3757317415863\\
84	-13.3380126179876\\
85	-13.2533902394187\\
86	-13.2045988050021\\
87	-13.1655524845174\\
88	-13.0945153926952\\
89	-13.0510911892412\\
90	-12.9750639953836\\
91	-12.8812759266548\\
92	-12.863487887342\\
93	-12.7765565266703\\
94	-12.7031988948418\\
95	-12.6096702238807\\
96	-12.595631928039\\
97	-12.5043966046002\\
98	-12.4498768605842\\
99	-12.3735230080325\\
100	-12.3735230080325\\
};
\addplot [color=white!55!mycolor3,solid,forget plot]
  table[row sep=crcr]{%
1	-15.9110391300612\\
2	-15.9001275941428\\
3	-15.9027894882714\\
4	-15.9516944749117\\
5	-15.9153633884466\\
6	-15.9275188151626\\
7	-15.9247569343653\\
8	-15.9201767492677\\
9	-15.8723868022902\\
10	-15.9303245475696\\
11	-15.8908831975905\\
12	-15.9260245200778\\
13	-15.9173915519054\\
14	-15.9446113004905\\
15	-15.9114870284967\\
16	-15.9463096514524\\
17	-15.9107722000449\\
18	-15.9482836704553\\
19	-15.9261680810672\\
20	-15.9157361690995\\
21	-15.8946631827286\\
22	-15.8301617355421\\
23	-15.7933938086135\\
24	-15.7486710122149\\
25	-15.725469001086\\
26	-15.6942519342178\\
27	-15.6366601343455\\
28	-15.5974386519158\\
29	-15.5799067722362\\
30	-15.5207192368968\\
31	-15.4809670750252\\
32	-15.4683918483885\\
33	-15.3892851665216\\
34	-15.3465874760297\\
35	-15.3048925270284\\
36	-15.249467593862\\
37	-15.2153810943333\\
38	-15.189205582183\\
39	-15.1308247384165\\
40	-15.0600084898386\\
41	-15.0313088525499\\
42	-14.9747345114508\\
43	-14.9253914229813\\
44	-14.871102689872\\
45	-14.8423150180051\\
46	-14.8026181170649\\
47	-14.7671117036185\\
48	-14.7163551468882\\
49	-14.6552192808454\\
50	-14.6343035634117\\
51	-14.5485787105068\\
52	-14.4966618243873\\
53	-14.4638861843168\\
54	-14.3821861628203\\
55	-14.3555636258494\\
56	-14.2955272900872\\
57	-14.2460311916392\\
58	-14.17419616485\\
59	-14.1120241507592\\
60	-14.1098234454961\\
61	-14.0463368225002\\
62	-13.9624550455674\\
63	-13.9124554213627\\
64	-13.8438144462215\\
65	-13.7568066790105\\
66	-13.7633427602914\\
67	-13.6681696381133\\
68	-13.6218988386581\\
69	-13.5717131042267\\
70	-13.4759130574596\\
71	-13.432807534544\\
72	-13.348953347201\\
73	-13.3062877862553\\
74	-13.229912330532\\
75	-13.1470134878631\\
76	-13.0836060041808\\
77	-13.0193699853913\\
78	-12.9414301012383\\
79	-12.8441919344345\\
80	-12.8100832270934\\
81	-12.7578026963016\\
82	-12.6135997468628\\
83	-12.6167062970115\\
84	-12.5005578364416\\
85	-12.4427636889029\\
86	-12.3142373763288\\
87	-12.2544605697836\\
88	-12.1476495970134\\
89	-12.0767991641684\\
90	-12.0120246212212\\
91	-11.9181012858725\\
92	-11.8220726629532\\
93	-11.7164252818235\\
94	-11.6317747359599\\
95	-11.5184086360456\\
96	-11.3794547952225\\
97	-11.3049621099047\\
98	-11.1731940238411\\
99	-11.0821310939859\\
100	-11.0821310939859\\
};
\addplot [color=mycolor3,solid,line width = 1.0pt,mark=o]
  table[row sep=crcr]{%
1	-15.964750496226\\
2	-15.9473297091655\\
3	-15.9483031985306\\
4	-15.9880300032927\\
5	-15.9571321131005\\
6	-15.9823476235917\\
7	-15.9758992681294\\
8	-15.9766632007188\\
9	-15.9247712820158\\
10	-15.9779433637948\\
11	-15.9574151293884\\
12	-15.9625238609102\\
13	-15.9675894140076\\
14	-15.9960671062047\\
15	-15.9710806426726\\
16	-15.9905271516121\\
17	-15.9677063045709\\
18	-15.9889228984629\\
19	-15.9664726443025\\
20	-15.9679626576288\\
21	-15.937016784717\\
22	-15.8918852752587\\
23	-15.8526765061029\\
24	-15.8233405551503\\
25	-15.781303866782\\
26	-15.7449177327319\\
27	-15.7090100884703\\
28	-15.6647248944302\\
29	-15.6350200188608\\
30	-15.5861273875721\\
31	-15.5538196594574\\
32	-15.5284547196276\\
33	-15.4603839124955\\
34	-15.4246672511723\\
35	-15.3680831262111\\
36	-15.3397616533586\\
37	-15.3096845771784\\
38	-15.2732934693626\\
39	-15.2258135823357\\
40	-15.1908433458181\\
41	-15.1321540523754\\
42	-15.0888358352949\\
43	-15.0435755263254\\
44	-15.0231950532682\\
45	-14.9768706125588\\
46	-14.937678784535\\
47	-14.877877514286\\
48	-14.8568946893944\\
49	-14.8001635096065\\
50	-14.7887819634217\\
51	-14.7113590178717\\
52	-14.652473235939\\
53	-14.6362882675609\\
54	-14.5759670038331\\
55	-14.5345862203503\\
56	-14.4858447609322\\
57	-14.4439646903743\\
58	-14.3905873848803\\
59	-14.3253782327557\\
60	-14.3209084939859\\
61	-14.2702574219917\\
62	-14.1879751762634\\
63	-14.1637596841495\\
64	-14.0833906699467\\
65	-14.0204843604759\\
66	-14.0160306408333\\
67	-13.9451032225248\\
68	-13.8935720653567\\
69	-13.8290930147671\\
70	-13.7610580190796\\
71	-13.7382837090963\\
72	-13.6482797070396\\
73	-13.6083554296081\\
74	-13.5409351609908\\
75	-13.5041863678826\\
76	-13.4135129599701\\
77	-13.361845206133\\
78	-13.2558780292171\\
79	-13.2246553278263\\
80	-13.1890303699882\\
81	-13.1241386171385\\
82	-13.0173287541139\\
83	-12.9962190192989\\
84	-12.9192852272146\\
85	-12.8480769641608\\
86	-12.7594180906655\\
87	-12.7100065271505\\
88	-12.6210824948543\\
89	-12.5639451767048\\
90	-12.4935443083024\\
91	-12.3996886062637\\
92	-12.3427802751476\\
93	-12.2464909042469\\
94	-12.1674868154008\\
95	-12.0640394299631\\
96	-11.9875433616308\\
97	-11.9046793572524\\
98	-11.8115354422127\\
99	-11.7278270510092\\
100	-11.7278270510092\\
};
\addlegendentry{C-MORE Ridge}

\addplot[area legend,solid,fill=mycolor4,opacity=1.000000e-01,draw=none,forget plot]
table[row sep=crcr] {%
x	y\\
1	-12.742807106613\\
2	-12.7781263481621\\
3	-12.7509538168399\\
4	-12.7918229499615\\
5	-12.742208467704\\
6	-12.7685842268508\\
7	-12.7253555603725\\
8	-12.7489251922435\\
9	-12.70173742811\\
10	-12.7078167901378\\
11	-12.6845834454685\\
12	-12.5771014263945\\
13	-12.5923704568809\\
14	-12.5907713558624\\
15	-12.5324791056998\\
16	-12.4835955368247\\
17	-12.4268409733584\\
18	-12.3708908082743\\
19	-12.3690149086522\\
20	-12.3214321511258\\
21	-12.2394167199806\\
22	-12.2054451791936\\
23	-12.1333686392159\\
24	-12.0906842207653\\
25	-12.0503739264531\\
26	-12.0112990189522\\
27	-11.9933253108454\\
28	-11.8902776655522\\
29	-11.8673752843382\\
30	-11.8298824252295\\
31	-11.7784586822574\\
32	-11.7258912593673\\
33	-11.6882985741659\\
34	-11.644134081009\\
35	-11.6059311104891\\
36	-11.5497250663506\\
37	-11.5211544381375\\
38	-11.4322542736418\\
39	-11.4091022960003\\
40	-11.3945505415398\\
41	-11.3437839962743\\
42	-11.3033866463191\\
43	-11.2640887870096\\
44	-11.2569415167204\\
45	-11.2048784088297\\
46	-11.1265725959097\\
47	-11.1337549128711\\
48	-11.0842717027108\\
49	-11.0380834357875\\
50	-10.9828459047402\\
51	-10.9531773791774\\
52	-10.958007606599\\
53	-10.8631890054094\\
54	-10.8556040769335\\
55	-10.7571520721692\\
56	-10.7791033544436\\
57	-10.7108763040078\\
58	-10.702205490788\\
59	-10.6098875368802\\
60	-10.593618859227\\
61	-10.5499289304827\\
62	-10.5146480650191\\
63	-10.4452372633994\\
64	-10.4184390708009\\
65	-10.3379310113711\\
66	-10.3533501873678\\
67	-10.2895677661221\\
68	-10.2179124403895\\
69	-10.2122239229711\\
70	-10.2128667894158\\
71	-10.1223280510109\\
72	-10.0851255897989\\
73	-10.0513094155237\\
74	-10.0132721372764\\
75	-9.97319384357839\\
76	-9.95578113850692\\
77	-9.9353629317495\\
78	-9.87545717883128\\
79	-9.84456166046916\\
80	-9.77255164435166\\
81	-9.75206197026876\\
82	-9.67588220392205\\
83	-9.63710110289563\\
84	-9.56509668326157\\
85	-9.5073332866977\\
86	-9.51080232644758\\
87	-9.45590090972188\\
88	-9.39142574751477\\
89	-9.36952939793516\\
90	-9.3127174029373\\
91	-9.25905451436759\\
92	-9.22086756115863\\
93	-9.21396258942426\\
94	-9.1289805646204\\
95	-9.12549522647755\\
96	-9.08998961526281\\
97	-9.07729355178516\\
98	-9.00365733726834\\
99	-8.98500096766313\\
100	-8.98500096766313\\
100	-6.23784162342916\\
99	-6.23784162342916\\
98	-6.26881024692568\\
97	-6.29728955892853\\
96	-6.35327349540778\\
95	-6.39829285226513\\
94	-6.42760036060124\\
93	-6.45743268245288\\
92	-6.47993548571967\\
91	-6.54126123810753\\
90	-6.60375437615349\\
89	-6.63887568277398\\
88	-6.67616330109121\\
87	-6.72723954641644\\
86	-6.77581439915657\\
85	-6.85204266131877\\
84	-6.92530044244612\\
83	-6.95899688186275\\
82	-7.02000981742997\\
81	-7.04967924823292\\
80	-7.12054168719292\\
79	-7.15100921213743\\
78	-7.23779838925212\\
77	-7.31469305259487\\
76	-7.38586741262133\\
75	-7.46077671273243\\
74	-7.52545737967378\\
73	-7.61499022708212\\
72	-7.65493221027782\\
71	-7.74372477086094\\
70	-7.84022328414124\\
69	-7.88394775423358\\
68	-7.97635448141528\\
67	-8.04204777563184\\
66	-8.1078972111239\\
65	-8.17063277808929\\
64	-8.27246034018279\\
63	-8.39092326546903\\
62	-8.44130145045453\\
61	-8.51778906361703\\
60	-8.58634832154631\\
59	-8.68620829542006\\
58	-8.70140889940108\\
57	-8.80581504775242\\
56	-8.8353214121586\\
55	-8.91735751274226\\
54	-8.95499442415307\\
53	-9.04820745721821\\
52	-9.16092636816303\\
51	-9.25142071097322\\
50	-9.34093198055917\\
49	-9.37468614039353\\
48	-9.47924905889908\\
47	-9.58047952736302\\
46	-9.63854111628923\\
45	-9.72037734799105\\
44	-9.78595187531235\\
43	-9.84486779669314\\
42	-9.97771575607192\\
41	-10.0330320496484\\
40	-10.118814178002\\
39	-10.2019584369158\\
38	-10.2918299209896\\
37	-10.3575008692123\\
36	-10.4783783813358\\
35	-10.598988995348\\
34	-10.6852647092766\\
33	-10.8050725710055\\
32	-10.8981064559162\\
31	-10.9816087887007\\
30	-11.0552441958852\\
29	-11.0906223253968\\
28	-11.1827299193742\\
27	-11.3125647550138\\
26	-11.3363950525897\\
25	-11.4731642750958\\
24	-11.5093604252914\\
23	-11.603983752825\\
22	-11.6949030791687\\
21	-11.7478715553014\\
20	-11.8577241236514\\
19	-11.9380959799812\\
18	-12.0359403139309\\
17	-12.1045037686339\\
16	-12.2024340412786\\
15	-12.2251687787961\\
14	-12.2800720467666\\
13	-12.3745270056034\\
12	-12.3927834161476\\
11	-12.4532709658004\\
10	-12.5180105026611\\
9	-12.5595296065785\\
8	-12.5439533201763\\
7	-12.5948564479863\\
6	-12.6058325556325\\
5	-12.5786633421242\\
4	-12.6339875054324\\
3	-12.6318061366274\\
2	-12.6637837881086\\
1	-12.6375804154268\\
}--cycle;

\addplot [color=white!55!mycolor4,solid,forget plot]
  table[row sep=crcr]{%
1	-12.742807106613\\
2	-12.7781263481621\\
3	-12.7509538168399\\
4	-12.7918229499615\\
5	-12.742208467704\\
6	-12.7685842268508\\
7	-12.7253555603725\\
8	-12.7489251922435\\
9	-12.70173742811\\
10	-12.7078167901378\\
11	-12.6845834454685\\
12	-12.5771014263945\\
13	-12.5923704568809\\
14	-12.5907713558624\\
15	-12.5324791056998\\
16	-12.4835955368247\\
17	-12.4268409733584\\
18	-12.3708908082743\\
19	-12.3690149086522\\
20	-12.3214321511258\\
21	-12.2394167199806\\
22	-12.2054451791936\\
23	-12.1333686392159\\
24	-12.0906842207653\\
25	-12.0503739264531\\
26	-12.0112990189522\\
27	-11.9933253108454\\
28	-11.8902776655522\\
29	-11.8673752843382\\
30	-11.8298824252295\\
31	-11.7784586822574\\
32	-11.7258912593673\\
33	-11.6882985741659\\
34	-11.644134081009\\
35	-11.6059311104891\\
36	-11.5497250663506\\
37	-11.5211544381375\\
38	-11.4322542736418\\
39	-11.4091022960003\\
40	-11.3945505415398\\
41	-11.3437839962743\\
42	-11.3033866463191\\
43	-11.2640887870096\\
44	-11.2569415167204\\
45	-11.2048784088297\\
46	-11.1265725959097\\
47	-11.1337549128711\\
48	-11.0842717027108\\
49	-11.0380834357875\\
50	-10.9828459047402\\
51	-10.9531773791774\\
52	-10.958007606599\\
53	-10.8631890054094\\
54	-10.8556040769335\\
55	-10.7571520721692\\
56	-10.7791033544436\\
57	-10.7108763040078\\
58	-10.702205490788\\
59	-10.6098875368802\\
60	-10.593618859227\\
61	-10.5499289304827\\
62	-10.5146480650191\\
63	-10.4452372633994\\
64	-10.4184390708009\\
65	-10.3379310113711\\
66	-10.3533501873678\\
67	-10.2895677661221\\
68	-10.2179124403895\\
69	-10.2122239229711\\
70	-10.2128667894158\\
71	-10.1223280510109\\
72	-10.0851255897989\\
73	-10.0513094155237\\
74	-10.0132721372764\\
75	-9.97319384357839\\
76	-9.95578113850692\\
77	-9.9353629317495\\
78	-9.87545717883128\\
79	-9.84456166046916\\
80	-9.77255164435166\\
81	-9.75206197026876\\
82	-9.67588220392205\\
83	-9.63710110289563\\
84	-9.56509668326157\\
85	-9.5073332866977\\
86	-9.51080232644758\\
87	-9.45590090972188\\
88	-9.39142574751477\\
89	-9.36952939793516\\
90	-9.3127174029373\\
91	-9.25905451436759\\
92	-9.22086756115863\\
93	-9.21396258942426\\
94	-9.1289805646204\\
95	-9.12549522647755\\
96	-9.08998961526281\\
97	-9.07729355178516\\
98	-9.00365733726834\\
99	-8.98500096766313\\
100	-8.98500096766313\\
};
\addplot [color=white!55!mycolor4,solid,forget plot]
  table[row sep=crcr]{%
1	-12.6375804154268\\
2	-12.6637837881086\\
3	-12.6318061366274\\
4	-12.6339875054324\\
5	-12.5786633421242\\
6	-12.6058325556325\\
7	-12.5948564479863\\
8	-12.5439533201763\\
9	-12.5595296065785\\
10	-12.5180105026611\\
11	-12.4532709658004\\
12	-12.3927834161476\\
13	-12.3745270056034\\
14	-12.2800720467666\\
15	-12.2251687787961\\
16	-12.2024340412786\\
17	-12.1045037686339\\
18	-12.0359403139309\\
19	-11.9380959799812\\
20	-11.8577241236514\\
21	-11.7478715553014\\
22	-11.6949030791687\\
23	-11.603983752825\\
24	-11.5093604252914\\
25	-11.4731642750958\\
26	-11.3363950525897\\
27	-11.3125647550138\\
28	-11.1827299193742\\
29	-11.0906223253968\\
30	-11.0552441958852\\
31	-10.9816087887007\\
32	-10.8981064559162\\
33	-10.8050725710055\\
34	-10.6852647092766\\
35	-10.598988995348\\
36	-10.4783783813358\\
37	-10.3575008692123\\
38	-10.2918299209896\\
39	-10.2019584369158\\
40	-10.118814178002\\
41	-10.0330320496484\\
42	-9.97771575607192\\
43	-9.84486779669314\\
44	-9.78595187531235\\
45	-9.72037734799105\\
46	-9.63854111628923\\
47	-9.58047952736302\\
48	-9.47924905889908\\
49	-9.37468614039353\\
50	-9.34093198055917\\
51	-9.25142071097322\\
52	-9.16092636816303\\
53	-9.04820745721821\\
54	-8.95499442415307\\
55	-8.91735751274226\\
56	-8.8353214121586\\
57	-8.80581504775242\\
58	-8.70140889940108\\
59	-8.68620829542006\\
60	-8.58634832154631\\
61	-8.51778906361703\\
62	-8.44130145045453\\
63	-8.39092326546903\\
64	-8.27246034018279\\
65	-8.17063277808929\\
66	-8.1078972111239\\
67	-8.04204777563184\\
68	-7.97635448141528\\
69	-7.88394775423358\\
70	-7.84022328414124\\
71	-7.74372477086094\\
72	-7.65493221027782\\
73	-7.61499022708212\\
74	-7.52545737967378\\
75	-7.46077671273243\\
76	-7.38586741262133\\
77	-7.31469305259487\\
78	-7.23779838925212\\
79	-7.15100921213743\\
80	-7.12054168719292\\
81	-7.04967924823292\\
82	-7.02000981742997\\
83	-6.95899688186275\\
84	-6.92530044244612\\
85	-6.85204266131877\\
86	-6.77581439915657\\
87	-6.72723954641644\\
88	-6.67616330109121\\
89	-6.63887568277398\\
90	-6.60375437615349\\
91	-6.54126123810753\\
92	-6.47993548571967\\
93	-6.45743268245288\\
94	-6.42760036060124\\
95	-6.39829285226513\\
96	-6.35327349540778\\
97	-6.29728955892853\\
98	-6.26881024692568\\
99	-6.23784162342916\\
100	-6.23784162342916\\
};
\addplot [color=mycolor4,dashed,line width = 1.0pt]
  table[row sep=crcr]{%
1	-12.6901937610199\\
2	-12.7209550681353\\
3	-12.6913799767336\\
4	-12.7129052276969\\
5	-12.6604359049141\\
6	-12.6872083912416\\
7	-12.6601060041794\\
8	-12.6464392562099\\
9	-12.6306335173442\\
10	-12.6129136463994\\
11	-12.5689272056344\\
12	-12.484942421271\\
13	-12.4834487312422\\
14	-12.4354217013145\\
15	-12.3788239422479\\
16	-12.3430147890517\\
17	-12.2656723709961\\
18	-12.2034155611026\\
19	-12.1535554443167\\
20	-12.0895781373886\\
21	-11.993644137641\\
22	-11.9501741291811\\
23	-11.8686761960204\\
24	-11.8000223230284\\
25	-11.7617691007745\\
26	-11.673847035771\\
27	-11.6529450329296\\
28	-11.5365037924632\\
29	-11.4789988048675\\
30	-11.4425633105573\\
31	-11.3800337354791\\
32	-11.3119988576417\\
33	-11.2466855725857\\
34	-11.1646993951428\\
35	-11.1024600529185\\
36	-11.0140517238432\\
37	-10.9393276536749\\
38	-10.8620420973157\\
39	-10.805530366458\\
40	-10.7566823597709\\
41	-10.6884080229613\\
42	-10.6405512011955\\
43	-10.5544782918514\\
44	-10.5214466960164\\
45	-10.4626278784104\\
46	-10.3825568560995\\
47	-10.3571172201171\\
48	-10.281760380805\\
49	-10.2063847880905\\
50	-10.1618889426497\\
51	-10.1022990450753\\
52	-10.059466987381\\
53	-9.95569823131379\\
54	-9.90529925054328\\
55	-9.83725479245573\\
56	-9.80721238330109\\
57	-9.75834567588011\\
58	-9.70180719509454\\
59	-9.64804791615012\\
60	-9.58998359038668\\
61	-9.53385899704988\\
62	-9.47797475773683\\
63	-9.41808026443421\\
64	-9.34544970549186\\
65	-9.25428189473021\\
66	-9.23062369924587\\
67	-9.16580777087694\\
68	-9.09713346090237\\
69	-9.04808583860236\\
70	-9.02654503677852\\
71	-8.93302641093593\\
72	-8.87002890003833\\
73	-8.83314982130292\\
74	-8.76936475847507\\
75	-8.71698527815541\\
76	-8.67082427556413\\
77	-8.62502799217219\\
78	-8.5566277840417\\
79	-8.4977854363033\\
80	-8.44654666577229\\
81	-8.40087060925084\\
82	-8.34794601067601\\
83	-8.29804899237919\\
84	-8.24519856285385\\
85	-8.17968797400823\\
86	-8.14330836280207\\
87	-8.09157022806916\\
88	-8.03379452430299\\
89	-8.00420254035457\\
90	-7.95823588954539\\
91	-7.90015787623756\\
92	-7.85040152343915\\
93	-7.83569763593857\\
94	-7.77829046261082\\
95	-7.76189403937134\\
96	-7.7216315553353\\
97	-7.68729155535685\\
98	-7.63623379209701\\
99	-7.61142129554615\\
100	-7.61142129554615\\
};
\addlegendentry{C-REPS PCA}

\end{axis}
\end{tikzpicture}%

%% file: fig_2dof_reward.tex
%
%
\definecolor{mycolor1}{rgb}{0.00000,0.14700,0.94100}%
\definecolor{mycolor2}{rgb}{0.85000,0.32500,0.09800}%
\definecolor{mycolor3}{rgb}{0.92900,0.69400,0.32500}%
\definecolor{mycolor4}{rgb}{0.15000,0.15000,0.15000}%
\begin{tikzpicture}
\begin{axis}[%
height = 5.7cm,
width = \columnwidth,
ylabel style = {yshift = -1.6em,
                xshift = -0.5em},
xlabel style = {yshift =  0.5em},
xlabel       = {Iteration},
ylabel       = {Average Reward},
axis lines*  = left,
enlargelimits = false,
mark repeat = {2},
ytick       = {-3,-1,1,3,5,7,9},
ymin        = -2.3,
xmin        = 1,
ymax         = 8.97
]

\addplot[area legend,solid,fill=mycolor1,opacity=1.000000e-01,draw=none,forget plot]
table[row sep=crcr] {%
x	y\\
1	-2.83829566361388\\
50	0.190260164550244\\
100	1.47914524451661\\
150	2.7005897485158\\
200	3.48064398625286\\
250	4.51659933954908\\
300	4.96036253267839\\
350	5.46272959137124\\
400	5.64760464208977\\
450	6.15383947496511\\
500	6.20821222824679\\
550	6.73262621020802\\
600	6.70715018224671\\
650	6.97431923475495\\
700	7.26585124348621\\
750	7.32884331817587\\
800	7.38103725278154\\
850	7.41652779497143\\
900	7.3880580868199\\
950	7.34621275790298\\
1000	7.31648652839049\\
1000	8.30244327160951\\
950	8.30123384209702\\
900	8.2059449131801\\
850	8.12586100502856\\
800	8.15476734721847\\
750	8.14788268182413\\
700	8.03417335651379\\
650	8.00199696524505\\
600	8.0371890177533\\
550	7.75885698979198\\
500	7.55393117175321\\
450	7.40865632503489\\
400	7.03945295791023\\
350	6.82739720862875\\
300	6.48999706732161\\
250	5.99937146045092\\
200	5.28393821374714\\
150	4.2621124514842\\
100	2.90211235548339\\
50	1.12877563544976\\
1	-1.20228693638612\\
}--cycle;

\addplot [color=white!55!mycolor1,solid,forget plot]
  table[row sep=crcr]{%
1	-2.83829566361388\\
50	0.190260164550244\\
100	1.47914524451661\\
150	2.7005897485158\\
200	3.48064398625286\\
250	4.51659933954908\\
300	4.96036253267839\\
350	5.46272959137124\\
400	5.64760464208977\\
450	6.15383947496511\\
500	6.20821222824679\\
550	6.73262621020802\\
600	6.70715018224671\\
650	6.97431923475495\\
700	7.26585124348621\\
750	7.32884331817587\\
800	7.38103725278154\\
850	7.41652779497143\\
900	7.3880580868199\\
950	7.34621275790298\\
1000	7.31648652839049\\
};
\addplot [color=white!55!mycolor1,solid,forget plot]
  table[row sep=crcr]{%
1	-1.20228693638612\\
50	1.12877563544976\\
100	2.90211235548339\\
150	4.2621124514842\\
200	5.28393821374714\\
250	5.99937146045092\\
300	6.48999706732161\\
350	6.82739720862875\\
400	7.03945295791023\\
450	7.40865632503489\\
500	7.55393117175321\\
550	7.75885698979198\\
600	8.0371890177533\\
650	8.00199696524505\\
700	8.03417335651379\\
750	8.14788268182413\\
800	8.15476734721847\\
850	8.12586100502856\\
900	8.2059449131801\\
950	8.30123384209702\\
1000	8.30244327160951\\
};
\addplot [color=mycolor1,solid,line width = 1.0pt,forget plot,mark=square]
  table[row sep=crcr]{%
1	-2.0202913\\
50	0.6595179\\
100	2.1906288\\
150	3.4813511\\
200	4.3822911\\
250	5.2579854\\
300	5.7251798\\
350	6.1450634\\
400	6.3435288\\
450	6.7812479\\
500	6.8810717\\
550	7.2457416\\
600	7.3721696\\
650	7.4881581\\
700	7.6500123\\
750	7.738363\\
800	7.7679023\\
850	7.7711944\\
900	7.7970015\\
950	7.8237233\\
1000	7.8094649\\
};

\addplot[area legend,solid,fill=mycolor2,opacity=1.000000e-01,draw=none,forget plot]
table[row sep=crcr] {%
x	y\\
1	-2.83829566361388\\
50	-0.219414910126783\\
100	0.512109285063016\\
150	1.40500524652532\\
200	2.03676996457372\\
250	2.75600105044867\\
300	3.32234136953212\\
350	3.94804317229214\\
400	4.2291829441941\\
450	4.55507660161299\\
500	4.94243965630486\\
550	5.09572390148774\\
600	5.41159360389008\\
650	5.62829576655713\\
700	5.72923180988476\\
750	5.98460066684439\\
800	6.05529796601189\\
850	6.12208375263541\\
900	6.19734498349733\\
950	6.2903173485985\\
1000	6.42313741322921\\
1000	7.01900678677079\\
950	6.9983220514015\\
900	6.96519221650267\\
850	6.78749264736459\\
800	6.63680203398811\\
750	6.55047473315561\\
700	6.47881419011524\\
650	6.23359603344287\\
600	6.06791619610992\\
550	5.86044169851226\\
500	5.65196454369514\\
450	5.43192319838701\\
400	5.2304590558059\\
350	4.81238922770786\\
300	4.21734063046788\\
250	3.64095534955133\\
200	2.95689283542628\\
150	2.23363995347468\\
100	1.54566211493698\\
50	0.479347110126783\\
1	-1.20228693638612\\
}--cycle;

\addplot [color=white!55!mycolor2,solid,forget plot]
  table[row sep=crcr]{%
1	-2.83829566361388\\
50	-0.219414910126783\\
100	0.512109285063016\\
150	1.40500524652532\\
200	2.03676996457372\\
250	2.75600105044867\\
300	3.32234136953212\\
350	3.94804317229214\\
400	4.2291829441941\\
450	4.55507660161299\\
500	4.94243965630486\\
550	5.09572390148774\\
600	5.41159360389008\\
650	5.62829576655713\\
700	5.72923180988476\\
750	5.98460066684439\\
800	6.05529796601189\\
850	6.12208375263541\\
900	6.19734498349733\\
950	6.2903173485985\\
1000	6.42313741322921\\
};
\addplot [color=white!55!mycolor2,solid,forget plot]
  table[row sep=crcr]{%
1	-1.20228693638612\\
50	0.479347110126783\\
100	1.54566211493698\\
150	2.23363995347468\\
200	2.95689283542628\\
250	3.64095534955133\\
300	4.21734063046788\\
350	4.81238922770786\\
400	5.2304590558059\\
450	5.43192319838701\\
500	5.65196454369514\\
550	5.86044169851226\\
600	6.06791619610992\\
650	6.23359603344287\\
700	6.47881419011524\\
750	6.55047473315561\\
800	6.63680203398811\\
850	6.78749264736459\\
900	6.96519221650267\\
950	6.9983220514015\\
1000	7.01900678677079\\
};
\addplot [color=mycolor2,solid,line width = 1.0pt,forget plot,mark=diamond]
  table[row sep=crcr]{%
1	-2.0202913\\
50	0.1299661\\
100	1.0288857\\
150	1.8193226\\
200	2.4968314\\
250	3.1984782\\
300	3.769841\\
350	4.3802162\\
400	4.729821\\
450	4.9934999\\
500	5.2972021\\
550	5.4780828\\
600	5.7397549\\
650	5.9309459\\
700	6.104023\\
750	6.2675377\\
800	6.34605\\
850	6.4547882\\
900	6.5812686\\
950	6.6443197\\
1000	6.7210721\\
};

\addplot[area legend,solid,fill=mycolor3,opacity=1.000000e-01,draw=none,forget plot]
table[row sep=crcr] {%
x	y\\
1	-2.83829566361388\\
50	-0.458578664755649\\
100	-0.0306926875160514\\
150	0.543509521709815\\
200	0.891331037699773\\
250	1.24238600829689\\
300	1.55945083752838\\
350	1.76753115232892\\
400	1.85386122375074\\
450	2.14618571973883\\
500	2.36027843562617\\
550	2.42356479950867\\
600	2.56869294349738\\
650	2.66822200010634\\
700	2.77861007877868\\
750	3.05956459051806\\
800	3.01137012549254\\
850	3.09554956538715\\
900	3.30296242706901\\
950	3.34320098152744\\
1000	3.47980110320503\\
1000	3.89076149679497\\
950	3.83819641847256\\
900	3.84277337293099\\
850	3.75982563461285\\
800	3.42303247450746\\
750	3.36567220948194\\
700	3.37566652122132\\
650	3.13529599989366\\
600	2.99031625650262\\
550	2.80526180049133\\
500	2.85094636437383\\
450	2.51763328026117\\
400	2.42423017624926\\
350	2.35695204767108\\
300	1.93633296247162\\
250	1.68658959170311\\
200	1.37799396230023\\
150	1.04374207829019\\
100	0.519835487516051\\
50	-0.00202673524435112\\
1	-1.20228693638612\\
}--cycle;

\addplot [color=white!55!mycolor3,solid,forget plot]
  table[row sep=crcr]{%
1	-2.83829566361388\\
50	-0.458578664755649\\
100	-0.0306926875160514\\
150	0.543509521709815\\
200	0.891331037699773\\
250	1.24238600829689\\
300	1.55945083752838\\
350	1.76753115232892\\
400	1.85386122375074\\
450	2.14618571973883\\
500	2.36027843562617\\
550	2.42356479950867\\
600	2.56869294349738\\
650	2.66822200010634\\
700	2.77861007877868\\
750	3.05956459051806\\
800	3.01137012549254\\
850	3.09554956538715\\
900	3.30296242706901\\
950	3.34320098152744\\
1000	3.47980110320503\\
};
\addplot [color=white!55!mycolor3,solid,forget plot]
  table[row sep=crcr]{%
1	-1.20228693638612\\
50	-0.00202673524435112\\
100	0.519835487516051\\
150	1.04374207829019\\
200	1.37799396230023\\
250	1.68658959170311\\
300	1.93633296247162\\
350	2.35695204767108\\
400	2.42423017624926\\
450	2.51763328026117\\
500	2.85094636437383\\
550	2.80526180049133\\
600	2.99031625650262\\
650	3.13529599989366\\
700	3.37566652122132\\
750	3.36567220948194\\
800	3.42303247450746\\
850	3.75982563461285\\
900	3.84277337293099\\
950	3.83819641847256\\
1000	3.89076149679497\\
};
\addplot [color=mycolor3,solid,line width = 1.0pt,forget plot,mark=o]
  table[row sep=crcr]{%
1	-2.0202913\\
50	-0.2303027\\
100	0.2445714\\
150	0.7936258\\
200	1.1346625\\
250	1.4644878\\
300	1.7478919\\
350	2.0622416\\
400	2.1390457\\
450	2.3319095\\
500	2.6056124\\
550	2.6144133\\
600	2.7795046\\
650	2.901759\\
700	3.0771383\\
750	3.2126184\\
800	3.2172013\\
850	3.4276876\\
900	3.5728679\\
950	3.5906987\\
1000	3.6852813\\
};

\addplot[area legend,solid,fill=mycolor4,opacity=1.000000e-01,draw=none,forget plot]
table[row sep=crcr] {%
x	y\\
1	-2.83829566361388\\
50	0.4685495096134\\
100	0.900226048758596\\
150	0.814642502019504\\
200	0.661633117892355\\
250	0.614810589156182\\
300	0.921645276701341\\
350	0.781591625456977\\
400	0.964133960567641\\
450	1.07842075471102\\
500	1.16682135389006\\
550	0.940136407621731\\
600	1.21820771536266\\
650	1.28724599727001\\
700	1.20097550334927\\
750	1.04724824123378\\
800	1.13578649743673\\
850	1.33890813085502\\
900	1.15211955201399\\
950	1.08918474509227\\
1000	1.32937710673433\\
1000	3.71016749326567\\
950	3.68556425490773\\
900	3.91826404798601\\
850	3.70792746914498\\
800	3.88466170256327\\
750	3.78971995876622\\
700	3.72431869665073\\
650	3.80539760272999\\
600	3.81887728463734\\
550	3.83688259237827\\
500	4.07646524610994\\
450	3.69472244528898\\
400	3.85867743943236\\
350	3.79409697454302\\
300	4.19238152329866\\
250	3.98007301084382\\
200	4.04581968210764\\
150	3.7927944979805\\
100	4.0443071512414\\
50	3.1125502903866\\
1	-1.20228693638612\\
}--cycle;

\addplot [color=white!55!mycolor4,solid,forget plot]
  table[row sep=crcr]{%
1	-2.83829566361388\\
50	0.4685495096134\\
100	0.900226048758596\\
150	0.814642502019504\\
200	0.661633117892355\\
250	0.614810589156182\\
300	0.921645276701341\\
350	0.781591625456977\\
400	0.964133960567641\\
450	1.07842075471102\\
500	1.16682135389006\\
550	0.940136407621731\\
600	1.21820771536266\\
650	1.28724599727001\\
700	1.20097550334927\\
750	1.04724824123378\\
800	1.13578649743673\\
850	1.33890813085502\\
900	1.15211955201399\\
950	1.08918474509227\\
1000	1.32937710673433\\
};
\addplot [color=white!55!mycolor4,solid,forget plot]
  table[row sep=crcr]{%
1	-1.20228693638612\\
50	3.1125502903866\\
100	4.0443071512414\\
150	3.7927944979805\\
200	4.04581968210764\\
250	3.98007301084382\\
300	4.19238152329866\\
350	3.79409697454302\\
400	3.85867743943236\\
450	3.69472244528898\\
500	4.07646524610994\\
550	3.83688259237827\\
600	3.81887728463734\\
650	3.80539760272999\\
700	3.72431869665073\\
750	3.78971995876622\\
800	3.88466170256327\\
850	3.70792746914498\\
900	3.91826404798601\\
950	3.68556425490773\\
1000	3.71016749326567\\
};
\addplot [color=mycolor4,dashed,line width = 1.0pt,forget plot]
  table[row sep=crcr]{%
1	-2.0202913\\
50	1.7905499\\
100	2.4722666\\
150	2.3037185\\
200	2.3537264\\
250	2.2974418\\
300	2.5570134\\
350	2.2878443\\
400	2.4114057\\
450	2.3865716\\
500	2.6216433\\
550	2.3885095\\
600	2.5185425\\
650	2.5463218\\
700	2.4626471\\
750	2.4184841\\
800	2.5102241\\
850	2.5234178\\
900	2.5351918\\
950	2.3873745\\
1000	2.5197723\\
};

\node at (axis cs:842,8.62765080611269) {\small $\sim76\%$ Hit Rate};
\node at (axis cs:842,5.60765080611269) {\small $\sim53\%$ Hit Rate};
\node at (axis cs:842,4.27765080611269) {\small $\sim18\%$ Hit Rate};
\node at (axis cs:842,1.80765080611269) {\small $\sim11\%$ Hit Rate};

\end{axis}
\end{tikzpicture}%

%% file: fig_6dof_reward.tex
%
%
\definecolor{mycolor1}{rgb}{0.00000,0.14700,0.94100}%
\definecolor{mycolor2}{rgb}{0.85000,0.32500,0.09800}%
\definecolor{mycolor3}{rgb}{0.92900,0.69400,0.32500}%
\definecolor{mycolor4}{rgb}{0.15000,0.15000,0.15000}%
\begin{tikzpicture}
\begin{axis}[%
height = 5.1cm,
width = 1.1\columnwidth,
ylabel style = {yshift = -1.9em,
                xshift = -0.5em},
xlabel style = {yshift =  0.5em},
xlabel       = {Iteration},
ylabel       = {Average Reward},
axis lines*  = left,
enlargelimits = false,
mark repeat  = {1},
ymax         = 5.9,
legend style = {at = {(1, 0.1)}, 
                anchor = south east,
                draw = black,
                fill = white,
                legend cell align = left},
]

\addplot[area legend,solid,fill=mycolor1,opacity=1.000000e-01,draw=none,forget plot]
table[row sep=crcr] {%
x	y\\
1	1.61579202216881\\
51	3.14494128052817\\
101	3.80725358656076\\
151	4.18132126872692\\
201	4.66907705060083\\
251	4.79176650430308\\
301	4.69360190472707\\
351	4.68916920979954\\
401	4.90991256315047\\
451	4.74233055170546\\
500	4.99899842569157\\
500	5.35543460949209\\
451	5.1786558319857\\
401	4.95073430198122\\
351	4.95968028281741\\
301	4.95767368861369\\
251	5.07634477903048\\
201	4.86753490650774\\
151	4.32760419051959\\
101	3.94399066108309\\
51	3.33682230121599\\
1	1.77635690256632\\
}--cycle;

\addplot [color=white!55!mycolor1,solid,forget plot]
  table[row sep=crcr]{%
1	1.61579202216881\\
51	3.14494128052817\\
101	3.80725358656076\\
151	4.18132126872692\\
201	4.66907705060083\\
251	4.79176650430308\\
301	4.69360190472707\\
351	4.68916920979954\\
401	4.90991256315047\\
451	4.74233055170546\\
500	4.99899842569157\\
};
\addplot [color=white!55!mycolor1,solid,forget plot]
  table[row sep=crcr]{%
1	1.77635690256632\\
51	3.33682230121599\\
101	3.94399066108309\\
151	4.32760419051959\\
201	4.86753490650774\\
251	5.07634477903048\\
301	4.95767368861369\\
351	4.95968028281741\\
401	4.95073430198122\\
451	5.1786558319857\\
500	5.35543460949209\\
};
\addplot [color=mycolor1,solid,line width = 1.0pt,mark=square]
  table[row sep=crcr]{%
1	1.69607446236756\\
51	3.24088179087208\\
101	3.87562212382192\\
151	4.25446272962326\\
201	4.76830597855428\\
251	4.93405564166678\\
301	4.82563779667038\\
351	4.82442474630848\\
401	4.93032343256584\\
451	4.96049319184558\\
500	5.17721651759183\\
};
\addlegendentry{C-MORE Nuc. Norm}

\addplot[area legend,solid,fill=mycolor2,opacity=1.000000e-01,draw=none,forget plot]
table[row sep=crcr] {%
x	y\\
1	1.73244434196328\\
51	2.30104219200464\\
101	3.00147199904517\\
151	3.30939652668588\\
201	3.5393086968903\\
251	3.61047360164421\\
301	3.58001351597468\\
351	3.87809896421586\\
401	3.71941306524256\\
451	4.01526876351329\\
500	4.22409308992303\\
500	4.39120852230235\\
451	4.43756440667483\\
401	4.19678287469262\\
351	4.31740004674147\\
301	3.95218644911659\\
251	3.80648751258574\\
201	3.73940036817333\\
151	3.59384223335329\\
101	3.16383335050607\\
51	2.8236973197559\\
1	1.80507889070218\\
}--cycle;

\addplot [color=white!55!mycolor2,solid,forget plot]
  table[row sep=crcr]{%
1	1.73244434196328\\
51	2.30104219200464\\
101	3.00147199904517\\
151	3.30939652668588\\
201	3.5393086968903\\
251	3.61047360164421\\
301	3.58001351597468\\
351	3.87809896421586\\
401	3.71941306524256\\
451	4.01526876351329\\
500	4.22409308992303\\
};
\addplot [color=white!55!mycolor2,solid,forget plot]
  table[row sep=crcr]{%
1	1.80507889070218\\
51	2.8236973197559\\
101	3.16383335050607\\
151	3.59384223335329\\
201	3.73940036817333\\
251	3.80648751258574\\
301	3.95218644911659\\
351	4.31740004674147\\
401	4.19678287469262\\
451	4.43756440667483\\
500	4.39120852230235\\
};
\addplot [color=mycolor2,solid,line width = 1.0pt,mark=diamond]
  table[row sep=crcr]{%
1	1.76876161633273\\
51	2.56236975588027\\
101	3.08265267477562\\
151	3.45161938001958\\
201	3.63935453253182\\
251	3.70848055711497\\
301	3.76609998254563\\
351	4.09774950547866\\
401	3.95809796996759\\
451	4.22641658509406\\
500	4.30765080611269\\
};
\addlegendentry{C-MORE Ridge+PCA}

\node at (axis cs:419,4.57765080611269) {\small $\sim60\%$ Hit Rate};
\node at (axis cs:419,5.50765080611269) {\small $\sim80\%$ Hit Rate};

\end{axis}
\end{tikzpicture}%

%% file: fig_quadexp_pca.tex
%
%
\definecolor{mycolor1}{rgb}{0.00000,0.14700,0.94100}%
\definecolor{mycolor2}{rgb}{0.85000,0.32500,0.09800}%
\definecolor{mycolor3}{rgb}{0.92900,0.69400,0.32500}%
\definecolor{mycolor4}{rgb}{0.15000,0.15000,0.15000}%
\definecolor{mycolor5}{rgb}{0.13500,0.67800,0.18400}%
\begin{tikzpicture}
\begin{axis}[%
height = 7.7cm,
width = 0.6\columnwidth,
ylabel style = {yshift = .5em,
                xshift = -0.5em},
xlabel       = {Iteration},
ylabel       = {Average Log-Reward},
axis lines*  = left,
enlargelimits = false,
mark repeat = {9},
legend style = {at = {(0.01, 1)}, 
                anchor = north west,
                draw = black,
                fill = white,
                legend cell align = left},
]

\addplot[area legend,solid,fill=mycolor1,opacity=1.000000e-01,draw=none,forget plot]
table[row sep=crcr] {%
x	y\\
1	-12.7672036764284\\
2	-12.7633346799139\\
3	-12.8642288274863\\
4	-13.0834879206377\\
5	-13.2233317813872\\
6	-13.3353381761405\\
7	-13.4412526516489\\
8	-13.459779375736\\
9	-13.4165223185272\\
10	-13.3222892491482\\
11	-13.2409634266717\\
12	-13.1917529190949\\
13	-13.1472545512663\\
14	-13.0507994764152\\
15	-12.9954720518804\\
16	-12.9908924973944\\
17	-12.9602569476634\\
18	-12.9204452505219\\
19	-12.8673905657729\\
20	-12.828311019402\\
21	-12.7481825284518\\
22	-12.6383586999887\\
23	-12.5569433238098\\
24	-12.4964234555576\\
25	-12.4053933485855\\
26	-12.3506898366278\\
27	-12.2698482880658\\
28	-12.1857111996409\\
29	-12.125250115571\\
30	-12.0373732377393\\
31	-11.9799590085422\\
32	-11.8835789095674\\
33	-11.8227066141415\\
34	-11.7692709070963\\
35	-11.7083722237572\\
36	-11.6734055971419\\
37	-11.5855990108743\\
38	-11.5207244098294\\
39	-11.429761314934\\
40	-11.370193257915\\
41	-11.3358823413496\\
42	-11.2501206600971\\
43	-11.2346832385912\\
44	-11.1553397600226\\
45	-11.0989401586335\\
46	-11.045157331835\\
47	-10.9475054788907\\
48	-10.930745389147\\
49	-10.8427114039051\\
50	-10.8290389835022\\
51	-10.7290148669311\\
52	-10.6933710404018\\
53	-10.6639098323849\\
54	-10.5839793873517\\
55	-10.5651267707795\\
56	-10.5002634052081\\
57	-10.4664197213448\\
58	-10.420918806593\\
59	-10.4269939152028\\
60	-10.3261221229453\\
61	-10.3237872062286\\
62	-10.2473692612741\\
63	-10.2561872663797\\
64	-10.22927744464\\
65	-10.1655193942148\\
66	-10.1135405561931\\
67	-10.1428329905671\\
68	-10.0726730480716\\
69	-10.0468826724752\\
70	-10.0216629471215\\
71	-10.0214390378168\\
72	-9.95283748923329\\
73	-9.97764999209097\\
74	-9.97066880372708\\
75	-9.92971939219762\\
76	-9.91973209258675\\
77	-9.86525109057528\\
78	-9.88057646718972\\
79	-9.89215473068989\\
80	-9.8527453832499\\
81	-9.83957374868536\\
82	-9.84860102890605\\
83	-9.84029867953215\\
84	-9.82095830133912\\
85	-9.77479820388471\\
86	-9.80729191826575\\
87	-9.80375498044971\\
88	-9.80904131430163\\
89	-9.7855861035168\\
90	-9.78361318632516\\
91	-9.76611927251932\\
92	-9.78580275709141\\
93	-9.81755558668988\\
94	-9.74114726806059\\
95	-9.76657754708884\\
96	-9.75148508236511\\
97	-9.74214055969719\\
98	-9.78752431400714\\
99	-9.72565388407972\\
100	-9.72565388407972\\
100	-6.88783034298903\\
99	-6.88783034298903\\
98	-6.85590001980811\\
97	-6.88953802542688\\
96	-6.89238936018912\\
95	-6.89599500437271\\
94	-6.87889106322189\\
93	-6.8544965264761\\
92	-6.88572396390131\\
91	-6.89646730238966\\
90	-6.86764334217063\\
89	-6.90232773672198\\
88	-6.86912718736661\\
87	-6.90681731263099\\
86	-6.88518666862094\\
85	-6.89225066505289\\
84	-6.93458752535951\\
83	-6.91545719057952\\
82	-6.93294047962513\\
81	-6.94136484306656\\
80	-6.95142128737052\\
79	-6.9764544569727\\
78	-6.98753213830709\\
77	-7.02199586787167\\
76	-7.05280461098865\\
75	-7.07616038775169\\
74	-7.08523100457196\\
73	-7.13978483134467\\
72	-7.14643401572877\\
71	-7.17719450705802\\
70	-7.22285611085795\\
69	-7.2507912034922\\
68	-7.28693836770799\\
67	-7.3189434020824\\
66	-7.38190504134375\\
65	-7.42057552999035\\
64	-7.46164716822864\\
63	-7.52842840086286\\
62	-7.59984955167325\\
61	-7.65197839946322\\
60	-7.72295817410288\\
59	-7.79994338637852\\
58	-7.8780382762225\\
57	-7.95241734786582\\
56	-8.00906406783596\\
55	-8.07024131121871\\
54	-8.16975221272388\\
53	-8.27675839358899\\
52	-8.39102929534732\\
51	-8.47698612956372\\
50	-8.56025061363303\\
49	-8.6785946759563\\
48	-8.7671355269229\\
47	-8.85745962323147\\
46	-8.99303071116363\\
45	-9.10088737720755\\
44	-9.20367756869951\\
43	-9.31917676514105\\
42	-9.46090070717255\\
41	-9.56519632250323\\
40	-9.69039887302278\\
39	-9.81856134239236\\
38	-9.9257094859461\\
37	-10.0757841164266\\
36	-10.164814601535\\
35	-10.2758823090181\\
34	-10.4279603792568\\
33	-10.5217563198776\\
32	-10.6591965171608\\
31	-10.790553886661\\
30	-10.9015290660309\\
29	-11.0337565514359\\
28	-11.1613953974892\\
27	-11.2975302263477\\
26	-11.4180956314537\\
25	-11.5401085819441\\
24	-11.6930430724906\\
23	-11.8108842880881\\
22	-11.9359119338573\\
21	-12.0751331013074\\
20	-12.2005001581882\\
19	-12.3024296809429\\
18	-12.4165074518432\\
17	-12.4746647782673\\
16	-12.5878539078882\\
15	-12.6755803318275\\
14	-12.7142138039317\\
13	-12.7727769166378\\
12	-12.8790499567217\\
11	-12.9713391122329\\
10	-13.0869480594498\\
9	-13.1977088224354\\
8	-13.2031802562906\\
7	-13.1392357656322\\
6	-13.0182870556988\\
5	-12.9154124902439\\
4	-12.7756582952747\\
3	-12.6331471025563\\
2	-12.6327268557745\\
1	-12.6692242527908\\
}--cycle;

\addplot [color=white!55!mycolor1,solid,forget plot]
  table[row sep=crcr]{%
1	-12.7672036764284\\
2	-12.7633346799139\\
3	-12.8642288274863\\
4	-13.0834879206377\\
5	-13.2233317813872\\
6	-13.3353381761405\\
7	-13.4412526516489\\
8	-13.459779375736\\
9	-13.4165223185272\\
10	-13.3222892491482\\
11	-13.2409634266717\\
12	-13.1917529190949\\
13	-13.1472545512663\\
14	-13.0507994764152\\
15	-12.9954720518804\\
16	-12.9908924973944\\
17	-12.9602569476634\\
18	-12.9204452505219\\
19	-12.8673905657729\\
20	-12.828311019402\\
21	-12.7481825284518\\
22	-12.6383586999887\\
23	-12.5569433238098\\
24	-12.4964234555576\\
25	-12.4053933485855\\
26	-12.3506898366278\\
27	-12.2698482880658\\
28	-12.1857111996409\\
29	-12.125250115571\\
30	-12.0373732377393\\
31	-11.9799590085422\\
32	-11.8835789095674\\
33	-11.8227066141415\\
34	-11.7692709070963\\
35	-11.7083722237572\\
36	-11.6734055971419\\
37	-11.5855990108743\\
38	-11.5207244098294\\
39	-11.429761314934\\
40	-11.370193257915\\
41	-11.3358823413496\\
42	-11.2501206600971\\
43	-11.2346832385912\\
44	-11.1553397600226\\
45	-11.0989401586335\\
46	-11.045157331835\\
47	-10.9475054788907\\
48	-10.930745389147\\
49	-10.8427114039051\\
50	-10.8290389835022\\
51	-10.7290148669311\\
52	-10.6933710404018\\
53	-10.6639098323849\\
54	-10.5839793873517\\
55	-10.5651267707795\\
56	-10.5002634052081\\
57	-10.4664197213448\\
58	-10.420918806593\\
59	-10.4269939152028\\
60	-10.3261221229453\\
61	-10.3237872062286\\
62	-10.2473692612741\\
63	-10.2561872663797\\
64	-10.22927744464\\
65	-10.1655193942148\\
66	-10.1135405561931\\
67	-10.1428329905671\\
68	-10.0726730480716\\
69	-10.0468826724752\\
70	-10.0216629471215\\
71	-10.0214390378168\\
72	-9.95283748923329\\
73	-9.97764999209097\\
74	-9.97066880372708\\
75	-9.92971939219762\\
76	-9.91973209258675\\
77	-9.86525109057528\\
78	-9.88057646718972\\
79	-9.89215473068989\\
80	-9.8527453832499\\
81	-9.83957374868536\\
82	-9.84860102890605\\
83	-9.84029867953215\\
84	-9.82095830133912\\
85	-9.77479820388471\\
86	-9.80729191826575\\
87	-9.80375498044971\\
88	-9.80904131430163\\
89	-9.7855861035168\\
90	-9.78361318632516\\
91	-9.76611927251932\\
92	-9.78580275709141\\
93	-9.81755558668988\\
94	-9.74114726806059\\
95	-9.76657754708884\\
96	-9.75148508236511\\
97	-9.74214055969719\\
98	-9.78752431400714\\
99	-9.72565388407972\\
100	-9.72565388407972\\
};
\addplot [color=white!55!mycolor1,solid,forget plot]
  table[row sep=crcr]{%
1	-12.6692242527908\\
2	-12.6327268557745\\
3	-12.6331471025563\\
4	-12.7756582952747\\
5	-12.9154124902439\\
6	-13.0182870556988\\
7	-13.1392357656322\\
8	-13.2031802562906\\
9	-13.1977088224354\\
10	-13.0869480594498\\
11	-12.9713391122329\\
12	-12.8790499567217\\
13	-12.7727769166378\\
14	-12.7142138039317\\
15	-12.6755803318275\\
16	-12.5878539078882\\
17	-12.4746647782673\\
18	-12.4165074518432\\
19	-12.3024296809429\\
20	-12.2005001581882\\
21	-12.0751331013074\\
22	-11.9359119338573\\
23	-11.8108842880881\\
24	-11.6930430724906\\
25	-11.5401085819441\\
26	-11.4180956314537\\
27	-11.2975302263477\\
28	-11.1613953974892\\
29	-11.0337565514359\\
30	-10.9015290660309\\
31	-10.790553886661\\
32	-10.6591965171608\\
33	-10.5217563198776\\
34	-10.4279603792568\\
35	-10.2758823090181\\
36	-10.164814601535\\
37	-10.0757841164266\\
38	-9.9257094859461\\
39	-9.81856134239236\\
40	-9.69039887302278\\
41	-9.56519632250323\\
42	-9.46090070717255\\
43	-9.31917676514105\\
44	-9.20367756869951\\
45	-9.10088737720755\\
46	-8.99303071116363\\
47	-8.85745962323147\\
48	-8.7671355269229\\
49	-8.6785946759563\\
50	-8.56025061363303\\
51	-8.47698612956372\\
52	-8.39102929534732\\
53	-8.27675839358899\\
54	-8.16975221272388\\
55	-8.07024131121871\\
56	-8.00906406783596\\
57	-7.95241734786582\\
58	-7.8780382762225\\
59	-7.79994338637852\\
60	-7.72295817410288\\
61	-7.65197839946322\\
62	-7.59984955167325\\
63	-7.52842840086286\\
64	-7.46164716822864\\
65	-7.42057552999035\\
66	-7.38190504134375\\
67	-7.3189434020824\\
68	-7.28693836770799\\
69	-7.2507912034922\\
70	-7.22285611085795\\
71	-7.17719450705802\\
72	-7.14643401572877\\
73	-7.13978483134467\\
74	-7.08523100457196\\
75	-7.07616038775169\\
76	-7.05280461098865\\
77	-7.02199586787167\\
78	-6.98753213830709\\
79	-6.9764544569727\\
80	-6.95142128737052\\
81	-6.94136484306656\\
82	-6.93294047962513\\
83	-6.91545719057952\\
84	-6.93458752535951\\
85	-6.89225066505289\\
86	-6.88518666862094\\
87	-6.90681731263099\\
88	-6.86912718736661\\
89	-6.90232773672198\\
90	-6.86764334217063\\
91	-6.89646730238966\\
92	-6.88572396390131\\
93	-6.8544965264761\\
94	-6.87889106322189\\
95	-6.89599500437271\\
96	-6.89238936018912\\
97	-6.88953802542688\\
98	-6.85590001980811\\
99	-6.88783034298903\\
100	-6.88783034298903\\
};
\addplot [color=mycolor1,solid,line width = 1.0pt,mark=square]
  table[row sep=crcr]{%
1	-12.7182139646096\\
2	-12.6980307678442\\
3	-12.7486879650213\\
4	-12.9295731079562\\
5	-13.0693721358155\\
6	-13.1768126159196\\
7	-13.2902442086406\\
8	-13.3314798160133\\
9	-13.3071155704813\\
10	-13.204618654299\\
11	-13.1061512694523\\
12	-13.0354014379083\\
13	-12.960015733952\\
14	-12.8825066401735\\
15	-12.8355261918539\\
16	-12.7893732026413\\
17	-12.7174608629653\\
18	-12.6684763511826\\
19	-12.5849101233579\\
20	-12.5144055887951\\
21	-12.4116578148796\\
22	-12.287135316923\\
23	-12.183913805949\\
24	-12.0947332640241\\
25	-11.9727509652648\\
26	-11.8843927340408\\
27	-11.7836892572067\\
28	-11.673553298565\\
29	-11.5795033335035\\
30	-11.4694511518851\\
31	-11.3852564476016\\
32	-11.2713877133641\\
33	-11.1722314670096\\
34	-11.0986156431765\\
35	-10.9921272663877\\
36	-10.9191100993385\\
37	-10.8306915636505\\
38	-10.7232169478877\\
39	-10.6241613286632\\
40	-10.5302960654689\\
41	-10.4505393319264\\
42	-10.3555106836348\\
43	-10.2769300018661\\
44	-10.1795086643611\\
45	-10.0999137679206\\
46	-10.0190940214993\\
47	-9.90248255106107\\
48	-9.84894045803494\\
49	-9.76065303993072\\
50	-9.69464479856759\\
51	-9.60300049824741\\
52	-9.54220016787456\\
53	-9.47033411298693\\
54	-9.37686580003777\\
55	-9.31768404099911\\
56	-9.25466373652205\\
57	-9.20941853460531\\
58	-9.14947854140775\\
59	-9.11346865079068\\
60	-9.0245401485241\\
61	-8.9878828028459\\
62	-8.92360940647367\\
63	-8.89230783362128\\
64	-8.84546230643434\\
65	-8.79304746210256\\
66	-8.7477227987684\\
67	-8.73088819632473\\
68	-8.67980570788982\\
69	-8.64883693798372\\
70	-8.62225952898973\\
71	-8.59931677243742\\
72	-8.54963575248103\\
73	-8.55871741171782\\
74	-8.52794990414952\\
75	-8.50293988997466\\
76	-8.4862683517877\\
77	-8.44362347922347\\
78	-8.4340543027484\\
79	-8.43430459383129\\
80	-8.40208333531021\\
81	-8.39046929587596\\
82	-8.39077075426559\\
83	-8.37787793505583\\
84	-8.37777291334931\\
85	-8.3335244344688\\
86	-8.34623929344334\\
87	-8.35528614654035\\
88	-8.33908425083412\\
89	-8.34395692011939\\
90	-8.32562826424789\\
91	-8.33129328745449\\
92	-8.33576336049636\\
93	-8.33602605658299\\
94	-8.31001916564124\\
95	-8.33128627573078\\
96	-8.32193722127712\\
97	-8.31583929256203\\
98	-8.32171216690762\\
99	-8.30674211353437\\
100	-8.30674211353437\\
};
\addlegendentry{dz = 3}

\addplot[area legend,solid,fill=mycolor2,opacity=1.000000e-01,draw=none,forget plot]
table[row sep=crcr] {%
x	y\\
1	-12.8248886938533\\
2	-12.8956532603462\\
3	-13.003527255367\\
4	-13.1415623354951\\
5	-13.2862865296715\\
6	-13.4561358142939\\
7	-13.5142973951929\\
8	-13.5414354707422\\
9	-13.5123466747238\\
10	-13.4624943855272\\
11	-13.4102221758739\\
12	-13.3229783239037\\
13	-13.2779172803172\\
14	-13.2263598064568\\
15	-13.1917304193639\\
16	-13.1468440152209\\
17	-13.1291285657432\\
18	-13.0968246095857\\
19	-13.0680608332633\\
20	-13.0109432016852\\
21	-12.9765244090811\\
22	-12.878566987415\\
23	-12.761886172145\\
24	-12.7262594957311\\
25	-12.6286448191767\\
26	-12.5560920020346\\
27	-12.4773676095627\\
28	-12.4067982380633\\
29	-12.3280063416299\\
30	-12.2357194238671\\
31	-12.1561095746502\\
32	-12.0745834540245\\
33	-12.0005094053076\\
34	-11.9734555930064\\
35	-11.8841828418802\\
36	-11.8200373868148\\
37	-11.7272185901193\\
38	-11.6864221281706\\
39	-11.6322181268204\\
40	-11.5670391891847\\
41	-11.4552790649606\\
42	-11.4200107795129\\
43	-11.3608581518349\\
44	-11.3247758005646\\
45	-11.2175023667518\\
46	-11.1516007355897\\
47	-11.0982464232526\\
48	-11.0192087942799\\
49	-10.9720260010957\\
50	-10.9328088705555\\
51	-10.8567176773178\\
52	-10.7914070872182\\
53	-10.7612137358185\\
54	-10.6896160851661\\
55	-10.6010267954278\\
56	-10.5748588323853\\
57	-10.5242942362671\\
58	-10.4631252625644\\
59	-10.4034056661379\\
60	-10.3624304939445\\
61	-10.3219246685109\\
62	-10.2809283409946\\
63	-10.2290677892757\\
64	-10.2004080607374\\
65	-10.1482894893831\\
66	-10.1367627158003\\
67	-10.0588317706751\\
68	-10.0407964879401\\
69	-9.97247892450084\\
70	-9.96553369448413\\
71	-9.93256365434346\\
72	-9.88717110115839\\
73	-9.83156017747612\\
74	-9.83231590664709\\
75	-9.78145536714616\\
76	-9.8121775655923\\
77	-9.74977422783258\\
78	-9.70953133138\\
79	-9.72786798975911\\
80	-9.6989027199887\\
81	-9.65476582076206\\
82	-9.66368644467389\\
83	-9.64277151043928\\
84	-9.62284993834932\\
85	-9.59804260237983\\
86	-9.58636856310383\\
87	-9.5757138083357\\
88	-9.59259546421193\\
89	-9.55600031624976\\
90	-9.54710742140341\\
91	-9.5366841943098\\
92	-9.50522002478446\\
93	-9.51774874480623\\
94	-9.49730968875238\\
95	-9.49017702091804\\
96	-9.4747896836396\\
97	-9.48302806361476\\
98	-9.44644126347522\\
99	-9.47371312151123\\
100	-9.47371312151123\\
100	-7.73534578791656\\
99	-7.73534578791656\\
98	-7.74585835137252\\
97	-7.74319477696841\\
96	-7.77375247016672\\
95	-7.76228153952559\\
94	-7.76961200006916\\
93	-7.78335659242002\\
92	-7.80363474061029\\
91	-7.79546242674196\\
90	-7.80143879252488\\
89	-7.85352674991795\\
88	-7.8521968615239\\
87	-7.892906559686\\
86	-7.90201206367384\\
85	-7.95152302022701\\
84	-7.95830716682564\\
83	-7.9825031653401\\
82	-8.0275666857698\\
81	-8.01973188474357\\
80	-8.0523003284054\\
79	-8.09916304257903\\
78	-8.15364822583614\\
77	-8.16728633767675\\
76	-8.19843756754808\\
75	-8.23158286378037\\
74	-8.2457815683862\\
73	-8.31101640613662\\
72	-8.36190555955432\\
71	-8.40998499855976\\
70	-8.46150123925825\\
69	-8.49551850228856\\
68	-8.5239333728549\\
67	-8.59755695618678\\
66	-8.60076878023556\\
65	-8.68946930463897\\
64	-8.76003526148635\\
63	-8.79881315176295\\
62	-8.88161210216944\\
61	-8.91306954452585\\
60	-8.9748728699535\\
59	-9.0303682689955\\
58	-9.09537452765868\\
57	-9.15623438461139\\
56	-9.20526877309744\\
55	-9.28725940698457\\
54	-9.40197221643198\\
53	-9.47491932127672\\
52	-9.55634342175751\\
51	-9.64458444786698\\
50	-9.7117890423108\\
49	-9.84299126103518\\
48	-9.92101883535255\\
47	-9.99765217514483\\
46	-10.0625448063541\\
45	-10.1656730052012\\
44	-10.3091706289694\\
43	-10.3752764445461\\
42	-10.4838648907968\\
41	-10.5807291305986\\
40	-10.6669916257162\\
39	-10.7630175068255\\
38	-10.8425735214968\\
37	-10.9537823014628\\
36	-11.0375012125521\\
35	-11.1203745345027\\
34	-11.2505506650949\\
33	-11.3622579532375\\
32	-11.482193987832\\
31	-11.5545160040017\\
30	-11.6580811407645\\
29	-11.7800722498556\\
28	-11.8779583157528\\
27	-11.9700246329829\\
26	-12.0817494384556\\
25	-12.1926954326133\\
24	-12.2629923240157\\
23	-12.4170982407092\\
22	-12.4965782672939\\
21	-12.59593910156\\
20	-12.6934244699092\\
19	-12.767670616804\\
18	-12.8277262289026\\
17	-12.8494601942228\\
16	-12.9085905623551\\
15	-12.9360215626052\\
14	-12.992069096965\\
13	-13.0119297567111\\
12	-13.0438413701383\\
11	-13.1215674683459\\
10	-13.1945064119952\\
9	-13.298278597666\\
8	-13.2987450498851\\
7	-13.2381298451749\\
6	-13.2291036806842\\
5	-13.1230766772751\\
4	-13.019408012264\\
3	-12.8712969634938\\
2	-12.7668648646708\\
1	-12.7559775180985\\
}--cycle;
\addplot [color=white!55!mycolor2,solid,forget plot]
  table[row sep=crcr]{%
1	-12.8248886938533\\
2	-12.8956532603462\\
3	-13.003527255367\\
4	-13.1415623354951\\
5	-13.2862865296715\\
6	-13.4561358142939\\
7	-13.5142973951929\\
8	-13.5414354707422\\
9	-13.5123466747238\\
10	-13.4624943855272\\
11	-13.4102221758739\\
12	-13.3229783239037\\
13	-13.2779172803172\\
14	-13.2263598064568\\
15	-13.1917304193639\\
16	-13.1468440152209\\
17	-13.1291285657432\\
18	-13.0968246095857\\
19	-13.0680608332633\\
20	-13.0109432016852\\
21	-12.9765244090811\\
22	-12.878566987415\\
23	-12.761886172145\\
24	-12.7262594957311\\
25	-12.6286448191767\\
26	-12.5560920020346\\
27	-12.4773676095627\\
28	-12.4067982380633\\
29	-12.3280063416299\\
30	-12.2357194238671\\
31	-12.1561095746502\\
32	-12.0745834540245\\
33	-12.0005094053076\\
34	-11.9734555930064\\
35	-11.8841828418802\\
36	-11.8200373868148\\
37	-11.7272185901193\\
38	-11.6864221281706\\
39	-11.6322181268204\\
40	-11.5670391891847\\
41	-11.4552790649606\\
42	-11.4200107795129\\
43	-11.3608581518349\\
44	-11.3247758005646\\
45	-11.2175023667518\\
46	-11.1516007355897\\
47	-11.0982464232526\\
48	-11.0192087942799\\
49	-10.9720260010957\\
50	-10.9328088705555\\
51	-10.8567176773178\\
52	-10.7914070872182\\
53	-10.7612137358185\\
54	-10.6896160851661\\
55	-10.6010267954278\\
56	-10.5748588323853\\
57	-10.5242942362671\\
58	-10.4631252625644\\
59	-10.4034056661379\\
60	-10.3624304939445\\
61	-10.3219246685109\\
62	-10.2809283409946\\
63	-10.2290677892757\\
64	-10.2004080607374\\
65	-10.1482894893831\\
66	-10.1367627158003\\
67	-10.0588317706751\\
68	-10.0407964879401\\
69	-9.97247892450084\\
70	-9.96553369448413\\
71	-9.93256365434346\\
72	-9.88717110115839\\
73	-9.83156017747612\\
74	-9.83231590664709\\
75	-9.78145536714616\\
76	-9.8121775655923\\
77	-9.74977422783258\\
78	-9.70953133138\\
79	-9.72786798975911\\
80	-9.6989027199887\\
81	-9.65476582076206\\
82	-9.66368644467389\\
83	-9.64277151043928\\
84	-9.62284993834932\\
85	-9.59804260237983\\
86	-9.58636856310383\\
87	-9.5757138083357\\
88	-9.59259546421193\\
89	-9.55600031624976\\
90	-9.54710742140341\\
91	-9.5366841943098\\
92	-9.50522002478446\\
93	-9.51774874480623\\
94	-9.49730968875238\\
95	-9.49017702091804\\
96	-9.4747896836396\\
97	-9.48302806361476\\
98	-9.44644126347522\\
99	-9.47371312151123\\
100	-9.47371312151123\\
};
\addplot [color=white!55!mycolor2,solid,forget plot]
  table[row sep=crcr]{%
1	-12.7559775180985\\
2	-12.7668648646708\\
3	-12.8712969634938\\
4	-13.019408012264\\
5	-13.1230766772751\\
6	-13.2291036806842\\
7	-13.2381298451749\\
8	-13.2987450498851\\
9	-13.298278597666\\
10	-13.1945064119952\\
11	-13.1215674683459\\
12	-13.0438413701383\\
13	-13.0119297567111\\
14	-12.992069096965\\
15	-12.9360215626052\\
16	-12.9085905623551\\
17	-12.8494601942228\\
18	-12.8277262289026\\
19	-12.767670616804\\
20	-12.6934244699092\\
21	-12.59593910156\\
22	-12.4965782672939\\
23	-12.4170982407092\\
24	-12.2629923240157\\
25	-12.1926954326133\\
26	-12.0817494384556\\
27	-11.9700246329829\\
28	-11.8779583157528\\
29	-11.7800722498556\\
30	-11.6580811407645\\
31	-11.5545160040017\\
32	-11.482193987832\\
33	-11.3622579532375\\
34	-11.2505506650949\\
35	-11.1203745345027\\
36	-11.0375012125521\\
37	-10.9537823014628\\
38	-10.8425735214968\\
39	-10.7630175068255\\
40	-10.6669916257162\\
41	-10.5807291305986\\
42	-10.4838648907968\\
43	-10.3752764445461\\
44	-10.3091706289694\\
45	-10.1656730052012\\
46	-10.0625448063541\\
47	-9.99765217514483\\
48	-9.92101883535255\\
49	-9.84299126103518\\
50	-9.7117890423108\\
51	-9.64458444786698\\
52	-9.55634342175751\\
53	-9.47491932127672\\
54	-9.40197221643198\\
55	-9.28725940698457\\
56	-9.20526877309744\\
57	-9.15623438461139\\
58	-9.09537452765868\\
59	-9.0303682689955\\
60	-8.9748728699535\\
61	-8.91306954452585\\
62	-8.88161210216944\\
63	-8.79881315176295\\
64	-8.76003526148635\\
65	-8.68946930463897\\
66	-8.60076878023556\\
67	-8.59755695618678\\
68	-8.5239333728549\\
69	-8.49551850228856\\
70	-8.46150123925825\\
71	-8.40998499855976\\
72	-8.36190555955432\\
73	-8.31101640613662\\
74	-8.2457815683862\\
75	-8.23158286378037\\
76	-8.19843756754808\\
77	-8.16728633767675\\
78	-8.15364822583614\\
79	-8.09916304257903\\
80	-8.0523003284054\\
81	-8.01973188474357\\
82	-8.0275666857698\\
83	-7.9825031653401\\
84	-7.95830716682564\\
85	-7.95152302022701\\
86	-7.90201206367384\\
87	-7.892906559686\\
88	-7.8521968615239\\
89	-7.85352674991795\\
90	-7.80143879252488\\
91	-7.79546242674196\\
92	-7.80363474061029\\
93	-7.78335659242002\\
94	-7.76961200006916\\
95	-7.76228153952559\\
96	-7.77375247016672\\
97	-7.74319477696841\\
98	-7.74585835137252\\
99	-7.73534578791656\\
100	-7.73534578791656\\
};
\addplot [color=mycolor2,solid,line width = 1.0pt,mark=diamond]
  table[row sep=crcr]{%
1	-12.7904331059759\\
2	-12.8312590625085\\
3	-12.9374121094304\\
4	-13.0804851738796\\
5	-13.2046816034733\\
6	-13.3426197474891\\
7	-13.3762136201839\\
8	-13.4200902603137\\
9	-13.4053126361949\\
10	-13.3285003987612\\
11	-13.2658948221099\\
12	-13.183409847021\\
13	-13.1449235185141\\
14	-13.1092144517109\\
15	-13.0638759909846\\
16	-13.027717288788\\
17	-12.989294379983\\
18	-12.9622754192442\\
19	-12.9178657250337\\
20	-12.8521838357972\\
21	-12.7862317553205\\
22	-12.6875726273544\\
23	-12.5894922064271\\
24	-12.4946259098734\\
25	-12.410670125895\\
26	-12.3189207202451\\
27	-12.2236961212728\\
28	-12.142378276908\\
29	-12.0540392957428\\
30	-11.9469002823158\\
31	-11.855312789326\\
32	-11.7783887209282\\
33	-11.6813836792725\\
34	-11.6120031290507\\
35	-11.5022786881915\\
36	-11.4287692996834\\
37	-11.340500445791\\
38	-11.2644978248337\\
39	-11.197617816823\\
40	-11.1170154074504\\
41	-11.0180040977796\\
42	-10.9519378351549\\
43	-10.8680672981905\\
44	-10.816973214767\\
45	-10.6915876859765\\
46	-10.6070727709719\\
47	-10.5479492991987\\
48	-10.4701138148162\\
49	-10.4075086310654\\
50	-10.3222989564331\\
51	-10.2506510625924\\
52	-10.1738752544878\\
53	-10.1180665285476\\
54	-10.045794150799\\
55	-9.94414310120621\\
56	-9.89006380274139\\
57	-9.84026431043925\\
58	-9.77924989511156\\
59	-9.71688696756668\\
60	-9.66865168194899\\
61	-9.61749710651839\\
62	-9.58127022158203\\
63	-9.51394047051931\\
64	-9.48022166111186\\
65	-9.41887939701106\\
66	-9.36876574801794\\
67	-9.32819436343092\\
68	-9.28236493039749\\
69	-9.2339987133947\\
70	-9.21351746687119\\
71	-9.17127432645161\\
72	-9.12453833035635\\
73	-9.07128829180637\\
74	-9.03904873751665\\
75	-9.00651911546327\\
76	-9.00530756657019\\
77	-8.95853028275466\\
78	-8.93158977860807\\
79	-8.91351551616907\\
80	-8.87560152419705\\
81	-8.83724885275282\\
82	-8.84562656522185\\
83	-8.81263733788969\\
84	-8.79057855258748\\
85	-8.77478281130342\\
86	-8.74419031338884\\
87	-8.73431018401085\\
88	-8.72239616286791\\
89	-8.70476353308386\\
90	-8.67427310696415\\
91	-8.66607331052588\\
92	-8.65442738269738\\
93	-8.65055266861313\\
94	-8.63346084441077\\
95	-8.62622928022182\\
96	-8.62427107690316\\
97	-8.61311142029158\\
98	-8.59614980742387\\
99	-8.60452945471389\\
100	-8.60452945471389\\
};
\addlegendentry{dz = 6}

\addplot[area legend,solid,fill=mycolor3,opacity=1.000000e-01,draw=none,forget plot]
table[row sep=crcr] {%
x	y\\
1	-12.9322347014806\\
2	-13.0095242004891\\
3	-13.1017715384092\\
4	-13.2981197730883\\
5	-13.430915191158\\
6	-13.5159149848724\\
7	-13.5563549574365\\
8	-13.5694395981143\\
9	-13.5322963808066\\
10	-13.4581249541529\\
11	-13.3853929414074\\
12	-13.349993017645\\
13	-13.2868329040913\\
14	-13.2603716336268\\
15	-13.247402648989\\
16	-13.2106462634961\\
17	-13.1563502691084\\
18	-13.1775350543842\\
19	-13.1328172899385\\
20	-13.0851262462001\\
21	-13.0202991558685\\
22	-12.9178832045742\\
23	-12.8323381036083\\
24	-12.7631493560865\\
25	-12.6834540555873\\
26	-12.6169401709716\\
27	-12.5616598106398\\
28	-12.4469764500756\\
29	-12.3740632687606\\
30	-12.3006110723645\\
31	-12.2142195667587\\
32	-12.1545359685878\\
33	-12.0740413410591\\
34	-11.9984944591339\\
35	-11.9425510635944\\
36	-11.8573901174916\\
37	-11.7863354416219\\
38	-11.7305450931553\\
39	-11.643965024019\\
40	-11.6224553076041\\
41	-11.5031544107653\\
42	-11.465257628898\\
43	-11.4179549181895\\
44	-11.3252126175166\\
45	-11.260977861194\\
46	-11.1961689709994\\
47	-11.1462026912051\\
48	-11.0799216335955\\
49	-11.0118108952012\\
50	-10.9482008518544\\
51	-10.8997986078294\\
52	-10.8327522966284\\
53	-10.7713882410808\\
54	-10.6765029933136\\
55	-10.6361244505785\\
56	-10.5662233144206\\
57	-10.5115303506242\\
58	-10.4444160506824\\
59	-10.3792511813253\\
60	-10.3300336719907\\
61	-10.2884040315383\\
62	-10.2461044004575\\
63	-10.1837398623502\\
64	-10.1098465424272\\
65	-10.0247000375904\\
66	-10.0137555337507\\
67	-9.95405241976507\\
68	-9.92709230162255\\
69	-9.84746219667929\\
70	-9.80336864765488\\
71	-9.75798072303103\\
72	-9.74312121867581\\
73	-9.68342929966209\\
74	-9.64124767123355\\
75	-9.59671962853753\\
76	-9.60695157208452\\
77	-9.55646351366437\\
78	-9.50908308815571\\
79	-9.45754556051774\\
80	-9.38636067497994\\
81	-9.37033781374462\\
82	-9.36767029995173\\
83	-9.3311102496807\\
84	-9.28280863801116\\
85	-9.2888479445723\\
86	-9.27788686212465\\
87	-9.22883743217734\\
88	-9.22388054946078\\
89	-9.18804093306757\\
90	-9.15948934432609\\
91	-9.15610249236961\\
92	-9.12816012126077\\
93	-9.1061816823333\\
94	-9.09950619336174\\
95	-9.07005104922814\\
96	-9.06063017468448\\
97	-9.00986931603618\\
98	-9.02436487868773\\
99	-9.00848077206592\\
100	-9.00848077206592\\
100	-7.52756516347543\\
99	-7.52756516347543\\
98	-7.50703508761956\\
97	-7.5298845153176\\
96	-7.53911100480769\\
95	-7.55673144935163\\
94	-7.54816444682695\\
93	-7.5618656972247\\
92	-7.57490845256365\\
91	-7.59859154524355\\
90	-7.63671557656485\\
89	-7.63636560946712\\
88	-7.66183372344072\\
87	-7.66742142907906\\
86	-7.70947245529419\\
85	-7.74610090941019\\
84	-7.75301709910105\\
83	-7.78827057877805\\
82	-7.81908454150282\\
81	-7.86946720658264\\
80	-7.90035856385354\\
79	-7.93128350448747\\
78	-7.95738072054221\\
77	-8.0133137764213\\
76	-8.04681511168041\\
75	-8.09985974956333\\
74	-8.13747041011114\\
73	-8.21767342614369\\
72	-8.22494597048634\\
71	-8.29527557935196\\
70	-8.36940900386269\\
69	-8.43467745684972\\
68	-8.48008897313136\\
67	-8.58112721121013\\
66	-8.63832615036469\\
65	-8.6908384503051\\
64	-8.74232676502489\\
63	-8.81549445029552\\
62	-8.90386412543385\\
61	-8.94945674229475\\
60	-9.04763499027455\\
59	-9.11577703184197\\
58	-9.20883399367772\\
57	-9.26462863766276\\
56	-9.39577360536699\\
55	-9.44878522145168\\
54	-9.53706886969175\\
53	-9.64604556458151\\
52	-9.71515528788501\\
51	-9.82338313562993\\
50	-9.89937653097599\\
49	-9.98682162747724\\
48	-10.0758314648675\\
47	-10.178286675042\\
46	-10.2942231719965\\
45	-10.3999519315684\\
44	-10.4762938963533\\
43	-10.5857146490186\\
42	-10.6959038500043\\
41	-10.7919695095174\\
40	-10.8975810313831\\
39	-10.9876330841981\\
38	-11.0900024712871\\
37	-11.2052551064994\\
36	-11.3127605223502\\
35	-11.3904855591328\\
34	-11.5318174137414\\
33	-11.5875328011972\\
32	-11.6980476397426\\
31	-11.8160434058546\\
30	-11.9038259607777\\
29	-12.011827531492\\
28	-12.0871305926921\\
27	-12.1906787336194\\
26	-12.2842935688498\\
25	-12.3790509381396\\
24	-12.5043097855571\\
23	-12.6112648787583\\
22	-12.7038852993791\\
21	-12.7740234617254\\
20	-12.9044870630136\\
19	-12.9370237787094\\
18	-12.9928766222292\\
17	-13.0053349398632\\
16	-13.0267355416155\\
15	-13.0617502118511\\
14	-13.0921645421823\\
13	-13.1136870155823\\
12	-13.1494089120715\\
11	-13.1804493840922\\
10	-13.1982032521587\\
9	-13.2511277794872\\
8	-13.2889547201154\\
7	-13.2991144499075\\
6	-13.3029770637766\\
5	-13.2137607663533\\
4	-13.0990822823163\\
3	-12.9614535017196\\
2	-12.8811524544365\\
1	-12.8771049993142\\
}--cycle;

\addplot [color=white!55!mycolor3,solid,forget plot]
  table[row sep=crcr]{%
1	-12.9322347014806\\
2	-13.0095242004891\\
3	-13.1017715384092\\
4	-13.2981197730883\\
5	-13.430915191158\\
6	-13.5159149848724\\
7	-13.5563549574365\\
8	-13.5694395981143\\
9	-13.5322963808066\\
10	-13.4581249541529\\
11	-13.3853929414074\\
12	-13.349993017645\\
13	-13.2868329040913\\
14	-13.2603716336268\\
15	-13.247402648989\\
16	-13.2106462634961\\
17	-13.1563502691084\\
18	-13.1775350543842\\
19	-13.1328172899385\\
20	-13.0851262462001\\
21	-13.0202991558685\\
22	-12.9178832045742\\
23	-12.8323381036083\\
24	-12.7631493560865\\
25	-12.6834540555873\\
26	-12.6169401709716\\
27	-12.5616598106398\\
28	-12.4469764500756\\
29	-12.3740632687606\\
30	-12.3006110723645\\
31	-12.2142195667587\\
32	-12.1545359685878\\
33	-12.0740413410591\\
34	-11.9984944591339\\
35	-11.9425510635944\\
36	-11.8573901174916\\
37	-11.7863354416219\\
38	-11.7305450931553\\
39	-11.643965024019\\
40	-11.6224553076041\\
41	-11.5031544107653\\
42	-11.465257628898\\
43	-11.4179549181895\\
44	-11.3252126175166\\
45	-11.260977861194\\
46	-11.1961689709994\\
47	-11.1462026912051\\
48	-11.0799216335955\\
49	-11.0118108952012\\
50	-10.9482008518544\\
51	-10.8997986078294\\
52	-10.8327522966284\\
53	-10.7713882410808\\
54	-10.6765029933136\\
55	-10.6361244505785\\
56	-10.5662233144206\\
57	-10.5115303506242\\
58	-10.4444160506824\\
59	-10.3792511813253\\
60	-10.3300336719907\\
61	-10.2884040315383\\
62	-10.2461044004575\\
63	-10.1837398623502\\
64	-10.1098465424272\\
65	-10.0247000375904\\
66	-10.0137555337507\\
67	-9.95405241976507\\
68	-9.92709230162255\\
69	-9.84746219667929\\
70	-9.80336864765488\\
71	-9.75798072303103\\
72	-9.74312121867581\\
73	-9.68342929966209\\
74	-9.64124767123355\\
75	-9.59671962853753\\
76	-9.60695157208452\\
77	-9.55646351366437\\
78	-9.50908308815571\\
79	-9.45754556051774\\
80	-9.38636067497994\\
81	-9.37033781374462\\
82	-9.36767029995173\\
83	-9.3311102496807\\
84	-9.28280863801116\\
85	-9.2888479445723\\
86	-9.27788686212465\\
87	-9.22883743217734\\
88	-9.22388054946078\\
89	-9.18804093306757\\
90	-9.15948934432609\\
91	-9.15610249236961\\
92	-9.12816012126077\\
93	-9.1061816823333\\
94	-9.09950619336174\\
95	-9.07005104922814\\
96	-9.06063017468448\\
97	-9.00986931603618\\
98	-9.02436487868773\\
99	-9.00848077206592\\
100	-9.00848077206592\\
};
\addplot [color=white!55!mycolor3,solid,forget plot]
  table[row sep=crcr]{%
1	-12.8771049993142\\
2	-12.8811524544365\\
3	-12.9614535017196\\
4	-13.0990822823163\\
5	-13.2137607663533\\
6	-13.3029770637766\\
7	-13.2991144499075\\
8	-13.2889547201154\\
9	-13.2511277794872\\
10	-13.1982032521587\\
11	-13.1804493840922\\
12	-13.1494089120715\\
13	-13.1136870155823\\
14	-13.0921645421823\\
15	-13.0617502118511\\
16	-13.0267355416155\\
17	-13.0053349398632\\
18	-12.9928766222292\\
19	-12.9370237787094\\
20	-12.9044870630136\\
21	-12.7740234617254\\
22	-12.7038852993791\\
23	-12.6112648787583\\
24	-12.5043097855571\\
25	-12.3790509381396\\
26	-12.2842935688498\\
27	-12.1906787336194\\
28	-12.0871305926921\\
29	-12.011827531492\\
30	-11.9038259607777\\
31	-11.8160434058546\\
32	-11.6980476397426\\
33	-11.5875328011972\\
34	-11.5318174137414\\
35	-11.3904855591328\\
36	-11.3127605223502\\
37	-11.2052551064994\\
38	-11.0900024712871\\
39	-10.9876330841981\\
40	-10.8975810313831\\
41	-10.7919695095174\\
42	-10.6959038500043\\
43	-10.5857146490186\\
44	-10.4762938963533\\
45	-10.3999519315684\\
46	-10.2942231719965\\
47	-10.178286675042\\
48	-10.0758314648675\\
49	-9.98682162747724\\
50	-9.89937653097599\\
51	-9.82338313562993\\
52	-9.71515528788501\\
53	-9.64604556458151\\
54	-9.53706886969175\\
55	-9.44878522145168\\
56	-9.39577360536699\\
57	-9.26462863766276\\
58	-9.20883399367772\\
59	-9.11577703184197\\
60	-9.04763499027455\\
61	-8.94945674229475\\
62	-8.90386412543385\\
63	-8.81549445029552\\
64	-8.74232676502489\\
65	-8.6908384503051\\
66	-8.63832615036469\\
67	-8.58112721121013\\
68	-8.48008897313136\\
69	-8.43467745684972\\
70	-8.36940900386269\\
71	-8.29527557935196\\
72	-8.22494597048634\\
73	-8.21767342614369\\
74	-8.13747041011114\\
75	-8.09985974956333\\
76	-8.04681511168041\\
77	-8.0133137764213\\
78	-7.95738072054221\\
79	-7.93128350448747\\
80	-7.90035856385354\\
81	-7.86946720658264\\
82	-7.81908454150282\\
83	-7.78827057877805\\
84	-7.75301709910105\\
85	-7.74610090941019\\
86	-7.70947245529419\\
87	-7.66742142907906\\
88	-7.66183372344072\\
89	-7.63636560946712\\
90	-7.63671557656485\\
91	-7.59859154524355\\
92	-7.57490845256365\\
93	-7.5618656972247\\
94	-7.54816444682695\\
95	-7.55673144935163\\
96	-7.53911100480769\\
97	-7.5298845153176\\
98	-7.50703508761956\\
99	-7.52756516347543\\
100	-7.52756516347543\\
};
\addplot [color=mycolor3,solid,line width = 1.0pt,mark=o]
  table[row sep=crcr]{%
1	-12.9046698503974\\
2	-12.9453383274628\\
3	-13.0316125200644\\
4	-13.1986010277023\\
5	-13.3223379787557\\
6	-13.4094460243245\\
7	-13.427734703672\\
8	-13.4291971591148\\
9	-13.3917120801469\\
10	-13.3281641031558\\
11	-13.2829211627498\\
12	-13.2497009648583\\
13	-13.2002599598368\\
14	-13.1762680879046\\
15	-13.1545764304201\\
16	-13.1186909025558\\
17	-13.0808426044858\\
18	-13.0852058383067\\
19	-13.0349205343239\\
20	-12.9948066546068\\
21	-12.8971613087969\\
22	-12.8108842519767\\
23	-12.7218014911833\\
24	-12.6337295708218\\
25	-12.5312524968635\\
26	-12.4506168699107\\
27	-12.3761692721296\\
28	-12.2670535213839\\
29	-12.1929454001263\\
30	-12.1022185165711\\
31	-12.0151314863066\\
32	-11.9262918041652\\
33	-11.8307870711281\\
34	-11.7651559364376\\
35	-11.6665183113636\\
36	-11.5850753199209\\
37	-11.4957952740607\\
38	-11.4102737822212\\
39	-11.3157990541085\\
40	-11.2600181694936\\
41	-11.1475619601414\\
42	-11.0805807394512\\
43	-11.0018347836041\\
44	-10.900753256935\\
45	-10.8304648963812\\
46	-10.745196071498\\
47	-10.6622446831235\\
48	-10.5778765492315\\
49	-10.4993162613392\\
50	-10.4237886914152\\
51	-10.3615908717297\\
52	-10.2739537922567\\
53	-10.2087169028312\\
54	-10.1067859315027\\
55	-10.0424548360151\\
56	-9.98099845989382\\
57	-9.88807949414348\\
58	-9.82662502218006\\
59	-9.74751410658365\\
60	-9.68883433113261\\
61	-9.61893038691653\\
62	-9.57498426294569\\
63	-9.49961715632285\\
64	-9.42608665372604\\
65	-9.35776924394775\\
66	-9.32604084205768\\
67	-9.2675898154876\\
68	-9.20359063737695\\
69	-9.1410698267645\\
70	-9.08638882575879\\
71	-9.0266281511915\\
72	-8.98403359458107\\
73	-8.95055136290289\\
74	-8.88935904067234\\
75	-8.84828968905043\\
76	-8.82688334188247\\
77	-8.78488864504283\\
78	-8.73323190434896\\
79	-8.69441453250261\\
80	-8.64335961941674\\
81	-8.61990251016363\\
82	-8.59337742072728\\
83	-8.55969041422938\\
84	-8.51791286855611\\
85	-8.51747442699124\\
86	-8.49367965870942\\
87	-8.4481294306282\\
88	-8.44285713645075\\
89	-8.41220327126734\\
90	-8.39810246044547\\
91	-8.37734701880658\\
92	-8.35153428691221\\
93	-8.334023689779\\
94	-8.32383532009435\\
95	-8.31339124928988\\
96	-8.29987058974609\\
97	-8.26987691567689\\
98	-8.26569998315364\\
99	-8.26802296777068\\
100	-8.26802296777068\\
};
\addlegendentry{dz = 10}

\addplot[area legend,solid,fill=mycolor4,opacity=1.000000e-01,draw=none,forget plot]
table[row sep=crcr] {%
x	y\\
1	-13.0093233198995\\
2	-13.0699283968992\\
3	-13.168102924309\\
4	-13.25501886154\\
5	-13.3770557105259\\
6	-13.4702366920475\\
7	-13.5544976807834\\
8	-13.5688292938794\\
9	-13.5704057064082\\
10	-13.5284574121554\\
11	-13.5150600798025\\
12	-13.5013373942813\\
13	-13.4747664673337\\
14	-13.4574505957264\\
15	-13.4427396708535\\
16	-13.4585058302509\\
17	-13.4308619050974\\
18	-13.4016561023494\\
19	-13.3861045535946\\
20	-13.374915824943\\
21	-13.2604097701112\\
22	-13.1570485105217\\
23	-13.0693208589553\\
24	-12.9917609101417\\
25	-12.8752699689822\\
26	-12.7903819656445\\
27	-12.6956981765132\\
28	-12.6044847833274\\
29	-12.5426162261595\\
30	-12.4500913889141\\
31	-12.3206290372721\\
32	-12.208733215278\\
33	-12.1667798668277\\
34	-12.0754718039909\\
35	-11.972960549013\\
36	-11.9000781644055\\
37	-11.7960662574513\\
38	-11.7073036910642\\
39	-11.6231034095974\\
40	-11.5229957141268\\
41	-11.4535862026712\\
42	-11.3203362792201\\
43	-11.2459047618642\\
44	-11.1768073987171\\
45	-11.0808524752649\\
46	-10.9882341317043\\
47	-10.8868744109134\\
48	-10.790742078185\\
49	-10.7077031098216\\
50	-10.6283556036508\\
51	-10.5402774050154\\
52	-10.439581551006\\
53	-10.3390744284692\\
54	-10.2819274653387\\
55	-10.2017256759679\\
56	-10.0786151651484\\
57	-10.0052587326235\\
58	-9.9398189046773\\
59	-9.84401844432278\\
60	-9.77689129583759\\
61	-9.70700046755686\\
62	-9.62900381634818\\
63	-9.55608644204551\\
64	-9.48404870340319\\
65	-9.39530967996337\\
66	-9.33255096676027\\
67	-9.25655134008261\\
68	-9.16846265128628\\
69	-9.14880932678653\\
70	-9.03692183289189\\
71	-8.99093432891906\\
72	-8.92408240436307\\
73	-8.87935313613355\\
74	-8.83093527085826\\
75	-8.74159392891783\\
76	-8.71993240684156\\
77	-8.67448682624136\\
78	-8.61923937001541\\
79	-8.60008336748339\\
80	-8.55632225971454\\
81	-8.50271452320554\\
82	-8.4678168525144\\
83	-8.432475809912\\
84	-8.42861624061416\\
85	-8.36248626498229\\
86	-8.34555396152373\\
87	-8.29341682899329\\
88	-8.2706518310779\\
89	-8.23146305260459\\
90	-8.19611015981461\\
91	-8.19129651737881\\
92	-8.12311003288404\\
93	-8.11940326478939\\
94	-8.12536160646154\\
95	-8.09226065163175\\
96	-8.07543901068241\\
97	-8.04884433450759\\
98	-8.02298708558375\\
99	-8.02535601592429\\
100	-8.02535601592429\\
100	-6.58046777795557\\
99	-6.58046777795557\\
98	-6.59488942862289\\
97	-6.59006766688653\\
96	-6.60765302831959\\
95	-6.62954573567894\\
94	-6.65006362001673\\
93	-6.64310898727998\\
92	-6.63827349103688\\
91	-6.67857138605831\\
90	-6.69277782629426\\
89	-6.69809806526489\\
88	-6.73980650730608\\
87	-6.7393328499844\\
86	-6.78010635787696\\
85	-6.7990286434263\\
84	-6.82100048797313\\
83	-6.8626507205978\\
82	-6.88938063499297\\
81	-6.95273764254415\\
80	-6.98553353363493\\
79	-7.02349351957975\\
78	-7.08932593239333\\
77	-7.13039300343793\\
76	-7.19483513694402\\
75	-7.24067105602368\\
74	-7.32344195625018\\
73	-7.37870747701521\\
72	-7.45841996801815\\
71	-7.52539280161199\\
70	-7.60869845455595\\
69	-7.69225101001094\\
68	-7.77285820966874\\
67	-7.86428013281205\\
66	-7.93677753903439\\
65	-8.03598731863427\\
64	-8.13200115332712\\
63	-8.20911547878492\\
62	-8.30948794508434\\
61	-8.40770390913123\\
60	-8.52469043034843\\
59	-8.63342336722362\\
58	-8.74768200115894\\
57	-8.84750837957862\\
56	-8.94825650808802\\
55	-9.08014976479898\\
54	-9.18835879218385\\
53	-9.31427419652767\\
52	-9.4236882877969\\
51	-9.54633188849379\\
50	-9.6424570852914\\
49	-9.75848934378318\\
48	-9.89931149753373\\
47	-10.009603649416\\
46	-10.1256402597366\\
45	-10.245362594074\\
44	-10.3683927425022\\
43	-10.4779788784247\\
42	-10.5933728951894\\
41	-10.701542687769\\
40	-10.8229056941158\\
39	-10.931035787037\\
38	-11.0389222357352\\
37	-11.138400700256\\
36	-11.3078464624635\\
35	-11.3828181186863\\
34	-11.5123888888422\\
33	-11.6133089293896\\
32	-11.7341641405029\\
31	-11.840862689805\\
30	-11.9762275246306\\
29	-12.1159620796656\\
28	-12.2033343422051\\
27	-12.3131923329921\\
26	-12.4248897174479\\
25	-12.5521837304323\\
24	-12.6903602494109\\
23	-12.7515285427362\\
22	-12.8924367668367\\
21	-13.0231739459018\\
20	-13.1027492732071\\
19	-13.145850081978\\
18	-13.1729273640114\\
17	-13.2146250132171\\
16	-13.2187795481802\\
15	-13.269747052581\\
14	-13.2479478601829\\
13	-13.2562636124247\\
12	-13.3001120226899\\
11	-13.3172091215739\\
10	-13.3439259587994\\
9	-13.3603458338341\\
8	-13.3709349146314\\
7	-13.3677403798137\\
6	-13.361180810945\\
5	-13.2880984895006\\
4	-13.2051813007078\\
3	-13.0956851331818\\
2	-12.9544200102255\\
1	-12.9609797795641\\
}--cycle;

\addplot [color=white!55!mycolor4,solid,forget plot]
  table[row sep=crcr]{%
1	-13.0093233198995\\
2	-13.0699283968992\\
3	-13.168102924309\\
4	-13.25501886154\\
5	-13.3770557105259\\
6	-13.4702366920475\\
7	-13.5544976807834\\
8	-13.5688292938794\\
9	-13.5704057064082\\
10	-13.5284574121554\\
11	-13.5150600798025\\
12	-13.5013373942813\\
13	-13.4747664673337\\
14	-13.4574505957264\\
15	-13.4427396708535\\
16	-13.4585058302509\\
17	-13.4308619050974\\
18	-13.4016561023494\\
19	-13.3861045535946\\
20	-13.374915824943\\
21	-13.2604097701112\\
22	-13.1570485105217\\
23	-13.0693208589553\\
24	-12.9917609101417\\
25	-12.8752699689822\\
26	-12.7903819656445\\
27	-12.6956981765132\\
28	-12.6044847833274\\
29	-12.5426162261595\\
30	-12.4500913889141\\
31	-12.3206290372721\\
32	-12.208733215278\\
33	-12.1667798668277\\
34	-12.0754718039909\\
35	-11.972960549013\\
36	-11.9000781644055\\
37	-11.7960662574513\\
38	-11.7073036910642\\
39	-11.6231034095974\\
40	-11.5229957141268\\
41	-11.4535862026712\\
42	-11.3203362792201\\
43	-11.2459047618642\\
44	-11.1768073987171\\
45	-11.0808524752649\\
46	-10.9882341317043\\
47	-10.8868744109134\\
48	-10.790742078185\\
49	-10.7077031098216\\
50	-10.6283556036508\\
51	-10.5402774050154\\
52	-10.439581551006\\
53	-10.3390744284692\\
54	-10.2819274653387\\
55	-10.2017256759679\\
56	-10.0786151651484\\
57	-10.0052587326235\\
58	-9.9398189046773\\
59	-9.84401844432278\\
60	-9.77689129583759\\
61	-9.70700046755686\\
62	-9.62900381634818\\
63	-9.55608644204551\\
64	-9.48404870340319\\
65	-9.39530967996337\\
66	-9.33255096676027\\
67	-9.25655134008261\\
68	-9.16846265128628\\
69	-9.14880932678653\\
70	-9.03692183289189\\
71	-8.99093432891906\\
72	-8.92408240436307\\
73	-8.87935313613355\\
74	-8.83093527085826\\
75	-8.74159392891783\\
76	-8.71993240684156\\
77	-8.67448682624136\\
78	-8.61923937001541\\
79	-8.60008336748339\\
80	-8.55632225971454\\
81	-8.50271452320554\\
82	-8.4678168525144\\
83	-8.432475809912\\
84	-8.42861624061416\\
85	-8.36248626498229\\
86	-8.34555396152373\\
87	-8.29341682899329\\
88	-8.2706518310779\\
89	-8.23146305260459\\
90	-8.19611015981461\\
91	-8.19129651737881\\
92	-8.12311003288404\\
93	-8.11940326478939\\
94	-8.12536160646154\\
95	-8.09226065163175\\
96	-8.07543901068241\\
97	-8.04884433450759\\
98	-8.02298708558375\\
99	-8.02535601592429\\
100	-8.02535601592429\\
};
\addplot [color=white!55!mycolor4,solid,forget plot]
  table[row sep=crcr]{%
1	-12.9609797795641\\
2	-12.9544200102255\\
3	-13.0956851331818\\
4	-13.2051813007078\\
5	-13.2880984895006\\
6	-13.361180810945\\
7	-13.3677403798137\\
8	-13.3709349146314\\
9	-13.3603458338341\\
10	-13.3439259587994\\
11	-13.3172091215739\\
12	-13.3001120226899\\
13	-13.2562636124247\\
14	-13.2479478601829\\
15	-13.269747052581\\
16	-13.2187795481802\\
17	-13.2146250132171\\
18	-13.1729273640114\\
19	-13.145850081978\\
20	-13.1027492732071\\
21	-13.0231739459018\\
22	-12.8924367668367\\
23	-12.7515285427362\\
24	-12.6903602494109\\
25	-12.5521837304323\\
26	-12.4248897174479\\
27	-12.3131923329921\\
28	-12.2033343422051\\
29	-12.1159620796656\\
30	-11.9762275246306\\
31	-11.840862689805\\
32	-11.7341641405029\\
33	-11.6133089293896\\
34	-11.5123888888422\\
35	-11.3828181186863\\
36	-11.3078464624635\\
37	-11.138400700256\\
38	-11.0389222357352\\
39	-10.931035787037\\
40	-10.8229056941158\\
41	-10.701542687769\\
42	-10.5933728951894\\
43	-10.4779788784247\\
44	-10.3683927425022\\
45	-10.245362594074\\
46	-10.1256402597366\\
47	-10.009603649416\\
48	-9.89931149753373\\
49	-9.75848934378318\\
50	-9.6424570852914\\
51	-9.54633188849379\\
52	-9.4236882877969\\
53	-9.31427419652767\\
54	-9.18835879218385\\
55	-9.08014976479898\\
56	-8.94825650808802\\
57	-8.84750837957862\\
58	-8.74768200115894\\
59	-8.63342336722362\\
60	-8.52469043034843\\
61	-8.40770390913123\\
62	-8.30948794508434\\
63	-8.20911547878492\\
64	-8.13200115332712\\
65	-8.03598731863427\\
66	-7.93677753903439\\
67	-7.86428013281205\\
68	-7.77285820966874\\
69	-7.69225101001094\\
70	-7.60869845455595\\
71	-7.52539280161199\\
72	-7.45841996801815\\
73	-7.37870747701521\\
74	-7.32344195625018\\
75	-7.24067105602368\\
76	-7.19483513694402\\
77	-7.13039300343793\\
78	-7.08932593239333\\
79	-7.02349351957975\\
80	-6.98553353363493\\
81	-6.95273764254415\\
82	-6.88938063499297\\
83	-6.8626507205978\\
84	-6.82100048797313\\
85	-6.7990286434263\\
86	-6.78010635787696\\
87	-6.7393328499844\\
88	-6.73980650730608\\
89	-6.69809806526489\\
90	-6.69277782629426\\
91	-6.67857138605831\\
92	-6.63827349103688\\
93	-6.64310898727998\\
94	-6.65006362001673\\
95	-6.62954573567894\\
96	-6.60765302831959\\
97	-6.59006766688653\\
98	-6.59488942862289\\
99	-6.58046777795557\\
100	-6.58046777795557\\
};
\addplot [color=mycolor4,dashed,line width = 1.0pt]
  table[row sep=crcr]{%
1	-12.9851515497318\\
2	-13.0121742035624\\
3	-13.1318940287454\\
4	-13.2301000811239\\
5	-13.3325771000132\\
6	-13.4157087514963\\
7	-13.4611190302985\\
8	-13.4698821042554\\
9	-13.4653757701211\\
10	-13.4361916854774\\
11	-13.4161346006882\\
12	-13.4007247084856\\
13	-13.3655150398792\\
14	-13.3526992279546\\
15	-13.3562433617173\\
16	-13.3386426892155\\
17	-13.3227434591572\\
18	-13.2872917331804\\
19	-13.2659773177863\\
20	-13.2388325490751\\
21	-13.1417918580065\\
22	-13.0247426386792\\
23	-12.9104247008458\\
24	-12.8410605797763\\
25	-12.7137268497072\\
26	-12.6076358415462\\
27	-12.5044452547526\\
28	-12.4039095627662\\
29	-12.3292891529125\\
30	-12.2131594567724\\
31	-12.0807458635385\\
32	-11.9714486778904\\
33	-11.8900443981086\\
34	-11.7939303464165\\
35	-11.6778893338496\\
36	-11.6039623134345\\
37	-11.4672334788536\\
38	-11.3731129633997\\
39	-11.2770695983172\\
40	-11.1729507041213\\
41	-11.0775644452201\\
42	-10.9568545872048\\
43	-10.8619418201444\\
44	-10.7726000706096\\
45	-10.6631075346694\\
46	-10.5569371957204\\
47	-10.4482390301647\\
48	-10.3450267878594\\
49	-10.2330962268024\\
50	-10.1354063444711\\
51	-10.0433046467546\\
52	-9.93163491940147\\
53	-9.82667431249845\\
54	-9.7351431287613\\
55	-9.64093772038343\\
56	-9.51343583661824\\
57	-9.42638355610104\\
58	-9.34375045291812\\
59	-9.2387209057732\\
60	-9.15079086309301\\
61	-9.05735218834404\\
62	-8.96924588071626\\
63	-8.88260096041521\\
64	-8.80802492836516\\
65	-8.71564849929882\\
66	-8.63466425289733\\
67	-8.56041573644733\\
68	-8.47066043047751\\
69	-8.42053016839873\\
70	-8.32281014372392\\
71	-8.25816356526552\\
72	-8.19125118619061\\
73	-8.12903030657438\\
74	-8.07718861355422\\
75	-7.99113249247076\\
76	-7.95738377189279\\
77	-7.90243991483965\\
78	-7.85428265120437\\
79	-7.81178844353157\\
80	-7.77092789667473\\
81	-7.72772608287484\\
82	-7.67859874375368\\
83	-7.6475632652549\\
84	-7.62480836429364\\
85	-7.5807574542043\\
86	-7.56283015970034\\
87	-7.51637483948884\\
88	-7.50522916919199\\
89	-7.46478055893474\\
90	-7.44444399305444\\
91	-7.43493395171856\\
92	-7.38069176196046\\
93	-7.38125612603469\\
94	-7.38771261323913\\
95	-7.36090319365534\\
96	-7.341546019501\\
97	-7.31945600069706\\
98	-7.30893825710332\\
99	-7.30291189693993\\
100	-7.30291189693993\\
};
\addlegendentry{dz = 20}

\end{axis}
\end{tikzpicture}%